%% file: main.tex
\documentclass[10pt,twocolumn,letterpaper]{article}

\PassOptionsToPackage{table}{xcolor}
\usepackage[pagenumbers]{cvpr} 

\input{preamble}

\definecolor{cvprblue}{rgb}{0.21,0.49,0.74}
\usepackage[breaklinks,colorlinks,allcolors=cvprblue]{hyperref}

\def\confName{CVPR}
\def\confYear{2027}

\usepackage{textcomp}

\title{Self-Evolving Evaluation Environments for Computer-Use Agents}
\author{Wenhao Wang\textsuperscript{2,1}, Panrong Tong\textsuperscript{1}, Quyu Kong\textsuperscript{1}, Pengxiang Li\textsuperscript{1},\\
Haowen Yang\textsuperscript{3}, Chenglin Cai\textsuperscript{1}, Siheng Chen\textsuperscript{3$\dagger$}, Yue Wang\textsuperscript{1$\dagger$}\\
\textsuperscript{1}Alibaba Token Hub, Alibaba Group\\
\textsuperscript{2}Zhejiang University\\
\textsuperscript{3}Shanghai Jiao Tong University
}

\begin{document}

\maketitle

\begin{abstract}
Computer-use agents (CUAs) are rapidly improving, yet their evaluation remains largely static. Existing benchmarks provide a strong foundation, but fixed difficulty ceilings, limited feature coverage, and coarse aggregate scores progressively weaken their discriminative and diagnostic value. We introduce CUA-SEE, a closed-loop evaluation environment that co-evolves tasks and executable verifiers from rollout evidence. 
It disentangles three objectives: benchmark evolution expands evaluation scope through cohort-calibrated difficulty and coverage growth; diagnostic evolution increases resolution through agent-conditional capability-boundary probes; and verifier evolution improves precision by translating changing requirements into state-grounded checkpoints with adversarial validation. 
Across multiple agents, applications, and rounds, CUA-SEE generates challenging, measurable tasks while preserving process-level evidence. 
Our results position adaptive evaluation as a scalable counterpart to evolving CUAs, continuously tracking what agents can do, where their boundaries lie, and how reliably progress is measured.\par
\end{abstract}

\section{Introduction}
\label{sec:introduction}



Agent evaluation, which underpins both scientific progress and real-world deployment, has long been a fundamentally yet challenging problem \cite{evocodebench,yao$t$benchBenchmarkToolAgentUser2024,wang2026mcppersonabenchmarkingllmagents, sun2026agents}. 
For computer-use agents (CUAs) \cite{li-etal-2026-fully, shi2026cuauniversescalabledynamicenvironment, yan2026cuasandboxefficientenvironmentscomputeruse,zhou2026opengcuafullyopenstack}, evaluation is harder because success must be verified against intricate software-specific states and multifaceted task requirements, rather than textual responses (e.g., search agents \cite{mialon2023gaiabenchmarkgeneralai,li2025iterative,du2026recursiveselfevolvingagenticliterature}) or binary execution results (e.g., code agents \cite{jimenez2024swe}). 
Current benchmarks \cite{osworld,yangUltraCUAFoundationModel2025} have established an essential foundation for CUA evaluation by manually curating realistic tasks and task-specific verifiers.
However, this human-authored, one-shot construction paradigm is costly to scale and renders both tasks and verifiers largely fixed as agent capabilities evolve, giving rise to several critical limitations \cite{autoevoeval,jiang_sea-eval_2026,zheng_seagym_2026}.

This static paradigm creates three consequential limitations: (1) A fixed task set imposes a \emph{difficulty ceiling}: once agents master its tasks, scores saturate and cease to distinguish further progress. (2) Manual construction provides limited \emph{feature coverage}; in our audit of core OSWorld applications, the existing tasks exercise less than 20\% of the available software-feature taxonomy. (3) Static benchmarks lack \emph{evaluation dynamics}: they neither track a moving capability boundary nor actively search for newly emerging weaknesses, shortcut behaviors, and generalization gaps. 
These limitations reveal the necessity of an evaluation system that evolves in tandem with the LLM agents it assesses.

\begin{figure*}[t]
\centering
\begin{minipage}[t]{0.59\textwidth}
\vspace{0pt}
\centering
\includegraphics[width=\linewidth]{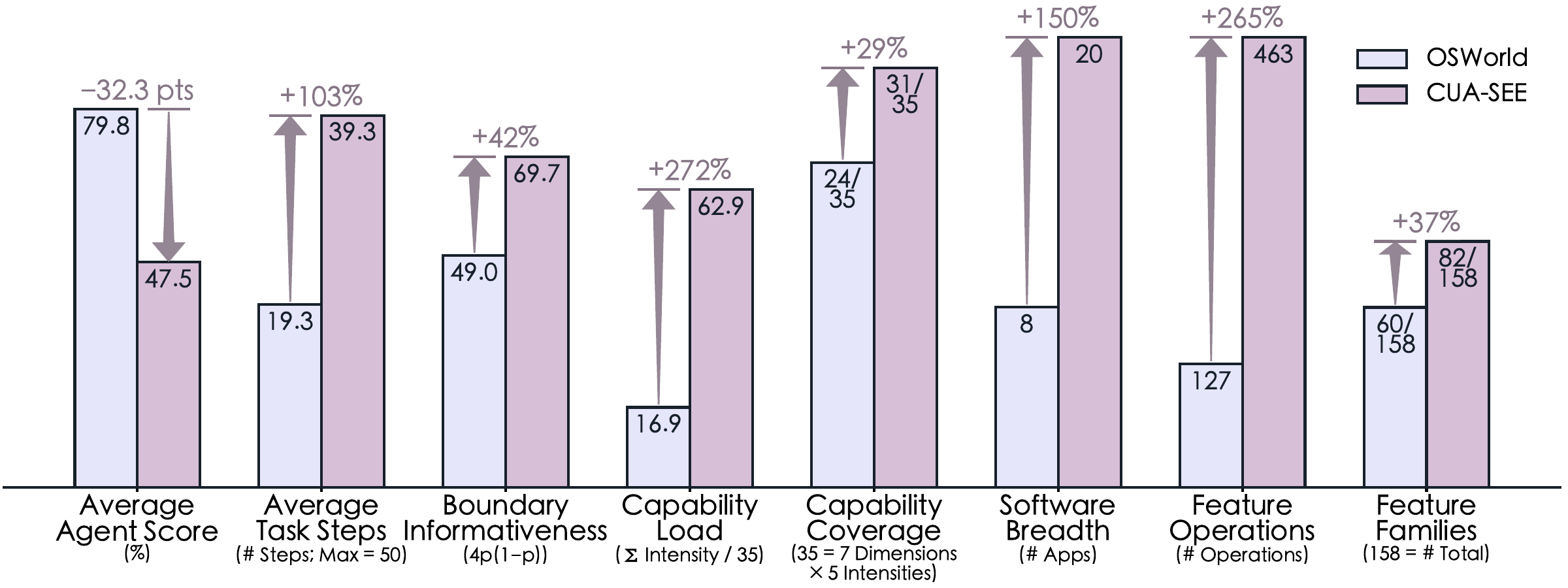}
\vspace{-5mm}
\captionof{figure}{Comparison between static benchmarks and CUA-SEE (Round 4). CUA-SEE progressively creates harder tasks requiring more complex
interactions and broadens coverage across capabilities, features,
and operations.}
\label{fig:osworld_vs_cua_see}
\end{minipage}
\hfill
\begin{minipage}[t]{0.39\textwidth}
\vspace{0pt}

\centering
\captionof{table}{Comparison of evolvable properties.}
\label{tab:benchmark_comparison}
\vspace{-2mm}
\setlength{\tabcolsep}{1.6pt}

\resizebox{\linewidth}{!}{%
\begin{tabular}{@{}lccccc@{}}
\toprule
\textbf{Benchmark}
& \makecell{Task\\Difficulty}
& \makecell{Feature\\Coverage}
& \makecell{Boundary\\Resolution}
& \makecell{Verifier\\Quality}
& \makecell{Process\\Signal} \\
\midrule
OSWorld & \xmark & \xmark & \xmark & \xmark & \xmark \\
TRACE & \cmark & \xmark & \xmark & \xmark & \cmark \\
CUA-Gym & \cmark & \cmark & \xmark & \cmark & \xmark \\
OpenComputer & \xmark & \xmark & \xmark & \cmark & \cmark \\
\textbf{CUA-SEE} & \cmark & \cmark & \cmark & \cmark & \cmark \\
\bottomrule
\end{tabular}%
}

\vspace{0.5em}
\centering
\includegraphics[width=\linewidth]{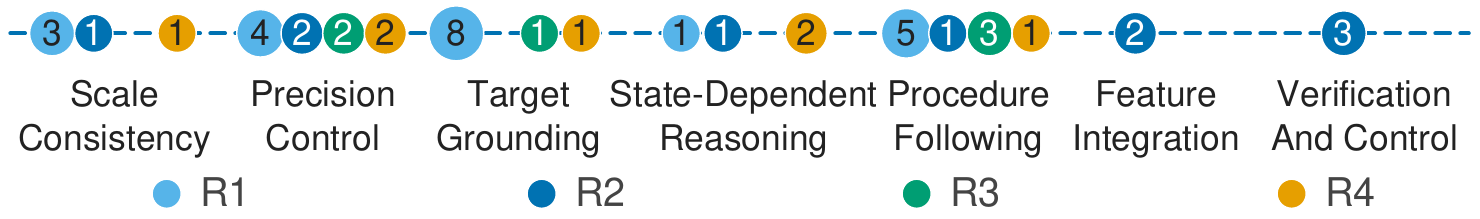}
\vspace{-5mm}
\captionof{figure}{Number of diagnostic capability boundaries found for Kimi across evolution rounds.}
\label{fig:kimi_diagnostic_boundary}
\end{minipage}
\vspace{-1mm}
\end{figure*}

To address the aforementioned challenges, we propose \textbf{CUA-SEE}, a \textbf{S}elf-\textbf{E}volving \textbf{E}valuation environment for computer-use agents. 
CUA-SEE reconceptualizes evaluation as a dynamic, closed-loop process where tasks, executable verifiers, and agent rollout evidence continuously inform and refine one another.
At each round, agents interact with the current task-verifier pairs; validity-filtered trajectories and checkpoint outcomes update the measurement state; and this evidence determines what tasks should be measured in the next round and how the corresponding verifiers should be implemented.
Specifically, our framework is designed around three core components that work in concert to disentangle three dimensions often conflated in conventional task-hardening pipelines: \emph{evaluation scope, diagnostic resolution, and measurement precision}.


\textbf{Scope.}
To address the fixed difficulty ceiling and feature coverage, we expand the evaluation scope of static benchmarks by introducing an evolution operator, called \textit{Benchmark Task Evolution}, which constructs a shared, model-neutral benchmark from aligned trajectories of a heterogeneous agent cohort. 
It retains all replicates, separates execution validity from task success, and uses cohort agreement and disagreement to calibrate whether a source task is mastered, discriminating, unstable, or confounded. 
Evolution then optimizes two suite-level objectives: advancing the measurable difficulty frontier and expanding coverage over exact software-feature-operation cells and capability-intensity cells. Candidate descendants may increase precision, scale, state dependence, procedure depth, feature composition, or add a previously uncovered but coherent operation. 
Each descendant preserves a semantic lineage anchor and must introduce explicit, independently observable measurement obligations, while a round-level registry coordinates coverage growth across tasks. Benchmark evolution thus broadens \emph{what} the suite can measure without tailoring the benchmark to an idiosyncratic failure of one agent.

\textbf{Resolution.}
The second operator, called \textit{Diagnostic Task Evolution}, is agent-conditional and concentrates evaluation budget near uncertain capability boundaries. 
A batch-level planner aggregates longitudinal evidence into capability, software, and feature maps to dispatch targeted experiments over seven calibrated dimensions such as precision control and state-dependent reasoning, as detailed in Appendix~\ref{sec:appendix_concepts}.
The task-level evolver realizes each experiment as a natural task with one primary dimension at a controlled intensity. 
The search is adaptive rather than monotonically harder: 
Clean successes and failures prompt upward and downward probes, respectively; partial or censored runs trigger deconfounding.
This adaptive bracketing refines \emph{where} an agent's pass-fail transition lies, turning coarse benchmark scores into locally attributable capability evidence.

\textbf{Precision.}
Both task operators share a requirement-preserving measurement backbone, termed as \textit{Verifier Evolution}.
The third component uses task-agnostic software adapters to probe native state-export surfaces and normalize them into a shared \texttt{CanonicalState} schema. Over this substrate, an information-isolated verifier agent generates atomic requirement-level checkpoints while a pass agent independently realizes a valid completion state; their adversarial cross-review improves checkpoint completeness and PRE/PASS discrimination \cite{cuagym, opencomputer}. Verifier evolution thus determines \emph{how} faithfully evolving agent behavior is measured while exposing process-level signals for diagnosis and future dense-feedback learning.

Extensive experiments across five SOTA agents demonstrate that CUA-SEE continuously restores evaluation informativeness. Over four evolution rounds, the macro-average agent score decreases from 60.6 to 47.5 while trajectory length increases from 27.4 to 39.3 steps, and exact feature coverage expands by 3.65$\times$.
Beyond producing harder and broader tasks, CUA-SEE adaptively concentrates diagnostic probes near each agent’s capability boundary, revealing agent-specific and software-specific weaknesses obscured by aggregate scores. 
Meanwhile, the evolved checkpoint verifiers achieve 84.0\% agreement with human trajectory judgments. 
To summarize, our main contributions are:
\begin{enumerate}[nosep]
\item We formulate self-evolving evaluation for CUAs as a closed loop in which task demand, measurement instruments, and validity-filtered rollout evidence are updated jointly across rounds, addressing the mismatch between continually improving agents and static benchmarks.
\item We introduce two complementary task-evolution operators: a cohort-grounded benchmark operator that expands evaluation scope through measurable frontier and feature-coverage growth, and an agent-conditional diagnostic operator that increases resolution through controlled, adaptive capability-boundary probes.
\item We develop a requirement-preserving verifier evolution pipeline that improves measurement precision by converting evolved task obligations into executable, state-grounded checkpoints, enabling validity-aware capability estimation and richer process supervision than terminal scalar rewards alone.
\end{enumerate}

\section{Related Work}
\label{sec:related-work}

\begin{figure*}[t]
\centering
\includegraphics[width=1\linewidth]{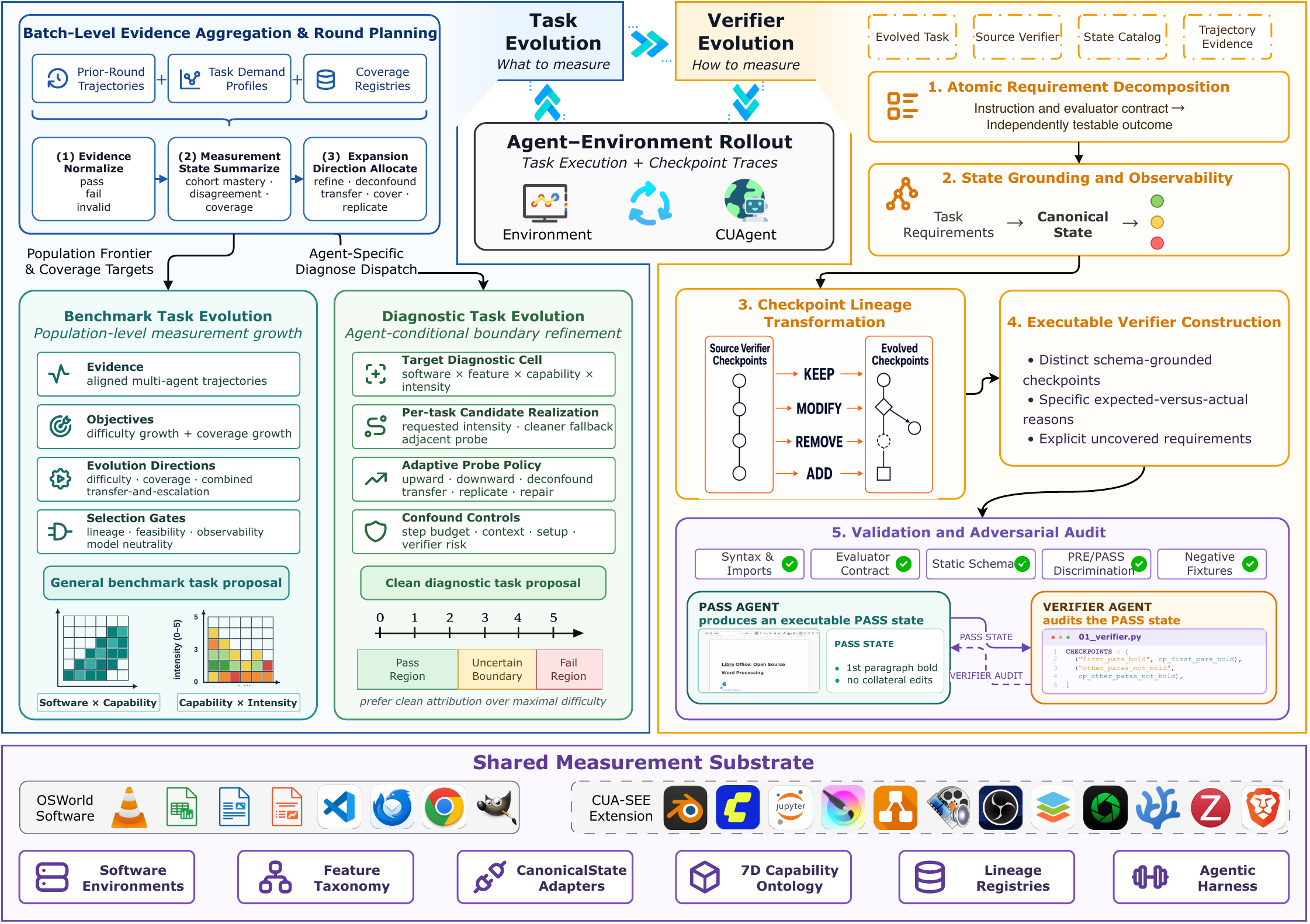}
\vspace{-6mm}
\caption{System overview. 
CUA-SEE forms a closed loop of task evolution, verifier evolution, and agent-environment rollout, as summarized by the central workflow.
The left panel details task evolution: rollout evidence is first aggregated at the batch level to plan the next round, followed by two complementary forms of per-task evolution: benchmark evolution for population-level measurement growth and diagnostic evolution for agent-conditional boundary refinement. The right panel presents verifier evolution, which progressively translates evolved task requirements into grounded, executable, and rigorously validated measurements. 
The bottom panel shows the shared substrate that supports all stages of the evaluation loop, expanding the 8 OSWorld applications to 20 verifiable desktop environments.}
\label{fig:main}
\vspace{-2mm}
\end{figure*}


\paragraph{CUA Environments \& Benchmarks.}
While popular benchmarks like OSWorld and WindowsAgentArena \cite{osworld,windowsagentarena,wang2026fedguibenchmarkingfederatedgui} offer robust, reproducible evaluation of computer-use agents in real desktop environments, they are limited to a fixed set of pre-curated tasks.
As agent capability improves, extending such benchmarks by hand is costly, difficult to scale, and slow to expose new capability limits.
Recent systems automate parts of this process: CUA-Gym \cite{cuagym} generates verifiable task-state-reward tuples, while OpenComputer \cite{opencomputer} builds tasks with state verifiers and partial-credit rewards.
Their main goal, however, is to scale verifiable environments or training data rather than repeatedly adapt a benchmark to the observed behavior of a fixed agent.

\paragraph{Evaluation Evolution.}
Early efforts have begun to explore benchmark evolution through rule-based editing, scheduled data refreshes, and agent-driven exploration \cite{benchmarkselfevolving,autoevoeval,evocodebench}.
AutoEvoEval applies predefined transformations to reasoning or question-answering tasks, while EvoCodeBench periodically adds new code tasks.
Most closely related, TRACE \cite{trace} uses test-time exploration to generate progressively harder web-search tasks and verify their solutions for reproducibility.
Unlike our approach, however, TRACE's scope is limited to task difficulty, and it leaves the critical role of verifier evolution unaddressed by employing a static, exact-match verifier.

\section{Methodology}
\label{sec:method}


As depicted in Figure \ref{fig:main}, CUA-SEE is designed to co-evolve tasks and verifiers through a closed loop fueled by evidence from agent-environment rollouts.

\begin{figure*}[t]
\centering

\begin{subfigure}[t]{0.49\textwidth}
\centering
\includegraphics[width=\linewidth]{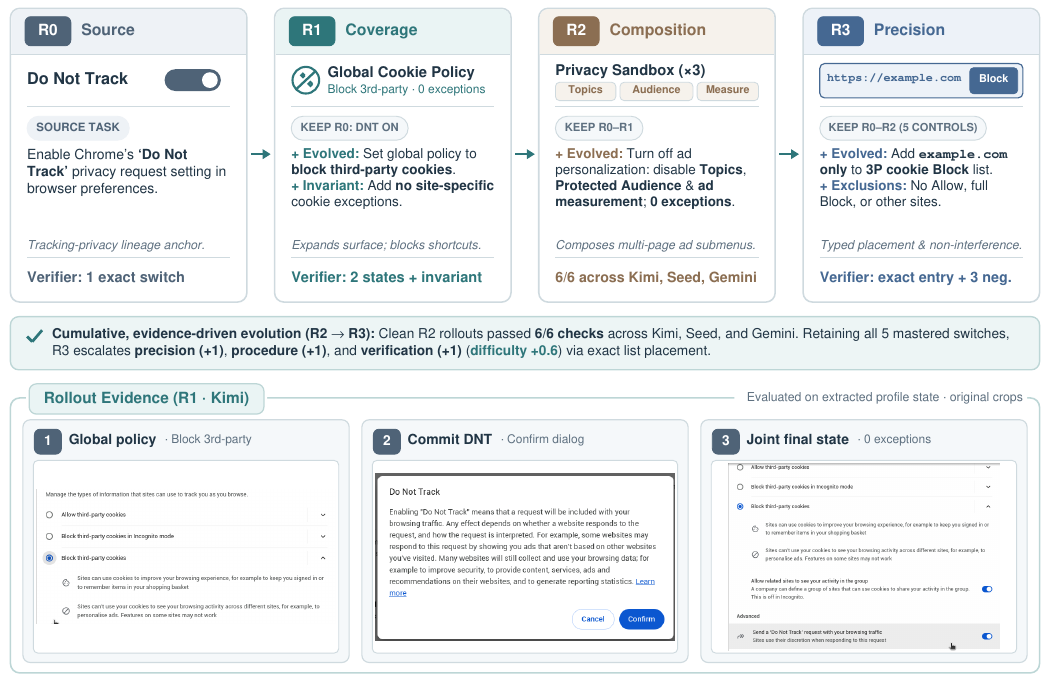}
\caption{Benchmark evolution using the Chrome privacy task.}
\label{fig:case_chrome_privacy}
\end{subfigure}
\hfill
\begin{subfigure}[t]{0.49\textwidth}
\centering
\includegraphics[width=\linewidth]{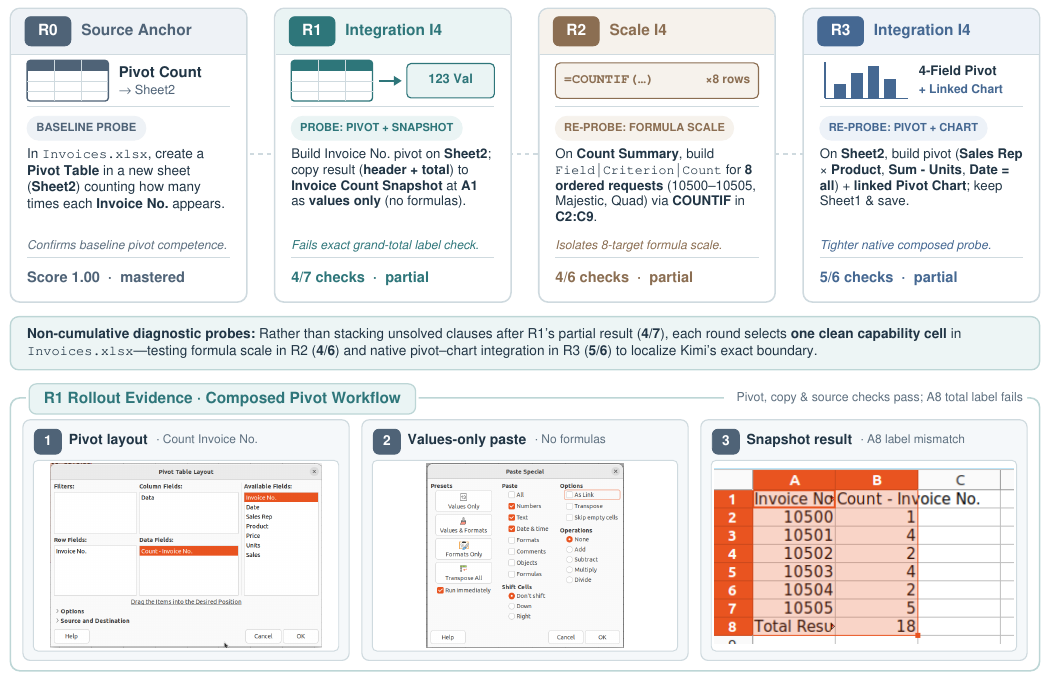}
\caption{Diagnosing Kimi using the LibreOffice Calc pivot-table task.}
\label{fig:case_gimp_export}
\end{subfigure}
\vspace{-1mm}
\caption{Case studies of benchmark evolution and diagnostic evolution. More examples and results are attached in the Appendix.}
\label{fig:benchmark_evolution_cases}
\vspace{-1mm}
\end{figure*}

\subsection{Scope: Benchmark Task Evolution}
\label{sec:benchmark_task_evolution}

Benchmark task evolution expands the population-level scope of evaluation. Given source tasks, aligned trajectories from a heterogeneous agent cohort, and the accumulated coverage registry, it produces a model-neutral batch of task descendants that advances the measurable difficulty frontier and enlarges the set of capabilities and software operations covered by the benchmark. 

\paragraph{Aligned Multi-Agent Evidence.} 
We join trajectories by canonical source-task identity and retain every compatible replicate. Evidence integrity is modeled independently of task outcome: valid successes and failures provide clean evidence, step-limit runs are retained as censored observations, and setup, controller, or environment failures are marked as invalid. Using scores, termination states, checkpoint transitions, regressions, and path diversity, we classify each source lineage as universally easy, mostly mastered, discriminating, universally hard, unstable, or confounded. 
This cohort-level state determines the admissible strength of evolution while preventing transient or model-specific errors from becoming benchmark objectives.

\paragraph{Bi-Objective Round Planning.} 
Evolution jointly optimizes difficulty-frontier and feature-coverage growth at the round level. Difficulty growth requires a measurable increase in one or more calibrated capability demands without material regression of the overall task, whereas coverage growth requires a strict set addition on either the exact software-taxonomy-operation surface or the capability-intensity surface. Required action families are tracked separately as exposure rather than treated as agent capabilities. A shared lineage and coverage registry coordinates parallel workers and prevents repeated references to an already measured feature from being counted as new breadth. Individual tasks may specialize in difficulty or coverage, but the selected batch must advance both objectives in aggregate.

\paragraph{Lineage-Anchored Evolution Directions.} 
For each eligible source, we construct candidates with four directions: \emph{difficulty escalation}, \emph{coverage expansion}, \emph{transfer with measurable escalation}, and their coherent \emph{combination}. Every candidate preserves a lineage anchor, the semantic demand that makes cross-round comparison meaningful, while allowing values, targets, scale, context, and supporting subgoals to change. 
Purely horizontal transfer is excluded: a transferred task must introduce an independently measurable demand or coverage cell. Candidates may increase precision, dependence, procedural depth, or verification requirements, but unrelated feature grafts and superficial elaboration do not constitute evolution.

\paragraph{Candidate Selection Gates.}
A candidate is eligible only if it preserves the lineage anchor, avoids material difficulty regression, remains feasible within the interaction budget, admits a uniquely observable final state, and is justified by model-neutral evidence. Eligible candidates are ranked by feasibility, verifiability, calibrated difficulty, coverage gain, model neutrality, and naturalness; step requirements above $0.85$ of the available budget are rejected. 
The output remains a general benchmark task proposal until every evolved obligation is instantiated by an executable verifier.


\subsection{Resolution: Diagnostic Task Evolution}
\label{sec:diagnostic_task_evolution}

Diagnostic task evolution increases the resolution of evaluation around an individual agent's capability boundary. 
This multi-agent system follows a hierarchical process: a batch-level planner converts longitudinal rollout evidence into agent-specific diagnostic dispatches, and a per-task evolver realizes each dispatch as a controlled task proposal. 
The objective is not maximal difficulty or scope (i.e., Benchmark Evolution), but a locally attributable observation that narrows the agent's pass-fail boundary.

\paragraph{Target Diagnostic Cell.}
To allocate well-targeted dispatches, the batch planner first separates pass, partial, confounded, and invalid outcomes, using only observations with valid setup and verifiers as clean boundary evidence. 
It then aggregates multiple analytical views to identify stable, failing, or insufficiently supported regions. 
Each dispatch targets a diagnostic cell defined by software, feature, primary capability, and intensity, together with control dimensions and confound limits. 
Priority is given to deconfounding blocked evidence, resolving uncertain transitions, testing transfer of suspected weaknesses, expanding under-supported cells, and replicating non-monotonic results.

\begin{table*}[t]
\caption{Evaluation results on evolved benchmarks of CUA-SEE.
The evolutionary process yields a consistent increase in task difficulty for all agents.
Pass is the percentage of tasks receiving a full score of 1.0.
}
\label{tab:benchmark-evolution-results}
\vspace{-1mm}
\centering
\small
\setlength{\tabcolsep}{5pt}
\resizebox{\textwidth}{!}{%
\begin{tabular}{lc*{8}{c}*{5}{>{\columncolor{gray!10}}c}}
\toprule
\multirow{2}{*}{\textbf{Agent}}
& \multirow{2}{*}{\textbf{Round}}
& \multirow{2}{*}{Chrome}
& \multirow{2}{*}{GIMP}
& \multirow{2}{*}{\makecell{LibreOffice\\Calc}}
& \multirow{2}{*}{\makecell{LibreOffice\\Impress}}
& \multirow{2}{*}{\makecell{LibreOffice\\Writer}}
& \multirow{2}{*}{ThunderBird}
& \multirow{2}{*}{VLC}
& \multirow{2}{*}{\makecell{VS Code}}
& \multicolumn{3}{c}{\cellcolor{gray!10}\textbf{Overall}}
& \multicolumn{2}{c}{\cellcolor{gray!10}\textbf{Tokens} (K)} \\
& & & & & & & & &
& \textbf{Score}$\uparrow$
& \textbf{Pass}$\uparrow$
& \textbf{Steps}$\downarrow$
& \textbf{Input}
& \textbf{Output} \\
\midrule

\multirow{5}{*}{\kimimoji{}~~Kimi-K2.6}
& R0 & 90.0 & 92.9 & 77.8 & 94.1 & 84.6 & 100.0 & 75.0 & 100.0 & 89.0 & 89.0 & 12.5 & 103.0 & 2.1 \\
& R1 & 80.8 & 54.9 & 68.3 & 83.4 & 54.5 & 36.0 & 57.3 & 64.7 & 64.8 & 41.0 & 23.7 & 201.5 & 4.4 \\
& R2 & 84.7 & 71.1 & 62.5 & 78.0 & 59.6 & 46.3 & 45.8 & 69.6 & 66.6 & 32.0 & 33.2 & 285.4 & 6.1 \\
& R3 & 93.5 & 60.4 & 49.3 & 65.8 & 53.8 & 47.7 & 49.1 & 52.0 & 58.9 & 22.0 & 39.9 & 346.7 & 7.9 \\
& R4 & 71.6 & 57.7 & 44.9 & 56.2 & 52.8 & 41.7 & 59.3 & 47.7 & 53.6 & 19.0 & 41.8 & 363.9 & 7.8 \\
\midrule

\multirow{5}{*}{\seedmoji{}~~Seed-2.1-Pro}
& R0 & 80.0 & 78.6 & 77.8 & 70.6 & 84.6 & 85.7 & 87.5 & 69.2 & 78.0 & 78.0 & 14.4 & -- & -- \\
& R1 & 80.8 & 48.5 & 76.7 & 76.0 & 54.5 & 32.1 & 34.4 & 62.0 & 61.7 & 34.0 & 24.2 & 142.4 & 12.7 \\
& R2 & 71.8 & 56.0 & 48.5 & 69.6 & 43.4 & 48.3 & 42.7 & 54.8 & 55.1 & 26.0 & 23.4 & 141.8 & 12.1 \\
& R3 & 86.7 & 58.4 & 47.0 & 58.2 & 37.7 & 45.9 & 48.4 & 48.2 & 53.5 & 20.0 & 29.7 & 184.1 & 14.5 \\
& R4 & 87.3 & 34.0 & 30.7 & 51.3 & 35.2 & 43.7 & 40.4 & 46.2 & 44.6 & 13.0 & 30.9 & 193.3 & 14.6 \\
\midrule

\multirow{5}{*}{\Geminiemoji{}~~Gemini-3.5-Flash}
& R0 & 90.0 & 57.1 & 94.4 & 70.6 & 92.3 & 100.0 & 75.0 & 69.2 & 80.0 & 80.0 & 14.1 & 74.4 & 0.3 \\
& R1 & 83.3 & 50.6 & 70.1 & 77.9 & 38.7 & 29.8 & 55.2 & 64.6 & 61.2 & 38.0 & 29.0 & 154.0 & 9.2 \\
& R2 & 73.0 & 48.6 & 56.7 & 69.5 & 36.1 & 48.3 & 43.8 & 73.5 & 57.2 & 29.0 & 36.7 & 203.9 & 10.6 \\
& R3 & 84.1 & 38.5 & 43.8 & 56.8 & 48.1 & 47.4 & 57.8 & 58.8 & 53.2 & 22.0 & 43.0 & 241.1 & 14.7 \\
& R4 & 82.4 & 36.8 & 35.0 & 47.4 & 48.9 & 46.7 & 60.9 & 54.4 & 49.3 & 16.0 & 44.6 & 254.5 & 15.4 \\
\midrule

\multirow{5}{*}{\Qwenemoji{}~~Qwen-3.7-Plus}
& R0 & 100.0 & 64.3 & 50.0 & 76.5 & 61.5 & 100.0 & 62.5 & 92.3 & 73.0 & 73.0 & 34.0 & 520.7 & 10.7 \\
& R1 & 75.0 & 54.9 & 48.1 & 82.2 & 52.2 & 40.7 & 36.5 & 70.3 & 59.5 & 35.0 & 35.0 & 398.8 & 8.1 \\
& R2 & 87.0 & 62.1 & 51.9 & 70.9 & 54.0 & 45.9 & 69.8 & 51.7 & 61.3 & 35.0 & 39.6 & 436.7 & 9.0 \\
& R3 & 82.6 & 49.4 & 35.4 & 68.2 & 46.9 & 37.5 & 56.4 & 49.5 & 52.8 & 19.0 & 43.7 & 471.5 & 9.3 \\
& R4 & 63.6 & 40.2 & 36.6 & 59.4 & 38.7 & 46.9 & 52.6 & 37.9 & 46.1 & 16.0 & 47.1 & 487.6 & 9.9 \\
\midrule

\multirow{5}{*}{\Openaiemoji{}~~GPT-5.5}
& R0 & 80.0 & 64.3 & 61.1 & 70.6 & 61.5 & 100.0 & 87.5 & 46.2 & 68.0 & 68.0 & 22.6 & 140.9 & 17.3 \\
& R1 & 63.3 & 47.7 & 49.7 & 76.7 & 36.4 & 33.6 & 67.7 & 65.4 & 56.0 & 31.0 & 24.9 & 183.9 & 22.4 \\
& R2 & 67.0 & 57.9 & 68.7 & 48.4 & 38.2 & 44.2 & 83.3 & 62.8 & 58.3 & 37.0 & 30.7 & 247.3 & 35.9 \\
& R3 & 83.7 & 60.3 & 49.1 & 40.5 & 21.5 & 26.8 & 72.9 & 66.1 & 51.6 & 32.0 & 30.8 & 283.6 & 39.7 \\
& R4 & 50.5 & 43.8 & 30.1 & 38.7 & 26.1 & 37.5 & 73.4 & 68.4 & 44.0 & 18.0 & 31.9 & 296.8 & 46.8 \\
\bottomrule
\end{tabular}%
}
\vspace{-2mm}
\end{table*}

\paragraph{KNN-Inspired Evolution Candidate Realization.} 
For one dispatched source lineage, the task evolver constructs a local neighborhood of candidates: the requested intensity, a cleaner lower-intensity fallback, and, when feasible, an adjacent higher-intensity or alternate-factor probe. 
Each task is defined by a primary demand dimension and several secondary dimensions, all drawn from the 7D capabilities: \emph{scale consistency, precision control, target grounding, state-dependent reasoning, procedure following, feature integration, or verification and control}. Intensities are scored from 0 to 5 using dimension-specific factors derived from the instruction, initial state, and verifier obligations; they describe task demand rather than predicted agent performance (detailed in Appendix~\ref{sec:appendix_concepts}). 
\paragraph{Adaptive Probe Policy.} 
The probe direction is evidence-driven rather than assumed to be monotonically upward. 
Clean passes or failures adjust intensity upward or downward, respectively, while ambiguous results trigger corrective actions like deconfounding, simplification, or replication.
Each proposal records expected failure modes and explicit verifier requirements, and it updates the agent's capability boundary only after verifier evolution and a valid rollout make the resulting evidence eligible.
\paragraph{Confound-Controlled Candidate Selection.} 
Capability demand is factorized from nuisance variables including step budget, context load and verifier risk. Candidates that violate dispatch limits, exceed the step budget, or lack an observable outcome are redesigned or rejected. 
Remaining ones are ranked by dispatch alignment, diagnostic cleanliness, secondary coverage value, etc. 
This criterion deliberately favors a slightly weaker but cleanly attributable probe over a harder task which cannot resolve agent boundaries.

\subsection{Precision: Unified Verifier Evolution}
\label{sec:verifier_evolution}

Verifier evolution co-evolves the measurement instrument with each task so that increased task demand remains faithfully observable. 
The objective is to maximize coverage of the evolved task contract while preserving verifier lineage, grounding every check in observable state, and explicitly representing verification gaps.


\paragraph{Requirement-Centric Contract Decomposition.} 
We reconstruct the evolved task contract as a set of independently testable outcome requirements derived from the explicit evaluator requirements and atomic clauses of the instruction. Target values, ordering constraints, artifact states, preservation conditions, and prohibited side effects are retained as first-class obligations rather than silently omitted or merged into coarse success criteria.

\begin{table*}[t]
\caption{Hierarchical feature coverage across benchmark-evolution rounds. Evolution steadily expands functional breadth across all apps ($3.65\times$ operations by $R4$).
A feature cell is an exact $\langle$software, taxonomy category, operation$\rangle$ tuple. Frontier reports the cumulative operation-level coverage registry.
Capability Coverage is cumulative coverage over the 35 capability-intensity cells.
}
\label{tab:feature-coverage}
\vspace{-1mm}
\centering
\small
\setlength{\tabcolsep}{5pt}
\resizebox{\textwidth}{!}{%
\begin{tabular}{
c*{8}{c}
*{5}{>{\columncolor{gray!10}}c}}
\toprule
\multirow{2}{*}{\textbf{Round}}
& \multirow{2}{*}{Chrome}
& \multirow{2}{*}{GIMP}
& \multirow{2}{*}{\makecell{LibreOffice\\Calc}}
& \multirow{2}{*}{\makecell{LibreOffice\\Impress}}
& \multirow{2}{*}{\makecell{LibreOffice\\Writer}}
& \multirow{2}{*}{\makecell{ThunderBird}}
& \multirow{2}{*}{VLC}
& \multirow{2}{*}{VS Code}
&
&
& \multicolumn{2}{c}{\cellcolor{gray!10}\textbf{Feature Operation}}
& \\
& & & & & & & &
& \multirow{-2}{*}{\makecell{\textbf{Feature}\\\textbf{Family (\%)}}}
& \multirow{-2}{*}{\makecell{\textbf{Feature}\\\textbf{Category (\%)}}}
& Current
& Frontier
& \multirow{-2}{*}{\makecell{\textbf{Capability}\\\textbf{Coverage}}} \\
\midrule
R0 & 12 & 10 & 27 & 30 & 14 & 12 &  7 & 15 & 38.0 & 12.9 & 127 & 127 & 68.6 \\
R1 & 20 & 25 & 46 & 47 & 27 & 26 & 15 & 29 & 45.6 & 20.6 & 235 & 240 (+113) & 85.7 \\
R2 & 31 & 33 & 58 & 57 & 38 & 31 & 22 & 41 & 50.6 & 24.4 & 311 & 341 (+101) & 100.0 \\
R3 & 38 & 38 & 75 & 74 & 50 & 37 & 28 & 51 & 51.9 & 26.6 & 391 & 421 (+80) & 100.0 \\
R4 & 44 & 46 & 92 & 86 & 60 & 43 & 33 & 60 & 53.8 & 27.5 & 464 & 487 (+66) & 100.0 \\
\bottomrule
\end{tabular}%
}
\vspace{-2mm}
\end{table*}

\paragraph{State Grounding and Observability.}
Each requirement maps to an observable state field or artifact, a planned checkpoint, and one of three coverage states: covered, partially observable, or unverifiable. State paths must be supported by the application's state catalog or captured snapshots; proxy checks based only on file existence, filenames, setup state, or unavailable pre-action values are disallowed. Unsupported clauses remain explicit coverage gaps rather than unsound surrogates, avoiding selection bias against complex, harder-to-inspect applications that discarding such tasks would introduce.

\begin{figure*}[t]
\centering
\begin{subfigure}[t]{0.19\textwidth}
\centering
\includegraphics[width=\linewidth]{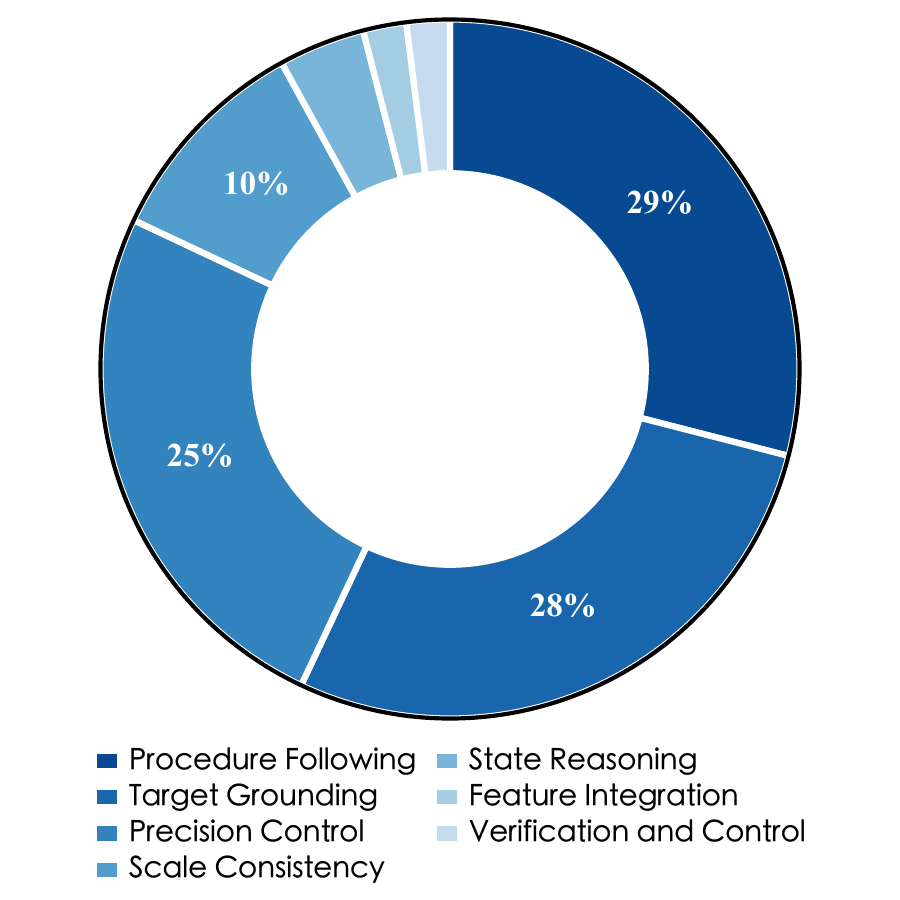}
\caption{Proportions of primary capability evaluated.}
\label{fig:primary-capability}
\end{subfigure}\hfill
\begin{subfigure}[t]{0.253333\textwidth}
\centering
\includegraphics[width=\linewidth]{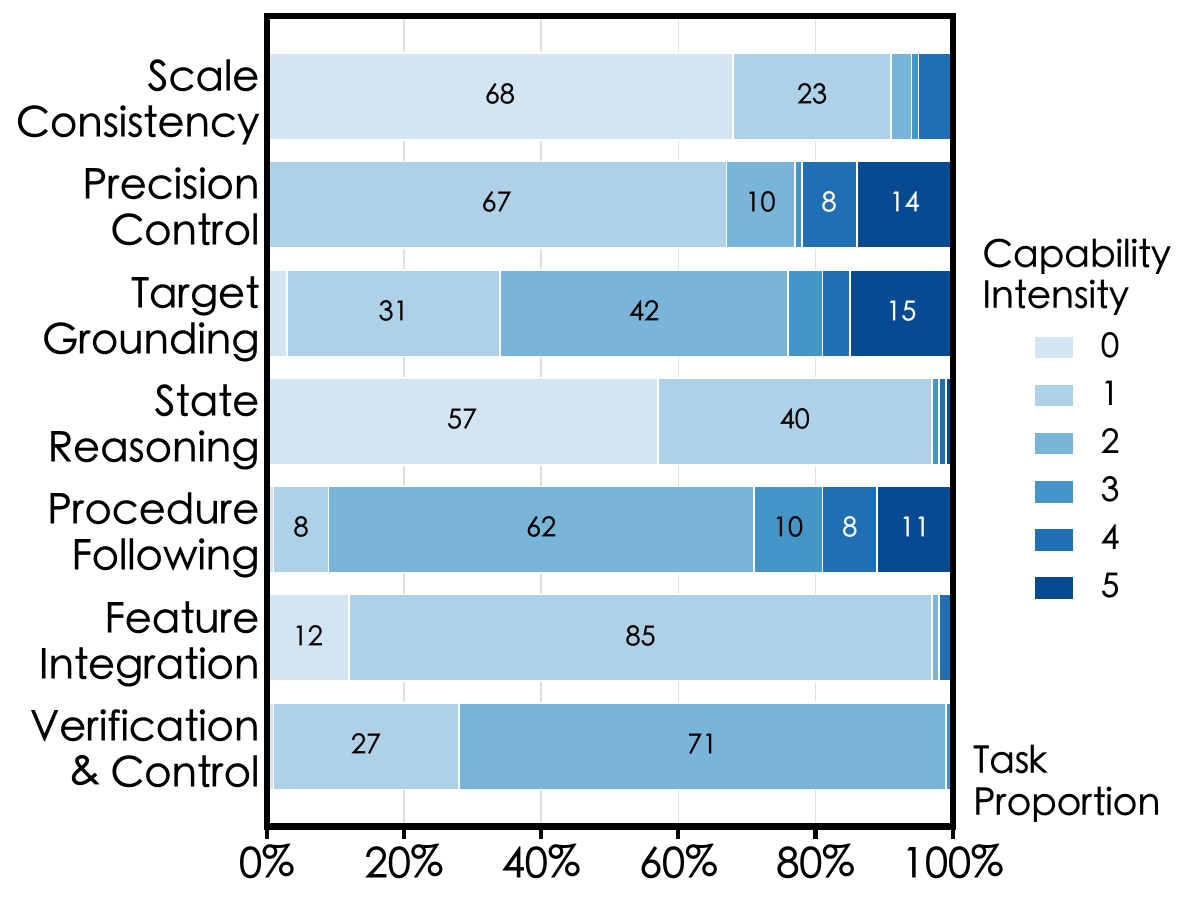}
\caption{Intensity distributions across capabilities.}
\label{fig:capability-intensity}
\end{subfigure}\hfill
\begin{subfigure}[t]{0.253333\textwidth}
\centering
\includegraphics[width=\linewidth]{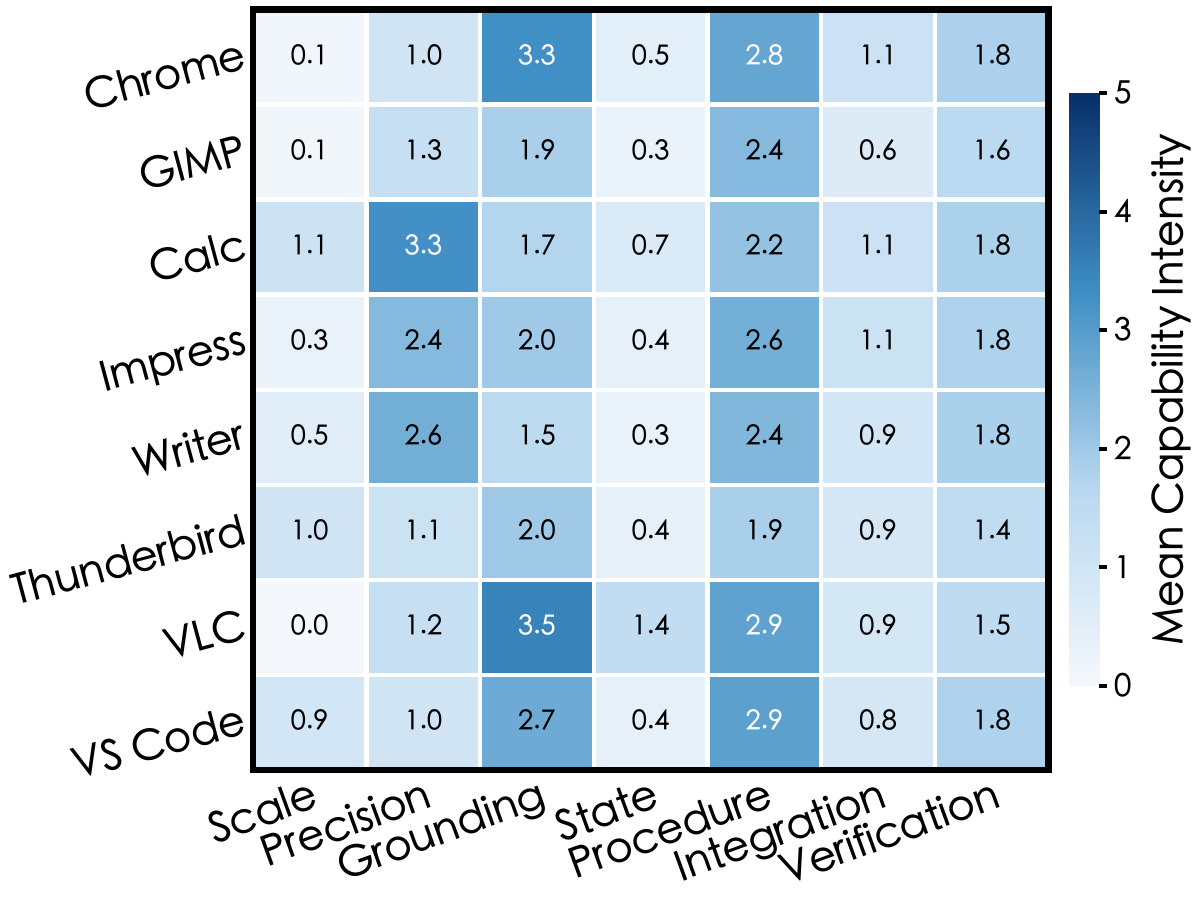}
\caption{Average capability demand across software applications.}
\label{fig:software-capability}
\end{subfigure}\hfill
\begin{subfigure}[t]{0.253333\textwidth}
\centering
\includegraphics[width=\linewidth]{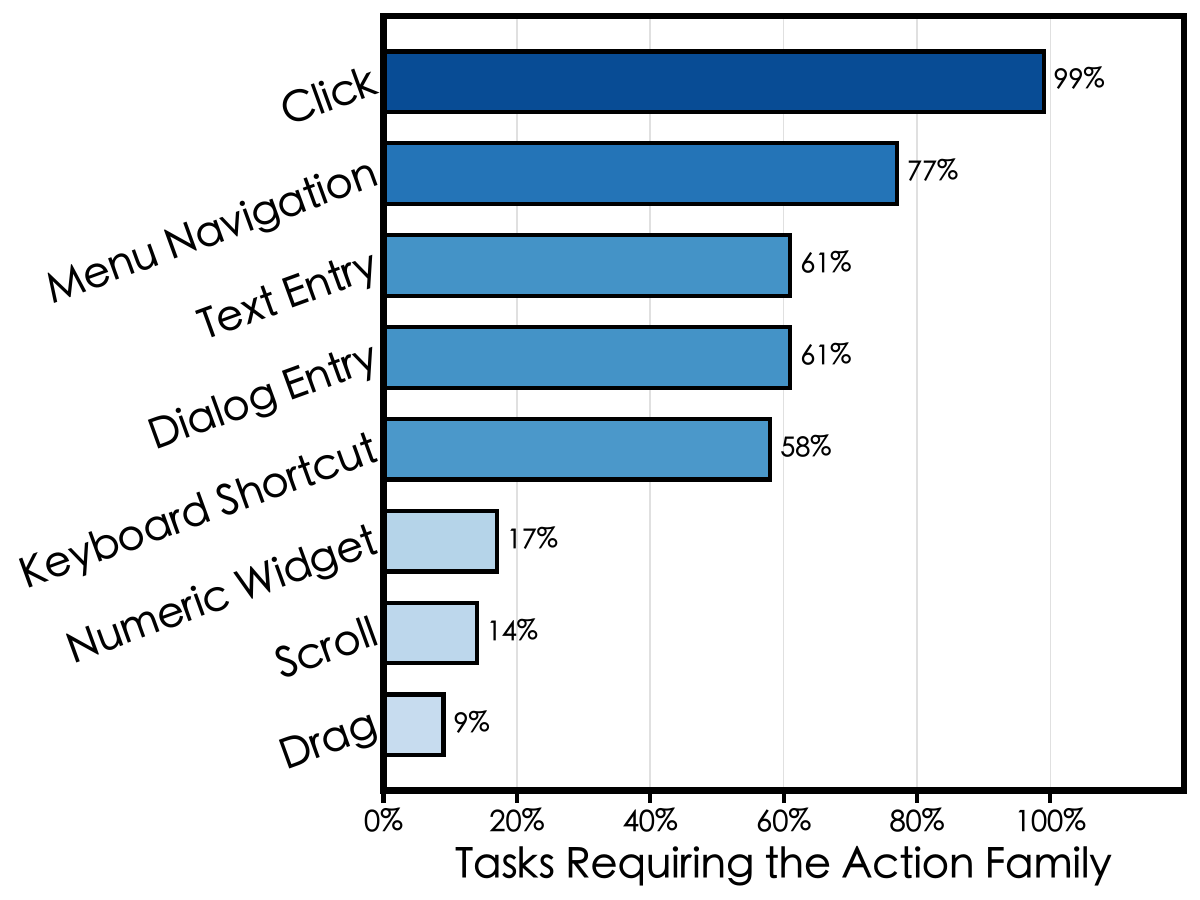}
\caption{Action-family exposure frequency among tasks.}
\label{fig:action-family-exposure}
\end{subfigure}
\vspace{-1mm}
\caption{Task-suite capability requirements
and action-family statistics.}
\label{fig:task-statistics}
\vspace{-2mm}
\end{figure*}

\begin{figure*}[t]
\centering
\begin{minipage}[t]{0.745\textwidth}
  \centering
  \begin{subfigure}[t]{0.32\linewidth}
    \centering
    \includegraphics[width=\linewidth]{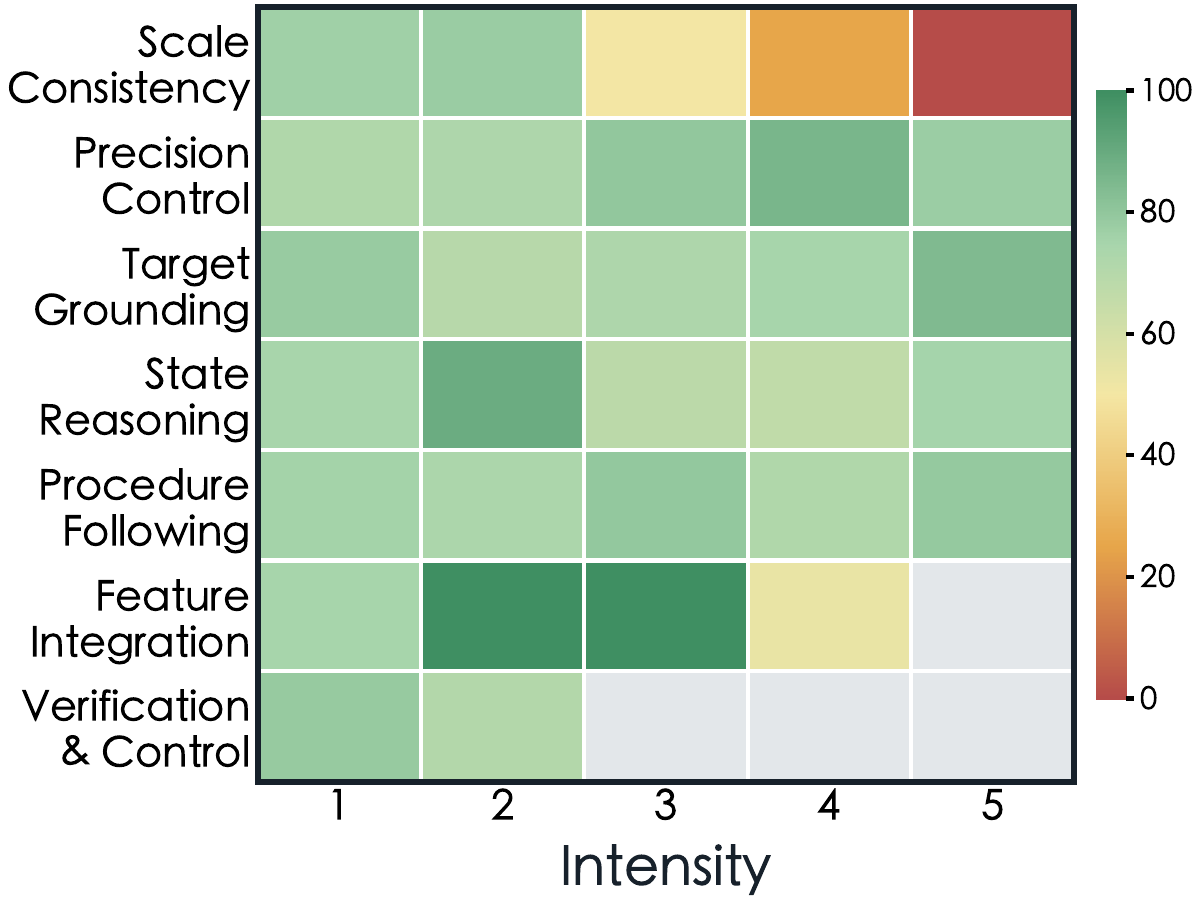}
    \caption{Capability intensity}
    \label{fig:gemini-r0-e3-intensity}
  \end{subfigure}\hfill
  \begin{subfigure}[t]{0.32\linewidth}
    \centering
    \includegraphics[width=\linewidth]{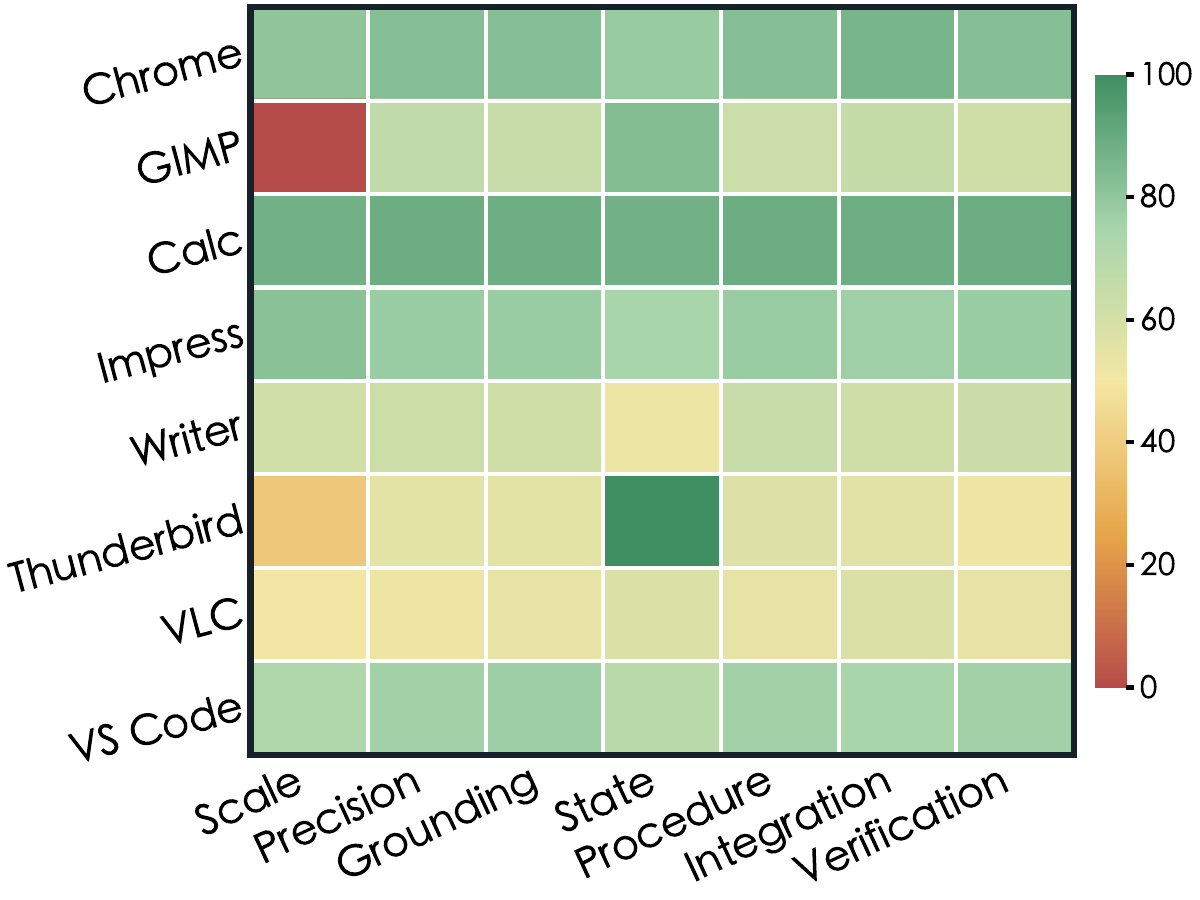}
    \caption{Software--capability}
    \label{fig:gemini-r0-e3-software-capability}
  \end{subfigure}\hfill
  \begin{subfigure}[t]{0.32\linewidth}
    \centering
    \includegraphics[width=\linewidth]{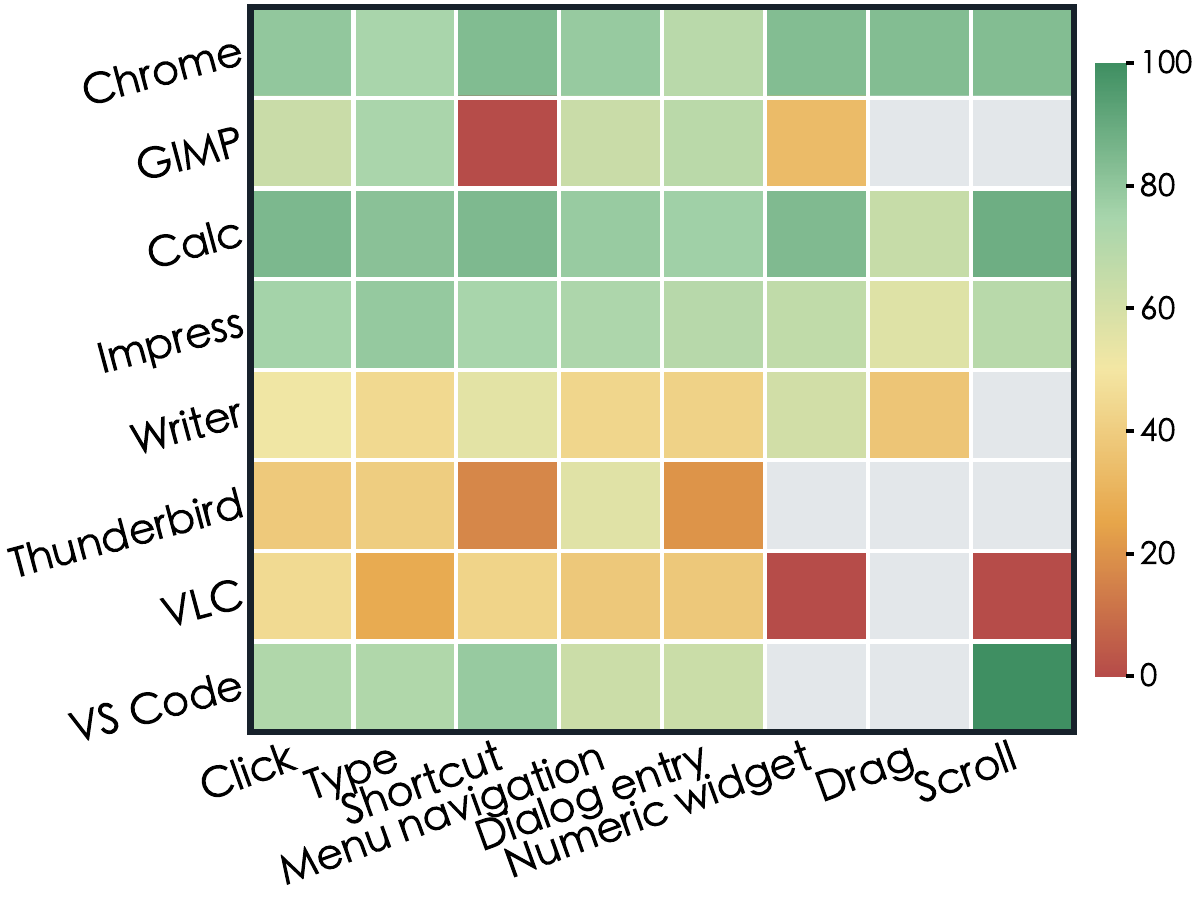}
    \caption{Software--action}
    \label{fig:gemini-r0-e3-software-action}
  \end{subfigure}
  \caption{Gemini diagnostic performance aggregated over $R0$--$R3$, with full matrices in Supplementary Figure~\ref{fig:diagnostic-matrices-all-agents}. Colors denote mean checkpoint scores; gray cells indicate unavailable evidence.}
  \label{fig:diagnose_heatmap_gemini}
\end{minipage}\hfill
\begin{minipage}[t]{0.24\textwidth}
  \centering
  \includegraphics[width=\linewidth]{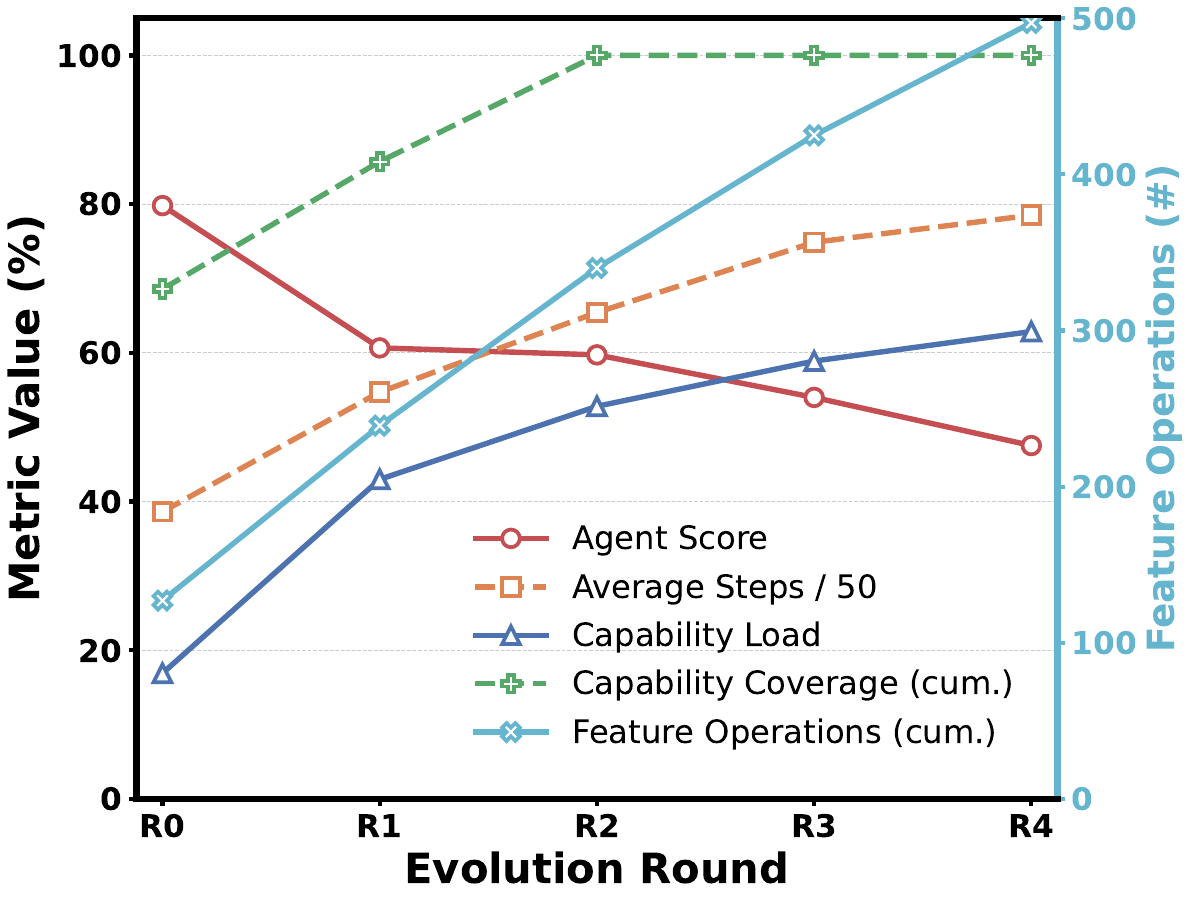}
  \vspace{-6mm}
  \caption{Round-wise benchmark evolution. Later rounds are harder and provide broader cumulative coverage.}
  \label{fig:benchmark-round-trends}
\end{minipage}
\vspace{-2mm}
\end{figure*}

\paragraph{Checkpoint-Lineage Transformation.} 
Given the source verifier, each existing checkpoint receives an explicit lineage operation: \textsc{Keep} for an unchanged obligation, \textsc{Modify} for a changed target or expected value, and \textsc{Remove} for an obligation absent from the evolved task; \textsc{Add} introduces checkpoints for newly required effects. 
The original OSWorld evaluator may provide semantic evidence about intended facts, tolerances, and conjunction structure, but the evolved task contract remains authoritative. Optional trajectories and checkpoint histories are used to confirm state paths and expose regressions, while an individual failed rollout is not treated as sufficient evidence of verifier error.

\paragraph{Executable Verifier Construction.} 
To construct a verifier, each software is equipped with a task-agnostic adapter that probes native state-export interfaces (e.g., debug protocols) to extract and normalize its snapshot into a shared \texttt{CanonicalState} schema. 
The transformed contract is compiled into distinct schema-grounded checkpoints, each of which checks one task effect and returns a specific expected-versus-actual reason. Every requirement must map to a checkpoint or be declared as an explicit observability gap, prohibiting shortcuts like unconditional passes or inference-based fallbacks. 
A partially observable task remains rollout-ready when at least one meaningful target effect is soundly verified and all omissions are declared.

\paragraph{Validation and Adversarial Audit.} 
Validation separates three static gates (i.e., syntax and imports, structural contract, and static schema), from two behavioral gates (i.e., PRE/PASS discrimination and checkpoint-wise negative fixtures). 
An information-isolated pass agent independently realizes an executable PASS state and audits checkpoint necessity and completeness, while a verifier agent audits whether that state genuinely satisfies the task without access to the pass implementation. Their adversarial cross-review exposes trivial checkpoints, missing obligations, and incorrect expected values. Revisions are adopted only when they preserve task semantics and do not reduce the best validated verifier score. 


\section{Experiments (\textit{More Results in Appendix})}

\paragraph{Experimental Setups.}
While CUA-SEE's substrate spans 20 applications (Appendix~\ref{sec:experimental_setup}), we track the eight core OSWorld apps under a 50-step budget for strict $R0$ alignment, reporting normalized scores, steps, and tokens.
For \emph{benchmark evolution}, 100 lineage-aligned task groups evolve from $R0$ through $R1$--$R4$ across five SOTA agents (2,500 paired observations controlling sample drift in total): Kimi-K2.6, Seed-2.1-Pro, and Gemini-3.5-Flash drive evolution, while Qwen-3.7-Plus and GPT-5.5 are strictly held out.
For \emph{diagnostic evolution}, independent boundary probes are generated over the seven $0$-$5$ capability dimensions (Section~\ref{sec:diagnostic_task_evolution}) for Gemini, Kimi, and Seed.
$R0$ uses native OSWorld tasks and verifiers; evolved rounds use CanonicalState verifiers.

\paragraph{Scoring and Metrics.}
Benchmark scores average normalized episode scores to retain partial credit. Diagnostic outcomes separate valid pass ($[0.99,1]$), partial ($(0.01,0.99)$), and fail ($[0,0.01]$) runs from 50-step censored and confounded/invalid runs; matrix values report task-weighted checkpoint means over eligible clean episodes rather than full-task pass rates or averages of round means (gray cells denote unavailable evidence).

\subsection{Evaluations on Evolved Benchmarks}


\paragraph{Higher Task Difficulty.}
As summarized in Table \ref{tab:benchmark-evolution-results}, we conclude the following:
(1) The evolved benchmark becomes substantially more demanding.
Across all five agents, the macro-average score decreases from 60.6 in $R1$ to 47.5 in $R4$, while the average trajectory length increases from 27.4 to 39.3 steps.
(2)
Kimi, Qwen, and GPT obtain slightly higher scores in $R2$ than in $R1$; nevertheless, their trajectory lengths increase by 9.5, 4.6, and 5.8 steps, respectively. This indicates that later tasks generally require longer interaction chains, while newly introduced features may align better with the capabilities of particular agents. 
(3) Difficulty growth is uneven across applications. Chrome remains comparatively easy in later rounds, whereas GIMP, Calc, and Writer exhibit larger score reductions.

\begin{figure*}[t]
\centering
\begin{subfigure}[t]{0.24\textwidth}
\centering
\includegraphics[width=\linewidth]{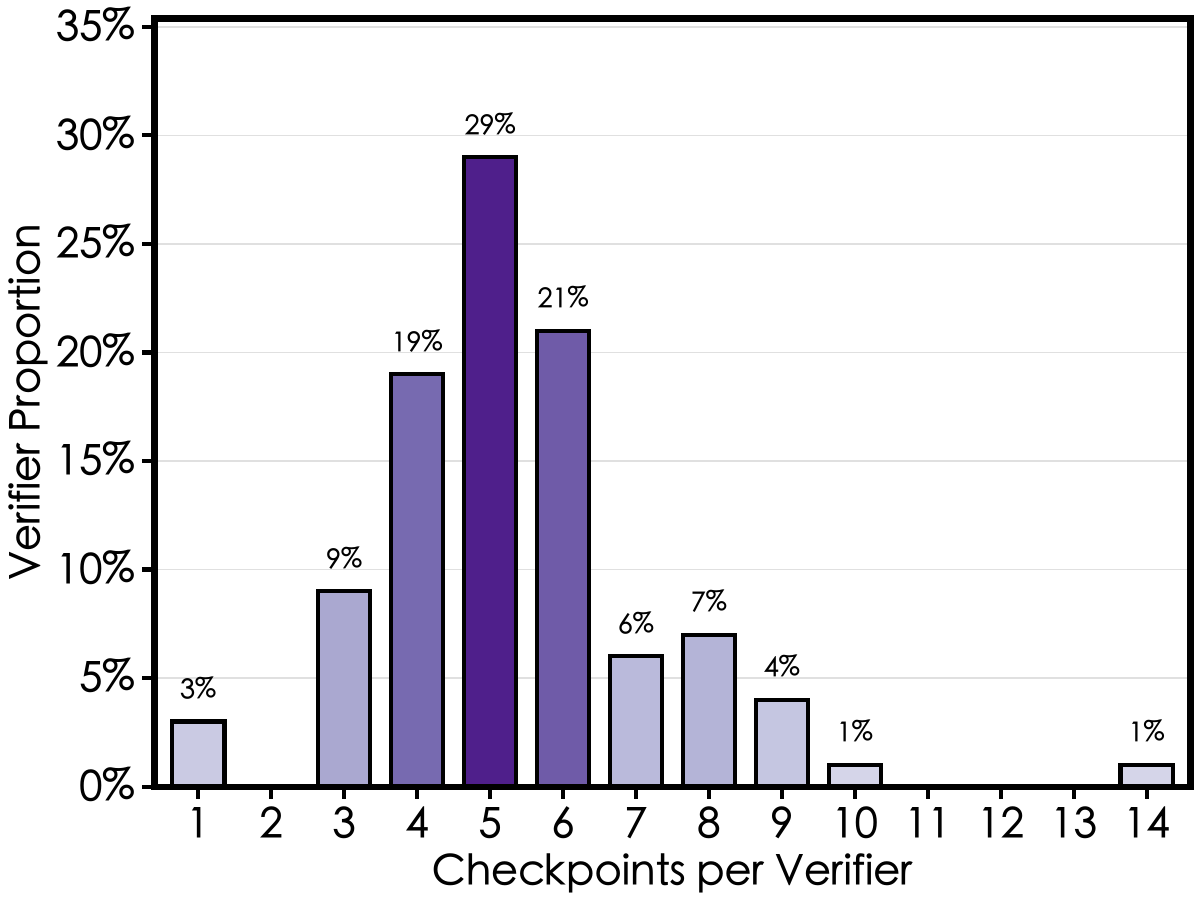}
\caption{Checkpoint distribution}
\label{fig:verifier-kimi-distribution}
\end{subfigure}\hfill
\begin{subfigure}[t]{0.24\textwidth}
\centering
\includegraphics[width=\linewidth]{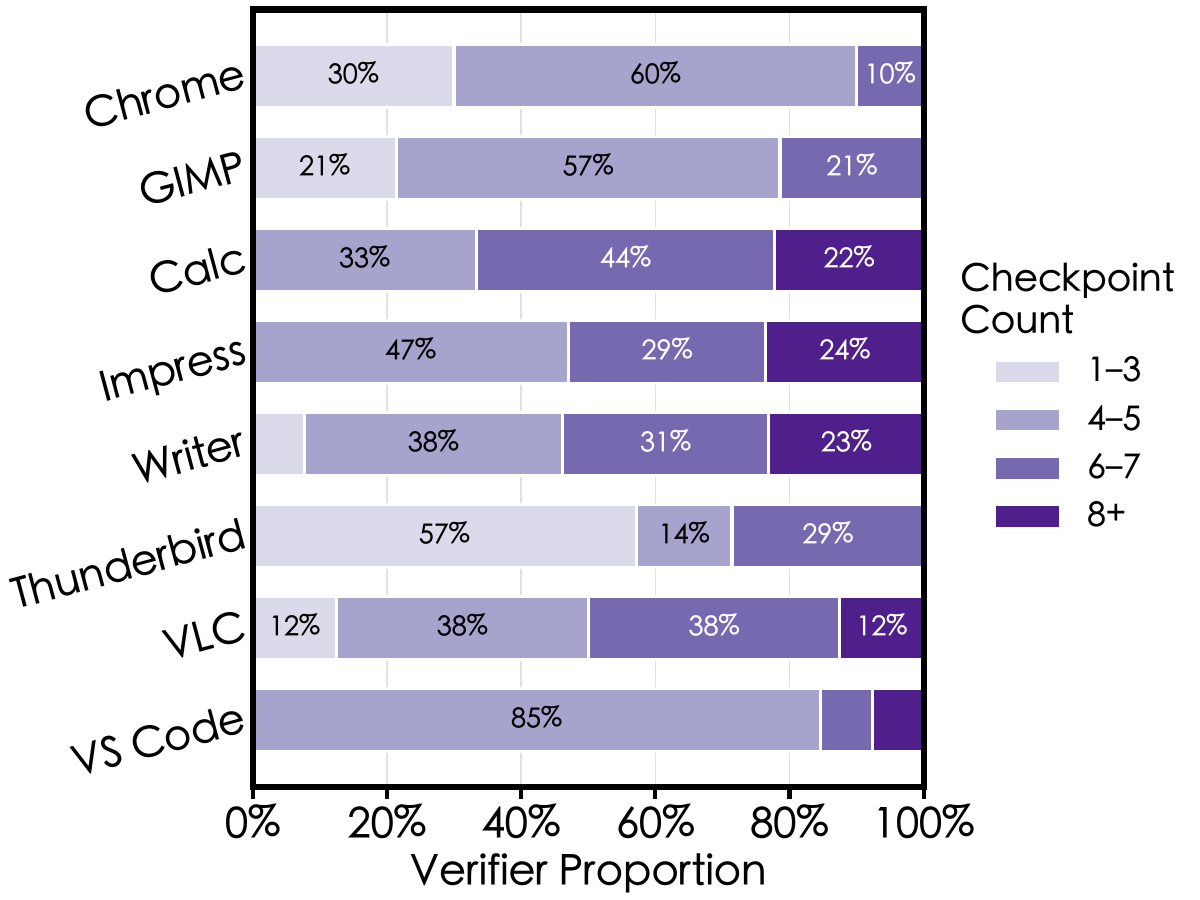}
\caption{Software profile}
\label{fig:verifier-kimi-profile}
\end{subfigure}\hfill
\begin{subfigure}[t]{0.24\textwidth}
\centering
\includegraphics[width=\linewidth]{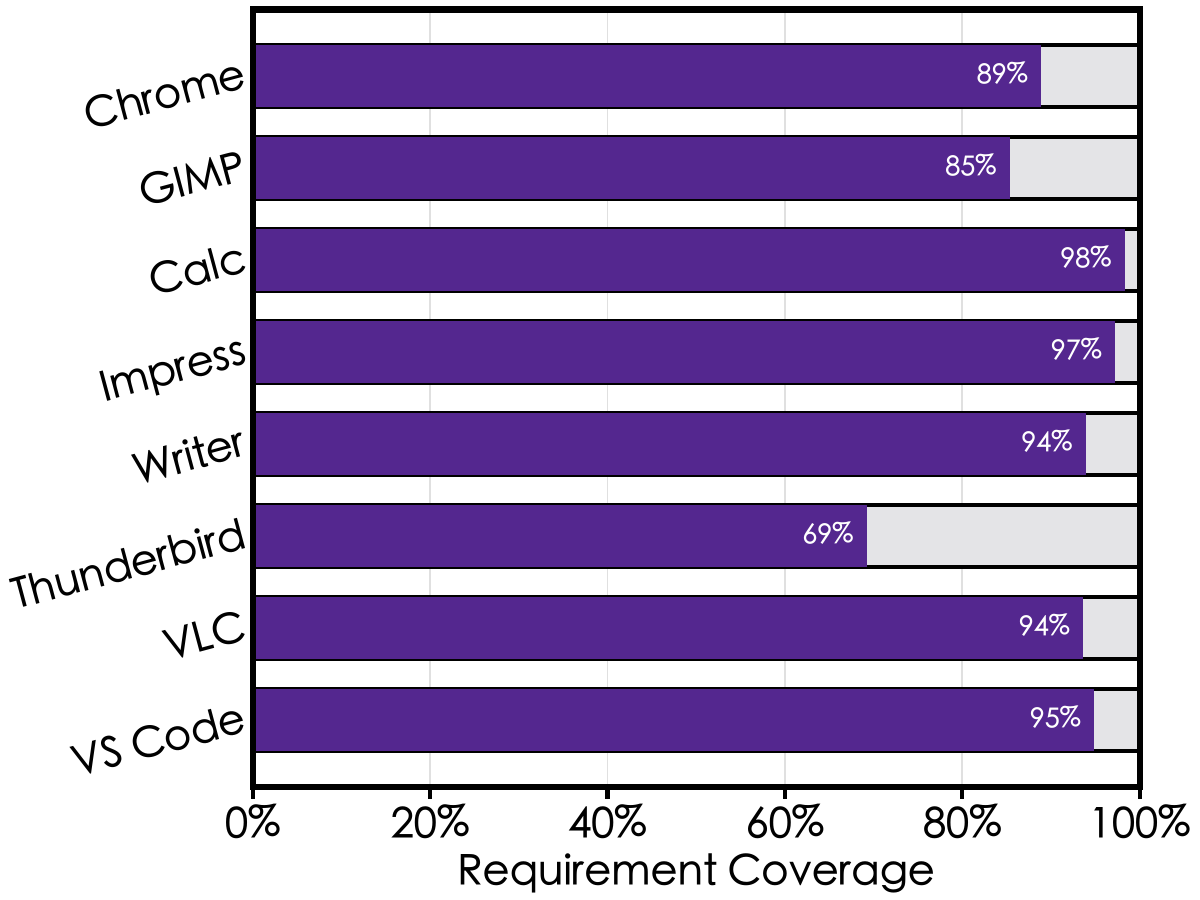}
\caption{Requirement coverage}
\label{fig:verifier-kimi-coverage}
\end{subfigure}\hfill
\begin{subfigure}[t]{0.24\textwidth}
\centering
\includegraphics[width=\linewidth]{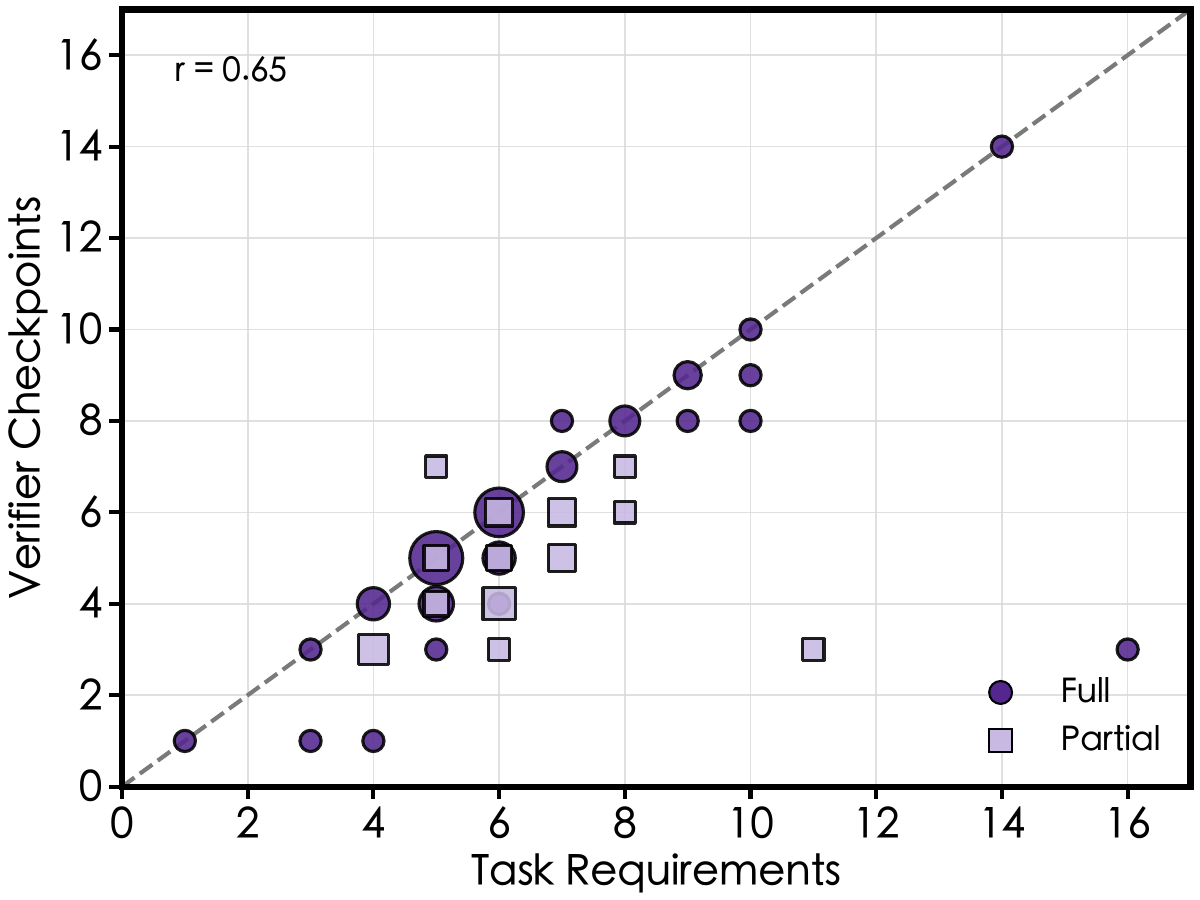}
\caption{Requirement alignment}
\label{fig:verifier-kimi-alignment}
\end{subfigure}
\vspace{-1mm}
\caption{Verifier construction statistics on the Kimi diagnostic benchmark across checkpoint-count distribution, software-specific checkpoint profile, atomic requirement coverage, and checkpoint--requirement alignment (Gemini and Seed in Supplementary Figure~\ref{fig:verifier-statistics-all-agents}).}
\label{fig:verifier-coverage-alignment}
\vspace{-2mm}
\end{figure*}

\begin{figure}[t]
\centering
\begin{minipage}[t]{0.23\textwidth}
\centering
\includegraphics[width=1\linewidth]{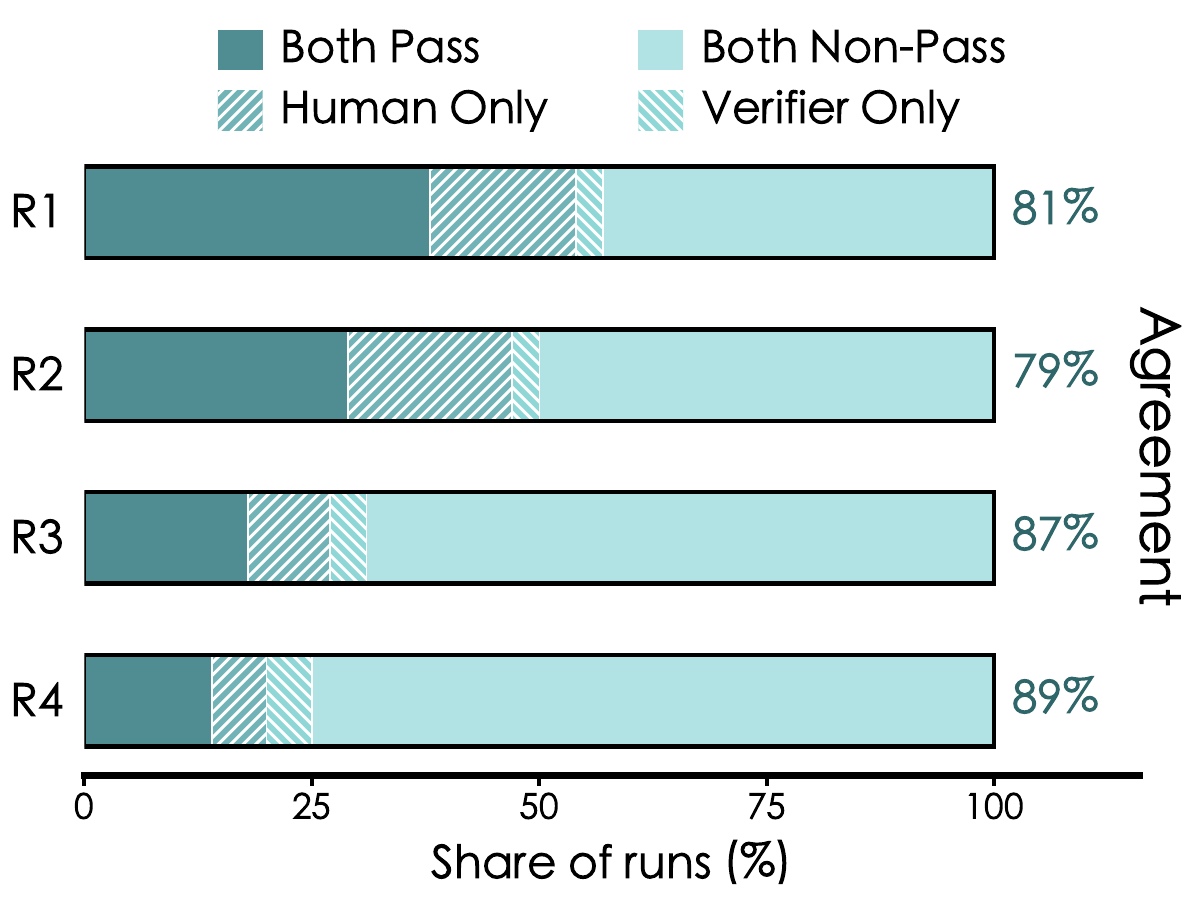}
\vspace{-7mm}
\caption{Binary agreement between human judgments and verifier outcomes.}
\label{fig:binary_judgement}
\end{minipage}\hfill
\begin{minipage}[t]{0.23\textwidth}
\centering
\includegraphics[width=1\linewidth]{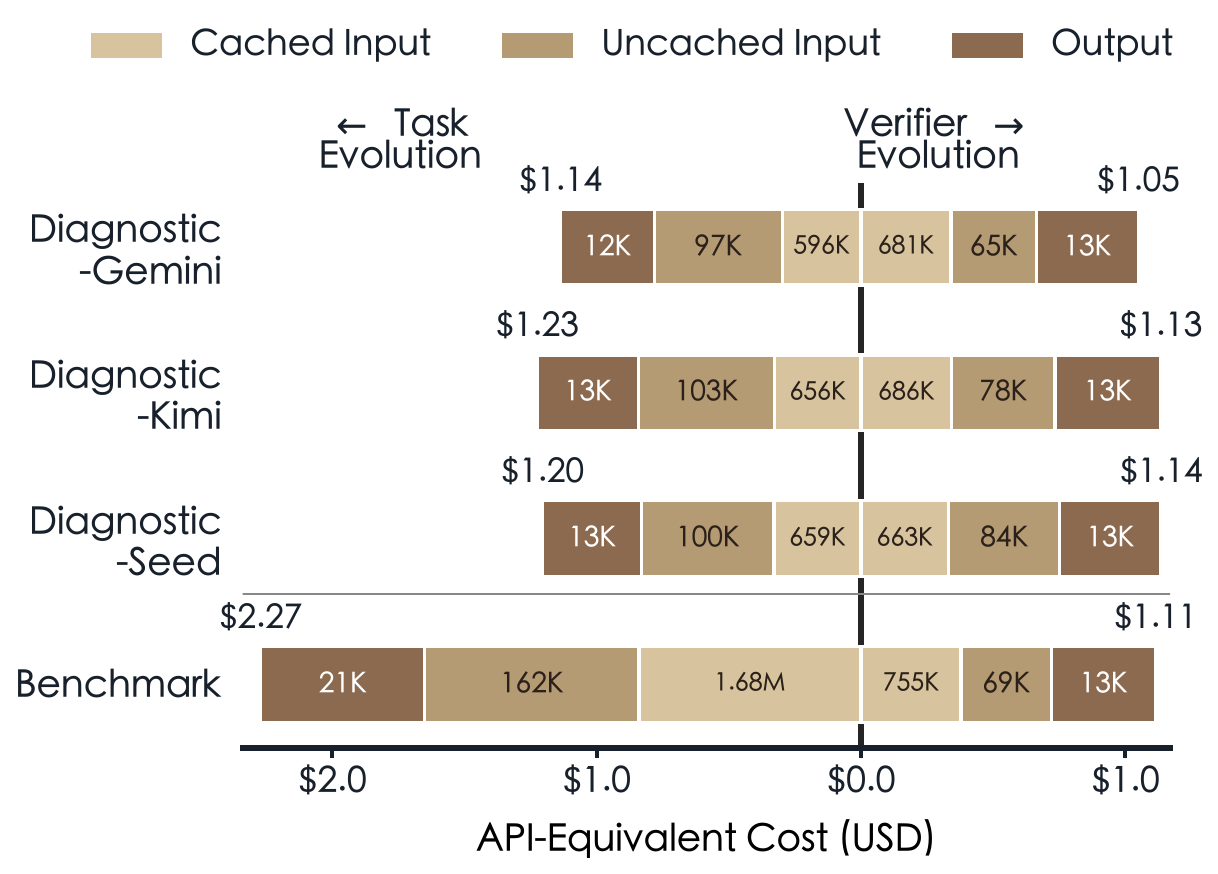}
\vspace{-7mm}
\caption{Token composition and API-equivalent cost per successful artifact.
}
\label{fig:efficiency_token}
\end{minipage}
\vspace{-1mm}
\end{figure}

\paragraph{Broader Feature Coverage.}
Feature coverage is organized hierarchically from broad feature families, through finer-grained feature categories, to exact executable feature operations within each software.
(1) The results in Table \ref{tab:feature-coverage} also demonstrate a consistent and hierarchical expansion of functional coverage across successive evolution rounds, with steady growth in feature families and categories and a $3.65\times$ increase in exact feature operations by $R4$.
(2) These gains appear across all eight applications, indicating that evolution broadens the benchmark as a whole rather than concentrating on a few software domains.
Overall, these results confirm that iterative benchmark evolution systematically broadens the functional and operational scope of the benchmark rather than merely rephrasing existing tasks.

\paragraph{Multi-Faceted Task Statistics.}
Beyond difficulty and coverage metrics, the evolved tasks exhibit a rich, multi-faceted structure, with detailed statistics shown in Figure~\ref{fig:task-statistics}.
(1) The tasks exhibit compound capability demands, with procedure following and target grounding frequently required at moderate-to-high intensities.
(2) Capability profiles vary substantially across applications—for example, Calc emphasizes precision control, whereas Chrome and VLC place greater demands on target grounding—demonstrating software-specific rather than templated difficulty.
(3) Finally, the suite covers diverse interaction families, spanning clicks, text, keyboard shortcuts, and long-tail operations such as scrolling and dragging.

\subsection{Evolved Diagnoses of Specific Agents}


\paragraph{Effective and Discriminative Diagnosis.}
(1) Figure \ref{fig:kimi_diagnostic_boundary} reveals that our Diagnostic Evolution continually mines the agent's pass/fail boundary, thereby shifting the original saturated benchmark from near-universal success toward a more informative regime where nuanced diagnosis becomes possible.
(2) Our method diagnoses heterogeneous capabilities and distinct software-specific strengths that are typically obscured by aggregate scores. As visualized in Figure \ref{fig:diagnose_heatmap_gemini} and Supplementary Figures \ref{fig:diagnose-seed-software-capability} and \ref{fig:diagnose-kimi-software-capability}, this analysis reveals unique performance signatures for each agent across various diagnostic matrices, such as capability-intensity and software-feature.
(3) Successive rounds expand coverage across capability, software, function, and action dimensions. Each round consistently introduces 57–73 new feature leaves, ensuring that our diagnosis grows in breadth and is not merely resampling similar challenges.

\paragraph{Agent-Specific Diagnostic Profiles.}
Based on the diagnosis, we construct specific agent profiles.
(1) Capability: Gemini is strong in precision control and target grounding, but high-scale consistency remains challenging.
(2) Operation: Kimi is reliable on clicking, typing, and navigation, while Seed is more sensitive to dragging, scrolling, and state-dependent controls.
(3) Feature: Kimi shows the most consistent feature-level performance, whereas Seed varies more across document and media features.
(4) Software: All three agents are more stable in Chrome and VS Code, while GIMP and Thunderbird expose clearer weaknesses.

\subsection{Validation of Evolved Verifiers}

\paragraph{Trajectory-Level Verifier Agreement.}
We manually assess convergent validity on 400 Kimi trajectories across four evolution rounds in Figure \ref{fig:binary_judgement}.
Human judgments and checkpoint verifiers agree on 84.0\% of runs (Cohen's $\kappa=0.640$), with agreement increasing from 81\% to 89\%. Continuous scores also show moderate concordance.
Audited disagreements mainly arise from incomplete adapter state, partial verifier coverage, and state-field mismatches, highlighting the remaining observability limitations.

\paragraph{Verifier Statistics.}
Figure~\ref{fig:verifier-coverage-alignment} and Supplementary Figure~\ref{fig:verifier-statistics-all-agents} confirm that evolved verifiers stay tightly aligned with growing task contracts rather than collapsing into minimal checks. Across 962 tasks (4.82 checkpoints/task), atomic requirement coverage reaches 92.5\% overall—exceeding 85\% in nearly all apps, with lower coverage isolated to native-adapter limits (e.g., Thunderbird). Moreover, checkpoint counts track requirement counts along the $y\!=\!x$ diagonal ($r\!=\!0.55\text{--}0.65$), with \textsc{Partial} tasks still grounding 3--7 checkpoints; the independent trajectory audit (Figure~\ref{fig:binary_judgement}) further validates behavioral precision.

\subsection{Evaluation Reliability and Evolution Efficiency}

\paragraph{Evaluation Reliability.}
Under live-VM budget and native observability limits, cross-round and cross-agent comparisons remain fair and discriminative:
(1)~Out-of-cohort generalization: held-out Qwen ($-26.9$) and GPT ($-24.0$) match the cohort's difficulty progression without driving evolution.
(2)~Calibrated co-evolution: \textsc{Keep}/\textsc{Modify} inheritance, isolated PASS-state solvability gates, rising human agreement ($81\%$ to $89\%$), verifier-independent $2\times$ step, and $2\text{--}3.5\times$ token growth rule out verifier drift.
(3)~Partial-verifier symmetry: retaining sound \textsc{Partial} verifiers preserves hard-to-inspect apps while applying identical checks across all agents without artificial score deflation.
(4)~Structured discrimination: while low $R4$ pass rates ($13\text{--}19\%$) narrow the macro spread, continuous scores lift boundary informativeness ($+20.8$~pp to $69.7$) and expose $20\text{--}37$~pp per-app rank reversals masked by scalar means.

\paragraph{Evolution Efficiency.}
In Figure \ref{fig:efficiency_token}, we measure the token usage and API-equivalent cost of successful task and verifier evolution runs, aggregated across rounds for three diagnostic agents and the multi-agent benchmark. Costs are estimated from cached input, uncached input, and output tokens using GPT-5.5 API rates. Verifier evolution exhibits stable costs across agents, while benchmark task evolution is more expensive due to its larger shared context, indicating that context construction—not verifier generation—is the primary efficiency bottleneck.

\section{Conclusion}
We present CUA-SEE, a closed-loop evaluation framework that co-evolves tasks and executable verifiers from computer-use agent trajectories. Benchmark evolution expands difficulty and coverage, diagnostic evolution probes agent-conditional capability boundaries, and verifier evolution grounds changing requirements in audited checkpoints. Experiments across agents, applications, and rounds show that adaptive evaluation remains informative beyond static benchmarks while producing attributable capability evidence. These results establish evolving evaluation as a practical counterpart to increasingly capable, self-evolving agents.

\section*{LLM Usage Statement}
Large language models were used during this project for manuscript language polishing, figure and layout drafting, and AI-assisted software engineering (``vibe coding'') of evaluation, data-processing, and visualization scripts. In addition, LLM and VLM inference APIs serve as core runtime components within the CUA-SEE synthesis pipeline and the evaluated computer-use agent rollouts, as detailed in Section~\ref{sec:method} and Appendix~\ref{sec:implementation}. All methodological designs, code implementations, experimental results, and scientific claims were inspected and verified by the authors, who take full responsibility for the contents of this paper.

{\small
\bibliographystyle{ieeenat_fullname}
\bibliography{main}
}

\clearpage
\setcounter{page}{1}
\maketitlesupplementary
\appendix

\section{Core Taxonomies, Capability Rubrics, and System Boundary}
\label{sec:appendix_concepts}
This section defines the foundational taxonomies, ordinal capability rubrics, and invariance boundaries that anchor CUA-SEE across evolution rounds.

\begin{figure}[H]
\centering
\includegraphics[width=0.8\linewidth]{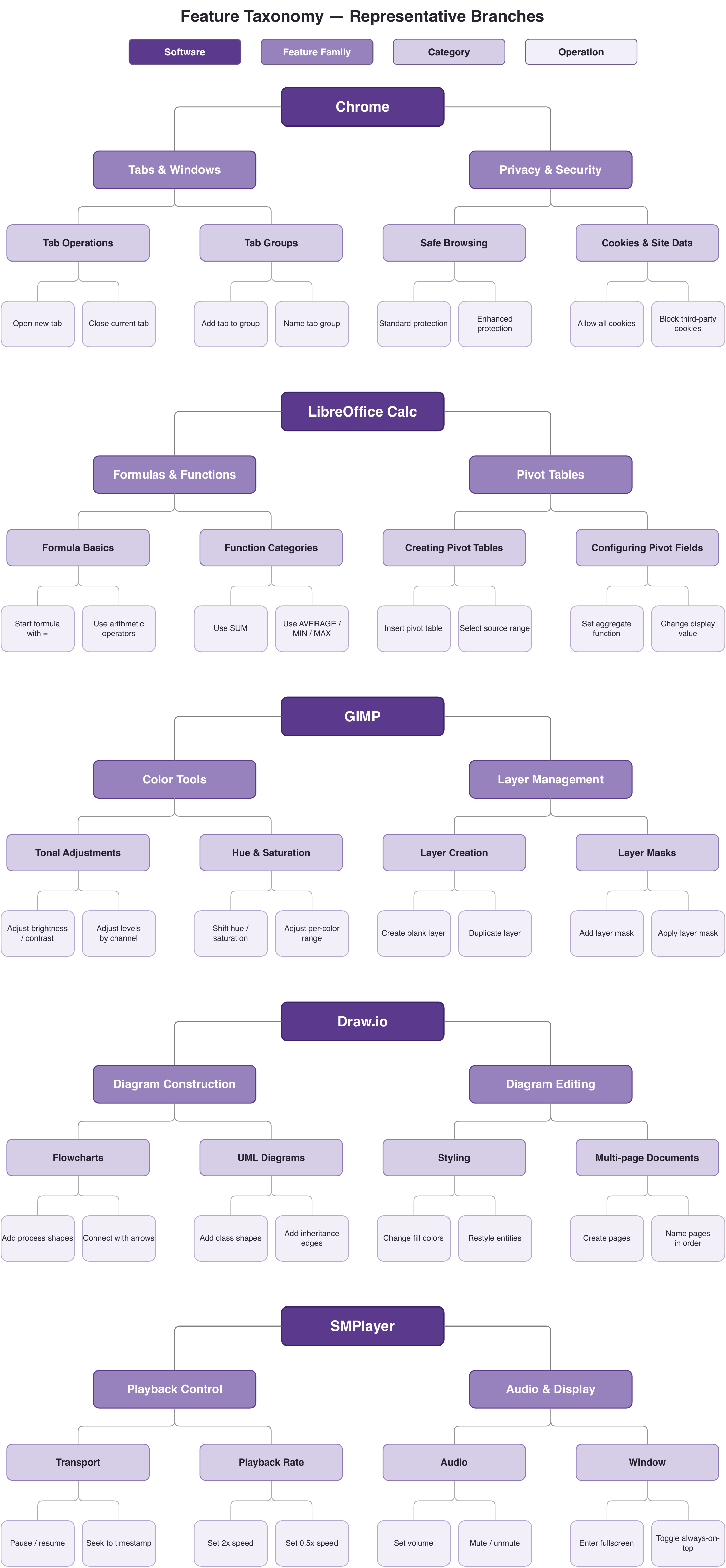}
\caption{Feature taxonomy hierarchy tree. Software functionality is organized into four levels—feature family, category, and concrete operation beneath each application—with representative branches shown for diverse software. The feature-, family-, category-, and operation-level coverage metrics reported in the main text are computed over this hierarchy.}
\label{fig:feature_tree}
\end{figure}

\paragraph{Feature Taxonomy.}
As illustrated in Figure \ref{fig:feature_tree}, we organize the feature space as a four-level hierarchy spanning software, feature families, functional categories, and concrete operations. The representative branches illustrate how application-level functionality is decomposed into directly executable and verifiable actions across browsers, productivity tools, creative applications, diagram editors, and media players. This structured taxonomy provides a consistent basis for measuring feature coverage, identifying underexplored functionality, and guiding task evolution toward broader and more diverse interaction capabilities.

\paragraph{Capability Dimensions and Intensity Rubric.}
Independent of software features, task demand is parameterized over seven orthogonal capability dimensions scored on an ordinal $0$--$5$ rubric (Table~\ref{tab:capability-rubric}). Scores are anchored in objective task factors—such as target count, value coupling, distractor similarity, state-derivation depth, and dependency chain length—derived strictly from the instruction, initial state, and verifier contract rather than predicted agent performance. To keep capability probes attributable, nuisance confounds (\emph{step-budget ratio, context load, navigation hiddenness,} and \emph{verifier risk}) are tracked separately and bounded during candidate selection.

\paragraph{Action-Family Exposure Taxonomy.}
To characterize low-level GUI interaction demands without conflating them with cognitive capabilities, each evolved profile records required exposure over eight canonical action families: \emph{click}, \emph{text entry}, \emph{keyboard shortcut}, \emph{menu navigation}, \emph{dialog entry}, \emph{numeric-widget manipulation}, \emph{drag}, and \emph{scroll}. Action annotations represent task-required interaction exposure rather than trajectory-inferred behavior, ensuring that software--action profiles remain well-defined even when an agent fails early or chooses an alternative UI path.

\begin{table*}[t]
\caption{Seven calibrated capability dimensions, grounded task factors, and condensed $0$--$5$ ordinal intensity anchors.}
\label{tab:capability-rubric}
\centering
\scriptsize
\setlength{\tabcolsep}{5pt}
\renewcommand{\arraystretch}{1.22}
\begin{tabularx}{\textwidth}{@{}
  >{\raggedright\arraybackslash}p{0.145\textwidth}
  >{\raggedright\arraybackslash}p{0.175\textwidth}
  >{\raggedright\arraybackslash\columncolor{evoBlue!3}}X
  >{\raggedright\arraybackslash\columncolor{evoTeal!4}}X
  >{\raggedright\arraybackslash\columncolor{evoOrange!5}}X@{}}
\toprule
\rowcolor{evoNavy!7}
\textbf{\textcolor{evoInk}{Dimension}}
& \textbf{\textcolor{evoInk}{Grounded Task Factors}}
& \textbf{\textcolor{evoBlue}{Level 0--1 (Minimal)}}
& \textbf{\textcolor{evoTeal}{Level 2--3 (Moderate)}}
& \textbf{\textcolor{evoOrange!90!black}{Level 4--5 (High / Dominant)}} \\
\midrule
\textbf{\textcolor{evoNavy}{Scale Consistency}} & Target count, distinct values, homogeneity & \textbf{0:} single target; \textbf{1:} two targets via one bulk action & \textbf{2:} 2--3 manual targets; \textbf{3:} 4--6 targets or distinct values & \textbf{4:} 7--12 targets across views; \textbf{5:} $>12$ targets within step budget \\
\addlinespace[2pt]
\textbf{\textcolor{evoNavy}{Precision Control}} & Value count/type, tolerance, unit conversion, coupling & \textbf{0:} no exact value; \textbf{1:} single categorical/default choice & \textbf{2:} 1--2 simple exact values; \textbf{3:} several exact values or strict format & \textbf{4:} coupled values, formulas, or unit conversion; \textbf{5:} coupled heterogeneous precision \\
\addlinespace[2pt]
\textbf{\textcolor{evoNavy}{Target Grounding}} & Candidate count, distractor similarity, visibility & \textbf{0:} explicit visible target; \textbf{1:} clear label without distractors & \textbf{2:} 2--4 distinct candidates; \textbf{3:} similar candidates by metadata/content & \textbf{4:} partly hidden targets via computed property; \textbf{5:} multi-stage cross-view grounding \\
\addlinespace[2pt]
\textbf{\textcolor{evoNavy}{State Reasoning}} & State variables read, derivation steps, branching & \textbf{0:} state-independent; \textbf{1:} read one directly visible state value & \textbf{2:} single condition/transform; \textbf{3:} 1--2 derivation steps or 2 branches & \textbf{4:} multi-variable or multi-branch derivation; \textbf{5:} iterative state updates/derivations \\
\addlinespace[2pt]
\textbf{\textcolor{evoNavy}{Procedure Following}} & Subgoal count, ordering strictness, irreversibility & \textbf{0:} one operation; \textbf{1:} two independent obvious operations & \textbf{2:} 3--4 weakly ordered subgoals; \textbf{3:} 4--6 dependent subgoals & \textbf{4:} long strict chain, irreversible step, or exact export; \textbf{5:} guarded multi-stage workflow \\
\addlinespace[2pt]
\textbf{\textcolor{evoNavy}{Feature Integration}} & Feature count, UI distance, cross-object data transfer & \textbf{0:} single feature area; \textbf{1:} two controls in one feature & \textbf{2:} two adjacent feature areas; \textbf{3:} three areas or explicit data transfer & \textbf{4:} distant features with cross-object linkage; \textbf{5:} multi-artifact/cross-app integration \\
\addlinespace[2pt]
\textbf{\textcolor{evoNavy}{Verification \& Control}} & Invariant count, inspect-and-correct, side effects & \textbf{0:} no invariant/constraint; \textbf{1:} basic completion check & \textbf{2:} one explicit preservation rule; \textbf{3:} multiple invariants or inspect-fix & \textbf{4:} conditional multi-object correction \& side-effect control; \textbf{5:} iterative recovery burden \\
\bottomrule
\end{tabularx}
\end{table*}

\paragraph{Fixed Substrate vs.\ Self-Evolving Artifacts.}
To clarify what ``self-evolving'' entails and why cross-round comparison remains stable, Table~\ref{tab:fixed-vs-evolving} separates CUA-SEE's invariant infrastructure from its round-wise generated artifacts. Holding the VM substrate, native state adapters, taxonomies, scoring rubrics, and validation gates fixed ensures that changes across rounds reflect genuine growth in task scope, boundary resolution, and contract coverage rather than shifting measurement standards.

\begin{table*}[t]
\caption{Lifecycle boundary separating fixed evaluation infrastructure from round-wise self-evolving artifacts in CUA-SEE.}
\label{tab:fixed-vs-evolving}
\centering
\footnotesize
\setlength{\tabcolsep}{6.5pt}
\renewcommand{\arraystretch}{1.24}
\begin{tabularx}{\textwidth}{@{}
  >{\raggedright\arraybackslash}p{0.19\textwidth}
  >{\raggedright\arraybackslash\columncolor{evoBlue!4}}X
  >{\raggedright\arraybackslash\columncolor{evoTeal!5}}X@{}}
\toprule
\rowcolor{evoNavy!7}
\textbf{\textcolor{evoInk}{System Layer}}
& \textbf{\textcolor{evoBlue}{Fixed Across All Rounds} \normalfont\scriptsize\textcolor{evoMuted}{(Invariant Substrate)}}
& \textbf{\textcolor{evoTeal}{Evolved Per Round} \normalfont\scriptsize\textcolor{evoMuted}{(Self-Evolving Artifacts)}} \\
\midrule
\textbf{\textcolor{evoNavy}{Environment \& State}}\newline
{\scriptsize\color{evoMuted}Execution \& snapshot runtime}
& \textbf{Ubuntu VM \& Adapters:} Isolated VM images, 20 app binaries, task-agnostic \texttt{CanonicalState} adapters, and 50-step rollout controller.
& \textbf{Task Instantiations:} Lineage-derived initial states, input files/fixtures, and evolved parameterizations per task instance. \\
\cmidrule(lr){1-3}
\textbf{\textcolor{evoNavy}{Taxonomies \& Rubrics}}\newline
{\scriptsize\color{evoMuted}Demand \& coverage ontology}
& \textbf{Calibrated Ontologies:} Four-level software--feature--category--operation tree, 7D capability $0$--$5$ rubric, 8 action families, and confound limits.
& \textbf{Dynamic Planning State:} Shared frontier coverage registry, per-task 7D demand profiles, and agent-conditional diagnostic dispatches. \\
\cmidrule(lr){1-3}
\textbf{\textcolor{evoNavy}{Measurement \& Gates}}\newline
{\scriptsize\color{evoMuted}Verification \& validity audit}
& \textbf{Audit Protocol:} Static contract/schema gates, isolated \texttt{PASS}-agent vs.\ verifier cross-audit, and validity classes (\emph{clean/censored/confounded}).
& \textbf{Executable Contracts:} Atomic requirement decompositions, \textsc{Keep/Modify/Remove/Add} checkpoints, and empirical boundary maps. \\
\bottomrule
\end{tabularx}
\end{table*}

\section{Experiment Details and Prompts}

\subsection{Experimental Setup}
\label{sec:experimental_setup}
\paragraph{Evaluation scope.}
We evaluate benchmark evolution and diagnostic evolution on realistic Ubuntu desktop environments derived from OSWorld. As shown in Figure~\ref{fig:osworld_vs_cua_see} and Figure~\ref{fig:main}, CUA-SEE's shared measurement substrate expands the eight core OSWorld applications (Chrome, GIMP, LibreOffice Calc, LibreOffice Impress, LibreOffice Writer, Thunderbird, VLC, and Visual Studio Code) with twelve additional native-\texttt{CanonicalState}-instrumented desktop environments—Blender, VSCodium, JupyterLab, Krita, Draw.io, SMPlayer, OBS Studio, ONLYOFFICE, OpenScreen, ComfyUI, Zotero, and Brave—bringing total software breadth to 20 applications across browsers, office suites, IDEs, creative/3D studios, diagramming, media production, and research workflows. To ensure strict task-lineage correspondence with the original $R0$ benchmark without confounding cross-round metrics with newly introduced software domains, the longitudinal multi-round panel ($R0$--$R4$) and diagnostic matrices ($R0$--$R3$) track the eight core OSWorld applications, while the full 20-application substrate supports broader benchmark synthesis and scope expansion. Benchmark evolution is evaluated on a heterogeneous panel of five computer-use agents: Kimi-K2.6, Seed-2.1-Pro, Gemini-3.5-Flash, Qwen-3.7-Plus, and GPT-5.5. Diagnostic evolution focuses on three representative agents—Kimi-K2.6, Seed-2.1-Pro, and Gemini-3.5-Flash—because it constructs a separate, agent-conditional task sequence for each target agent.

\paragraph{Benchmark-evolution protocol.} We start from a fixed panel of 100 OSWorld task lineages and evolve each lineage for four rounds, producing the original benchmark $R0$ and evolved suites $R1$--$R4$. Every round retains the same 100 lineage identifiers and is evaluated by all five agents, yielding 500 task--agent observations per round and 2,500 observations over the complete trajectory. This paired design prevents changes in aggregate performance from being attributed to a changing task sample. Each evolved task is also annotated with a seven-dimensional capability-demand vector on a 0--5 scale and with exact software--taxonomy--operation cells, enabling difficulty and coverage to be measured independently.

\paragraph{Diagnostic-evolution protocol.} Diagnostic evolution is run independently for each target agent. The current experiment contains four 100-task rounds for Kimi and three 100-task rounds each for Seed and Gemini, for a total of 1,000 evolved task profiles. The primary diagnostic matrices pool a matched 100-task $R0$ baseline with $R1$--$R3$ for each agent, yielding 1,200 planned agent--task--round slots before rollout availability and validity filtering. Kimi $R4$ is included only in task-generation and verifier analyses because it is outside the $R0$--$R3$ matrix scope. Profiles without eligible rollouts are retained as missing or unavailable evidence rather than imputed as failure.

\begin{figure*}[t]
\centering
\setlength{\tabcolsep}{2pt}

\begin{tabular}{c c c c c}
&
\textbf{Checkpoint Distribution} &
\textbf{Software Profile} &
\textbf{Requirement Coverage} &
\textbf{Requirement Alignment}
\\[2pt]

\rotatebox{90}{\textbf{Gemini}}
&
\begin{subfigure}[c]{0.225\textwidth}
  \centering
  \includegraphics[width=\linewidth]
  {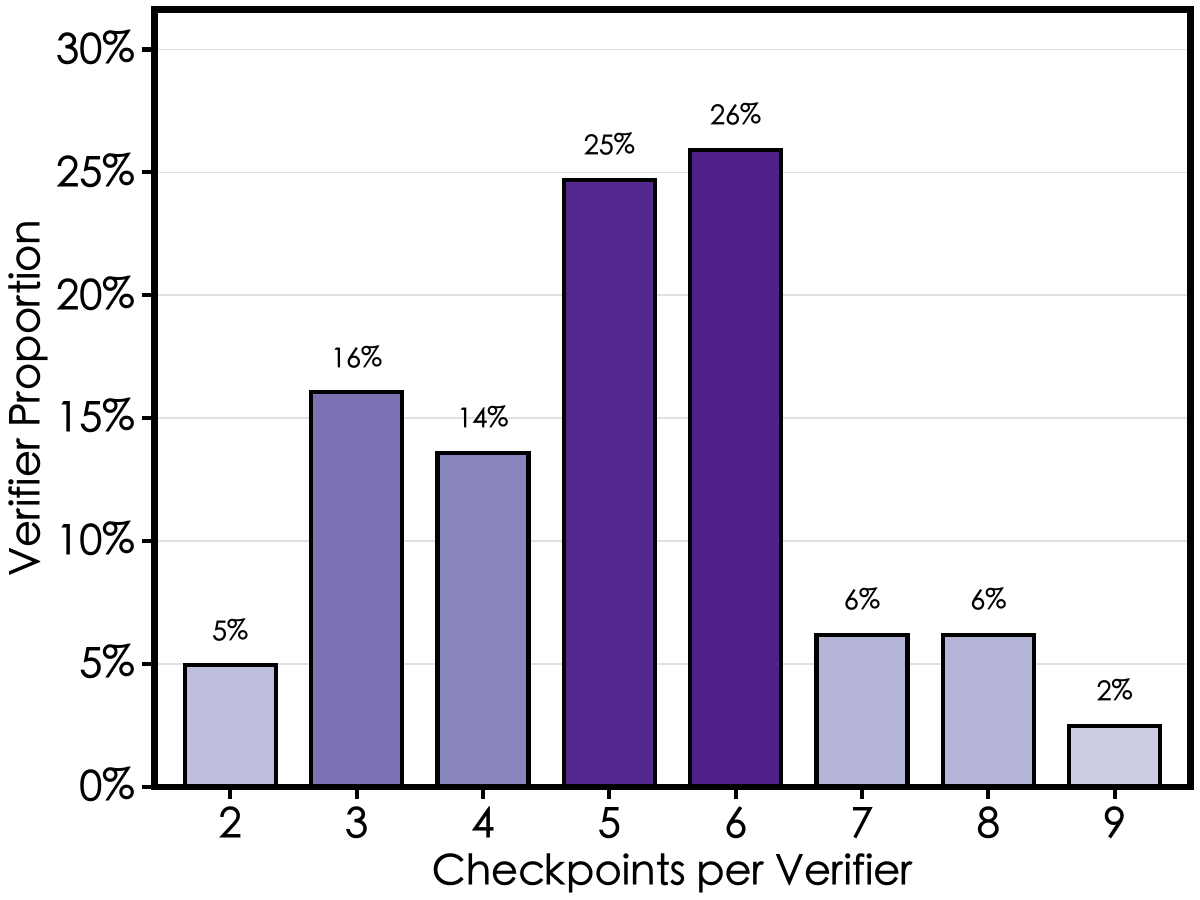}
  \phantomsubcaption
  \label{fig:verifier-gemini-distribution}
\end{subfigure}
&
\begin{subfigure}[c]{0.225\textwidth}
  \centering
  \includegraphics[width=\linewidth]
  {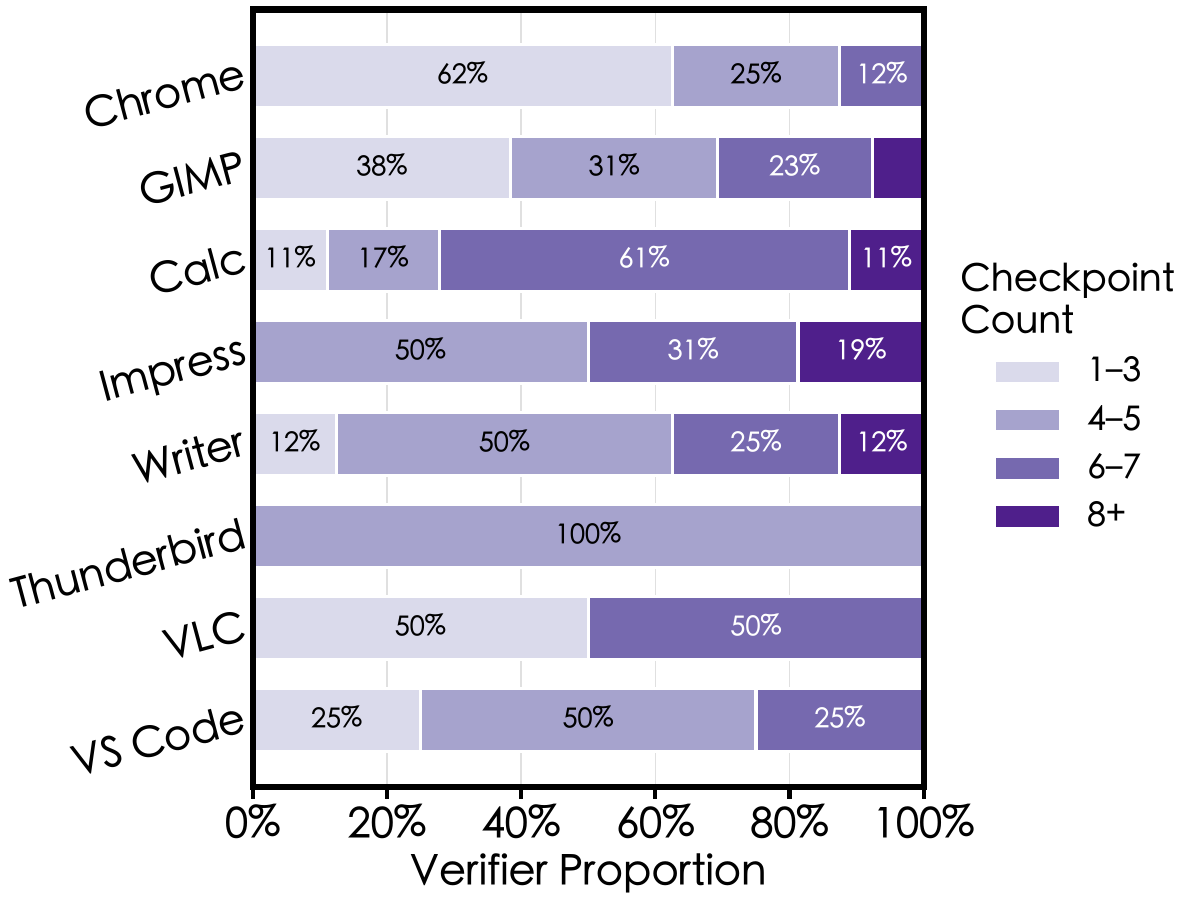}
  \phantomsubcaption
  \label{fig:verifier-gemini-profile}
\end{subfigure}
&
\begin{subfigure}[c]{0.225\textwidth}
  \centering
  \includegraphics[width=\linewidth]
  {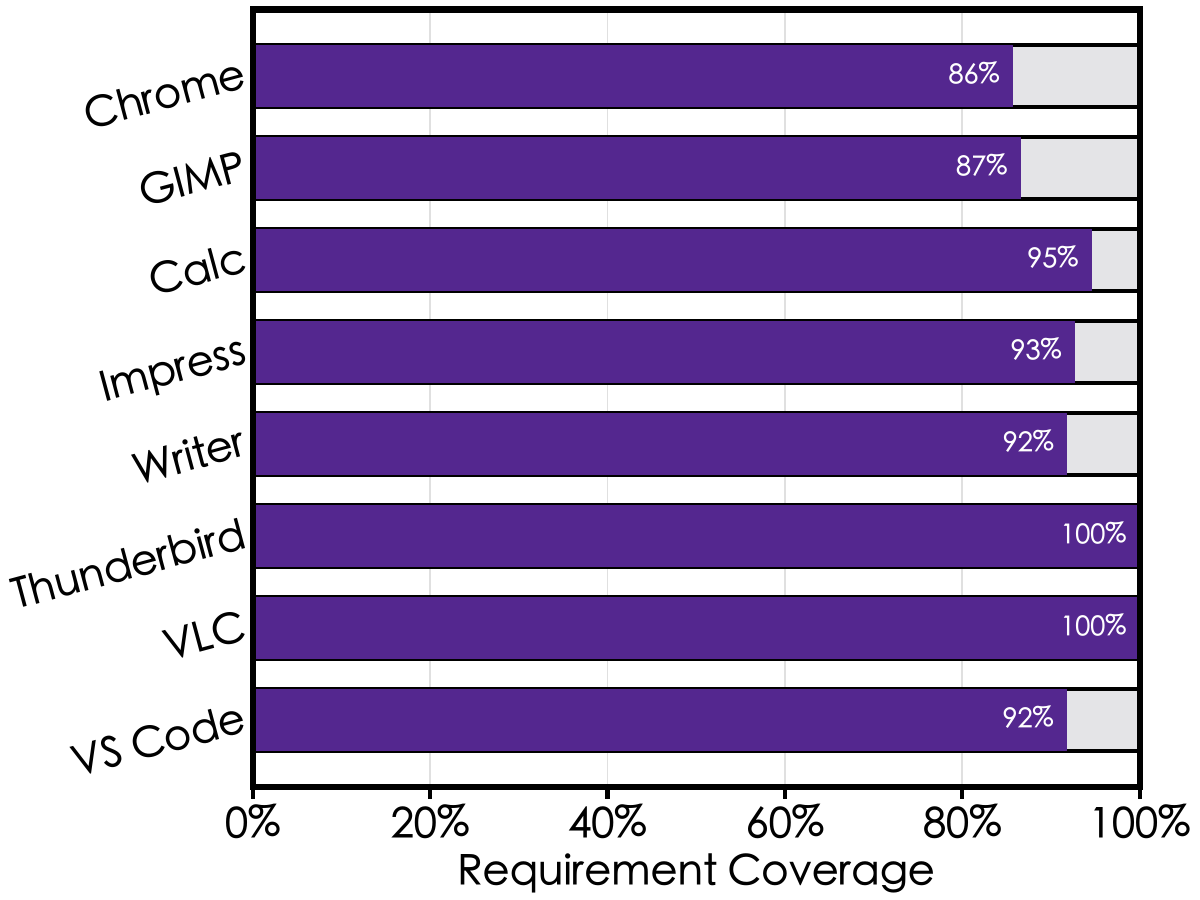}
  \phantomsubcaption
  \label{fig:verifier-gemini-coverage}
\end{subfigure}
&
\begin{subfigure}[c]{0.225\textwidth}
  \centering
  \includegraphics[width=\linewidth]
  {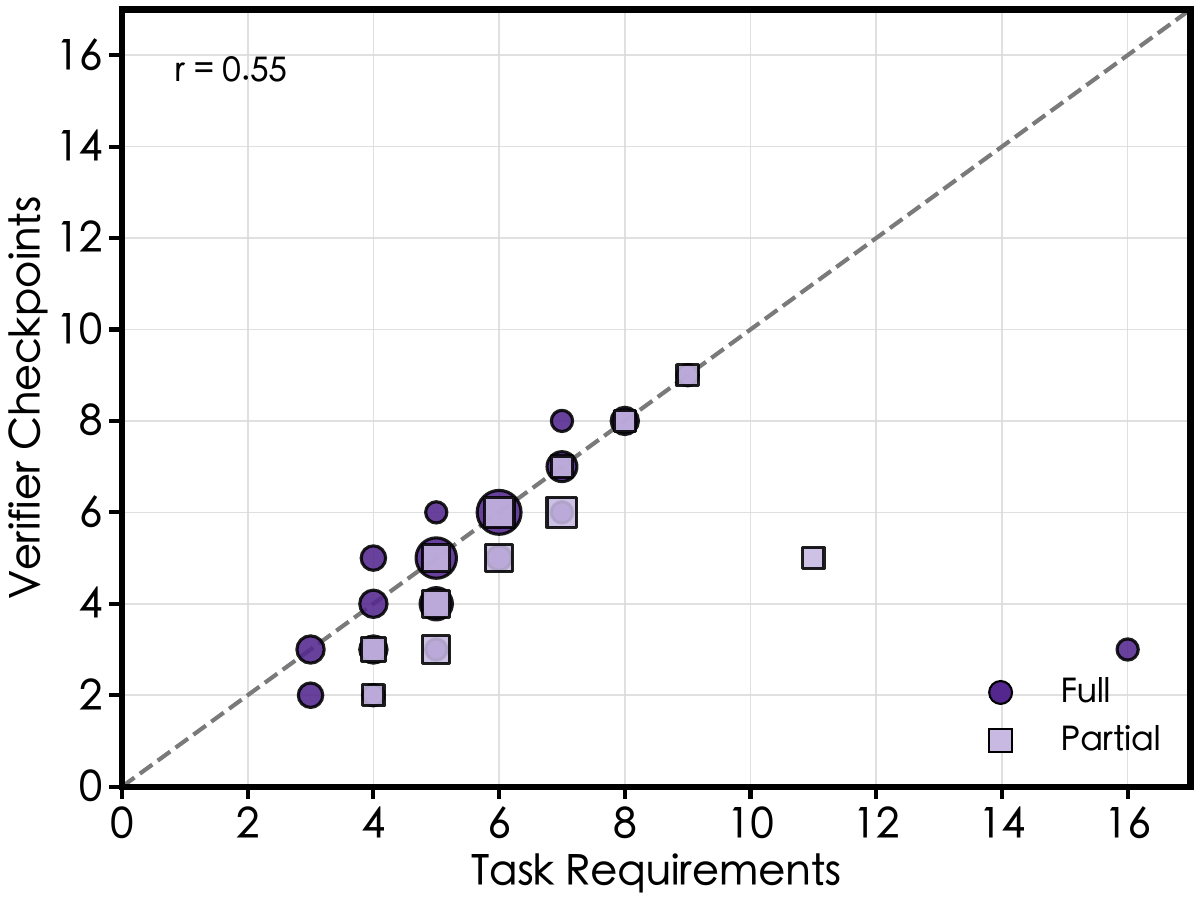}
  \phantomsubcaption
  \label{fig:verifier-gemini-alignment}
\end{subfigure}
\\[-1pt]

\rotatebox{90}{\textbf{Seed}}
&
\begin{subfigure}[c]{0.225\textwidth}
  \centering
  \includegraphics[width=\linewidth]
  {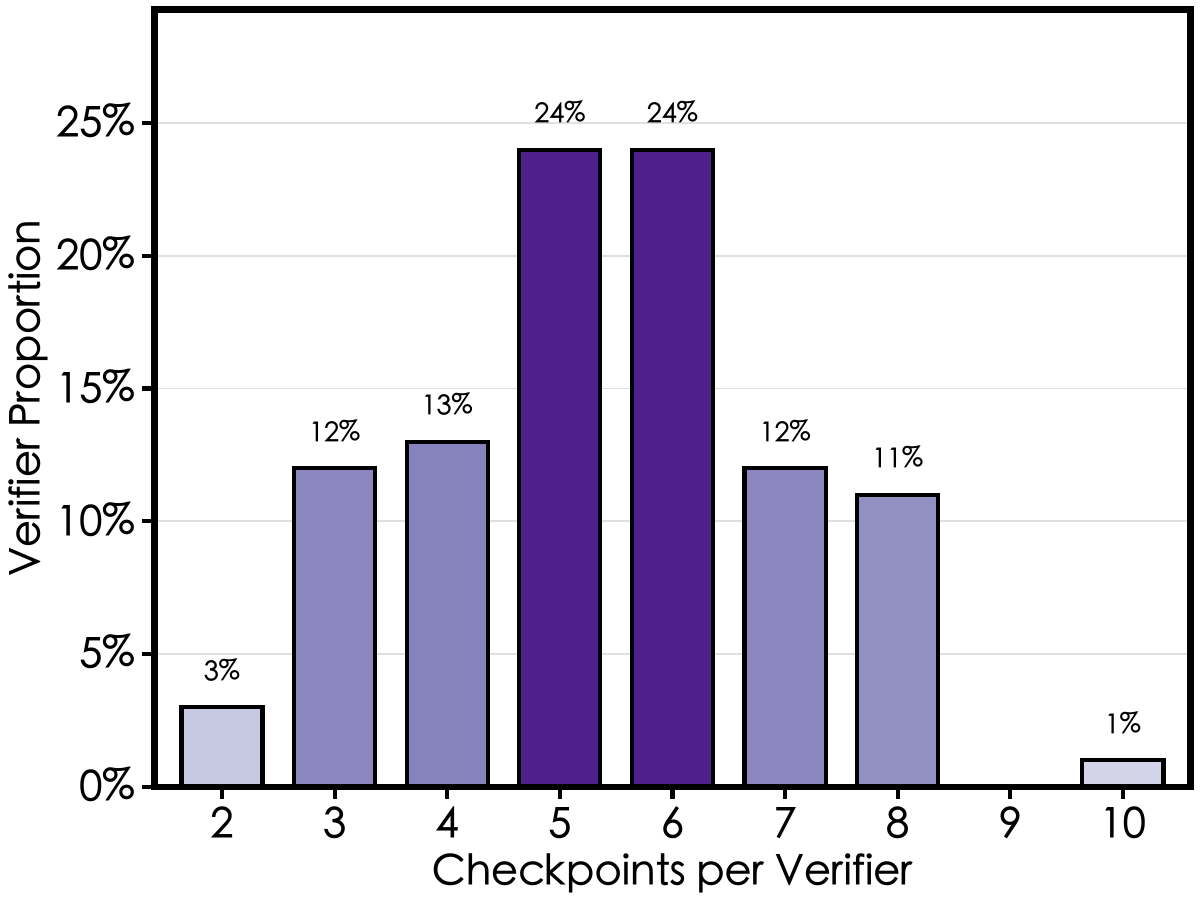}
  \phantomsubcaption
  \label{fig:verifier-seed-distribution}
\end{subfigure}
&
\begin{subfigure}[c]{0.225\textwidth}
  \centering
  \includegraphics[width=\linewidth]
  {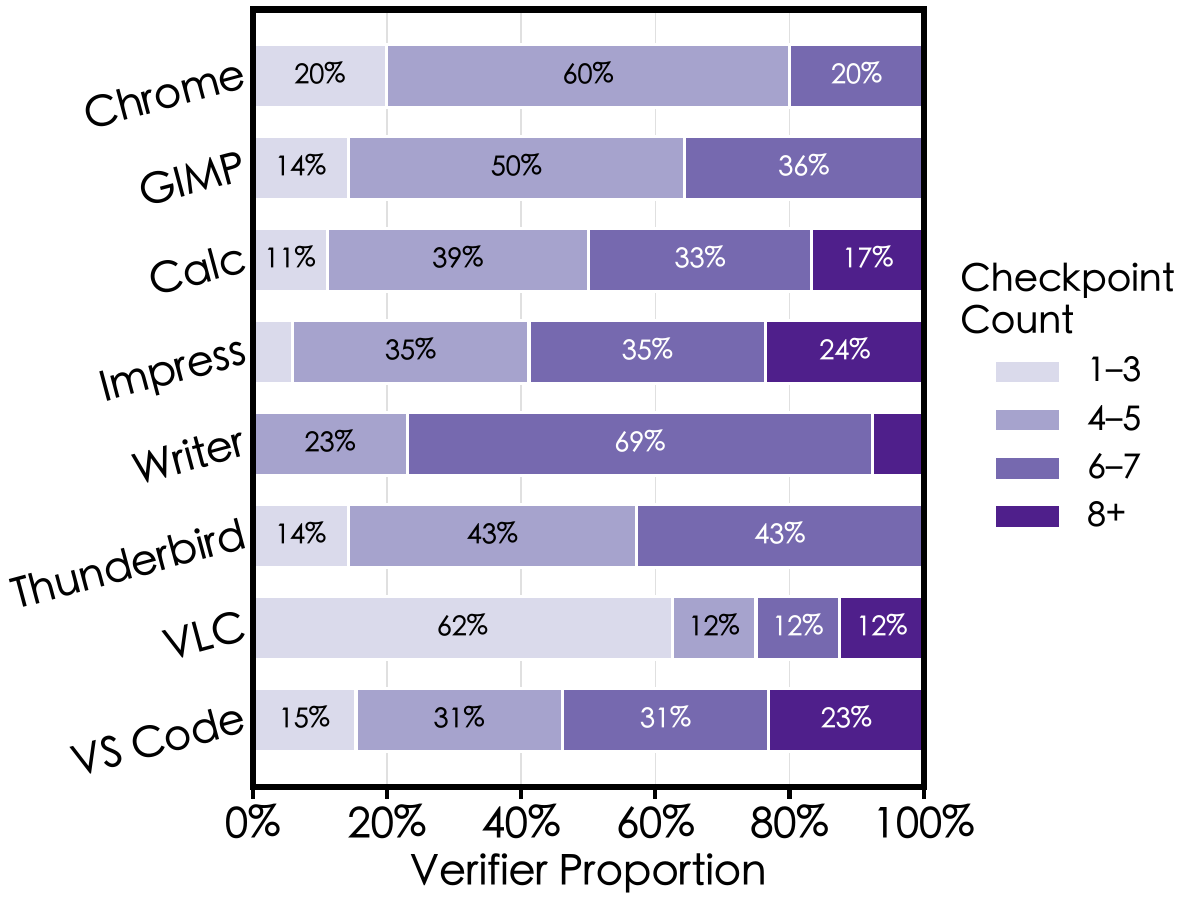}
  \phantomsubcaption
  \label{fig:verifier-seed-profile}
\end{subfigure}
&
\begin{subfigure}[c]{0.225\textwidth}
  \centering
  \includegraphics[width=\linewidth]
  {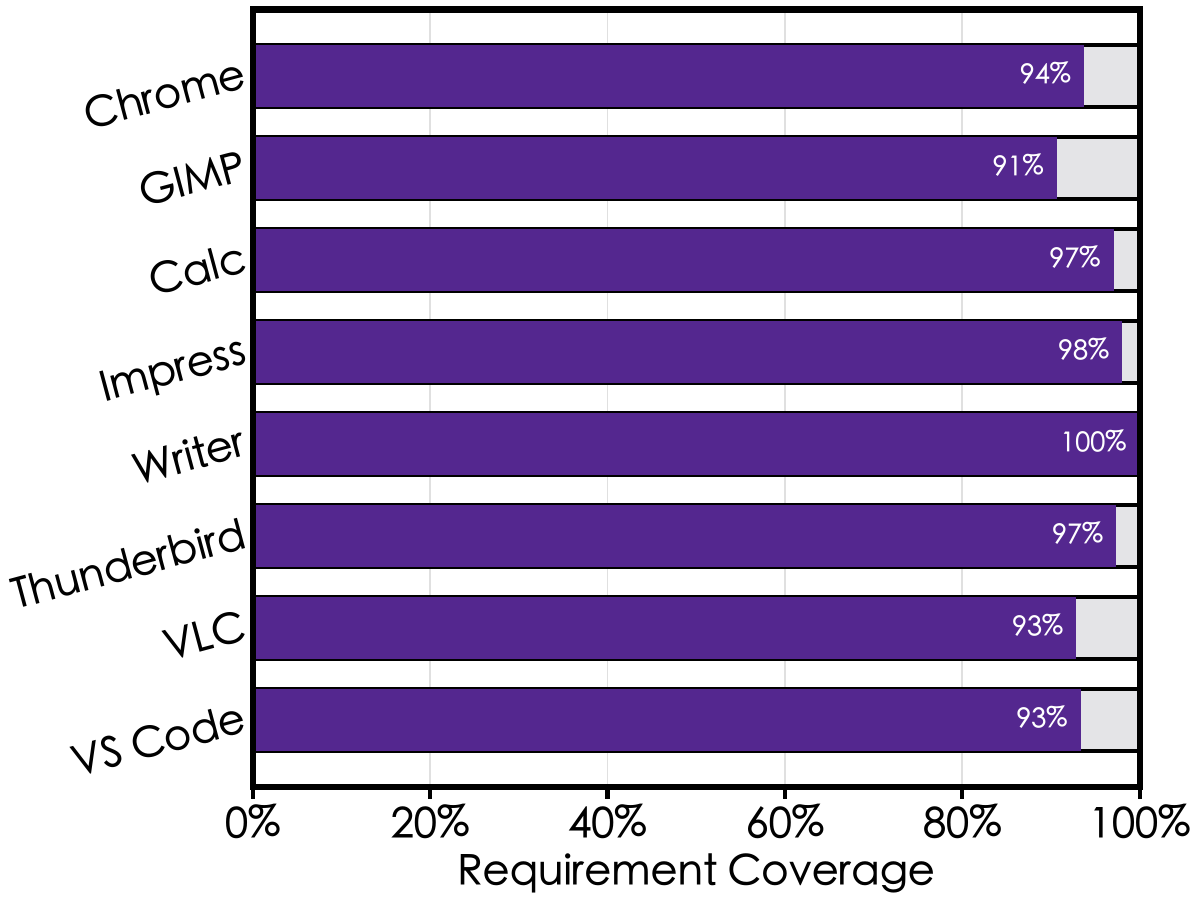}
  \phantomsubcaption
  \label{fig:verifier-seed-coverage}
\end{subfigure}
&
\begin{subfigure}[c]{0.225\textwidth}
  \centering
  \includegraphics[width=\linewidth]
  {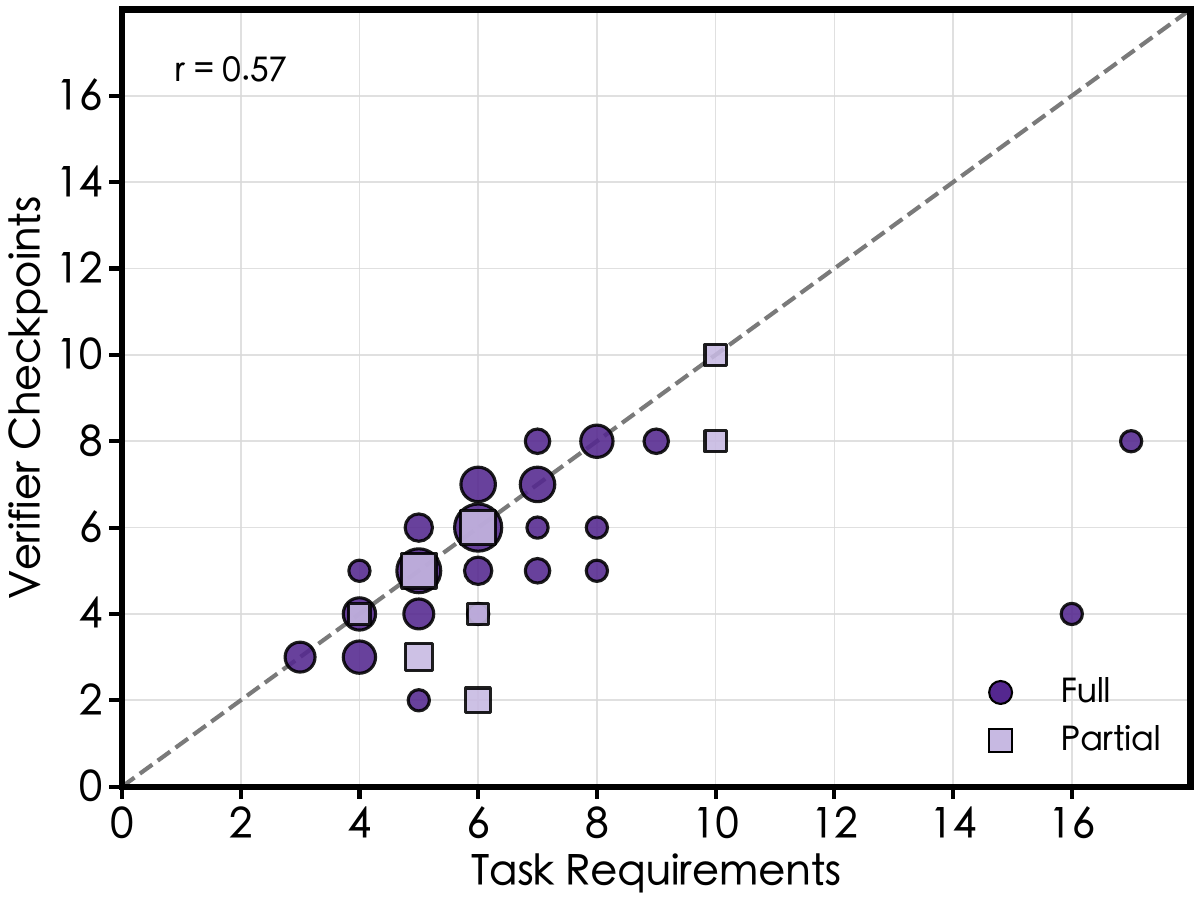}
  \phantomsubcaption
  \label{fig:verifier-seed-alignment}
\end{subfigure}
\end{tabular}

\caption{Verifier construction statistics across the remaining diagnostic benchmarks (Gemini and Seed; Kimi is shown in Figure~\ref{fig:verifier-coverage-alignment}). Rows correspond to the driving agents, while columns report checkpoint-count distributions, software-specific checkpoint profiles, requirement coverage, and checkpoint--requirement alignment.}
\label{fig:verifier-statistics-all-agents}
\end{figure*}


\begin{figure*}[t]
\centering
\setlength{\tabcolsep}{2pt}

\begin{tabular}{c c c c c}
&
\textbf{Capability Intensity} &
\textbf{Software--Capability} &
\textbf{Software--Action} &
\textbf{Software--Feature}
\\[2pt]

\rotatebox{90}{\textbf{Gemini}}
&
\begin{subfigure}[c]{0.225\textwidth}
  \centering
  \includegraphics[width=\linewidth]
  {Figures/gemini_r0-e3_capability_intensity.pdf}
  \phantomsubcaption
  \label{fig:diagnose-gemini-capability-intensity}
\end{subfigure}
&
\begin{subfigure}[c]{0.225\textwidth}
  \centering
  \includegraphics[width=\linewidth]
  {Figures/gemini_r0-e3_software_capability.pdf}
  \phantomsubcaption
  \label{fig:diagnose-gemini-software-capability}
\end{subfigure}
&
\begin{subfigure}[c]{0.225\textwidth}
  \centering
  \includegraphics[width=\linewidth]
  {Figures/gemini_r0-e3_software_action.pdf}
  \phantomsubcaption
  \label{fig:diagnose-gemini-software-action}
\end{subfigure}
&
\begin{subfigure}[c]{0.225\textwidth}
  \centering
  \includegraphics[width=\linewidth]
  {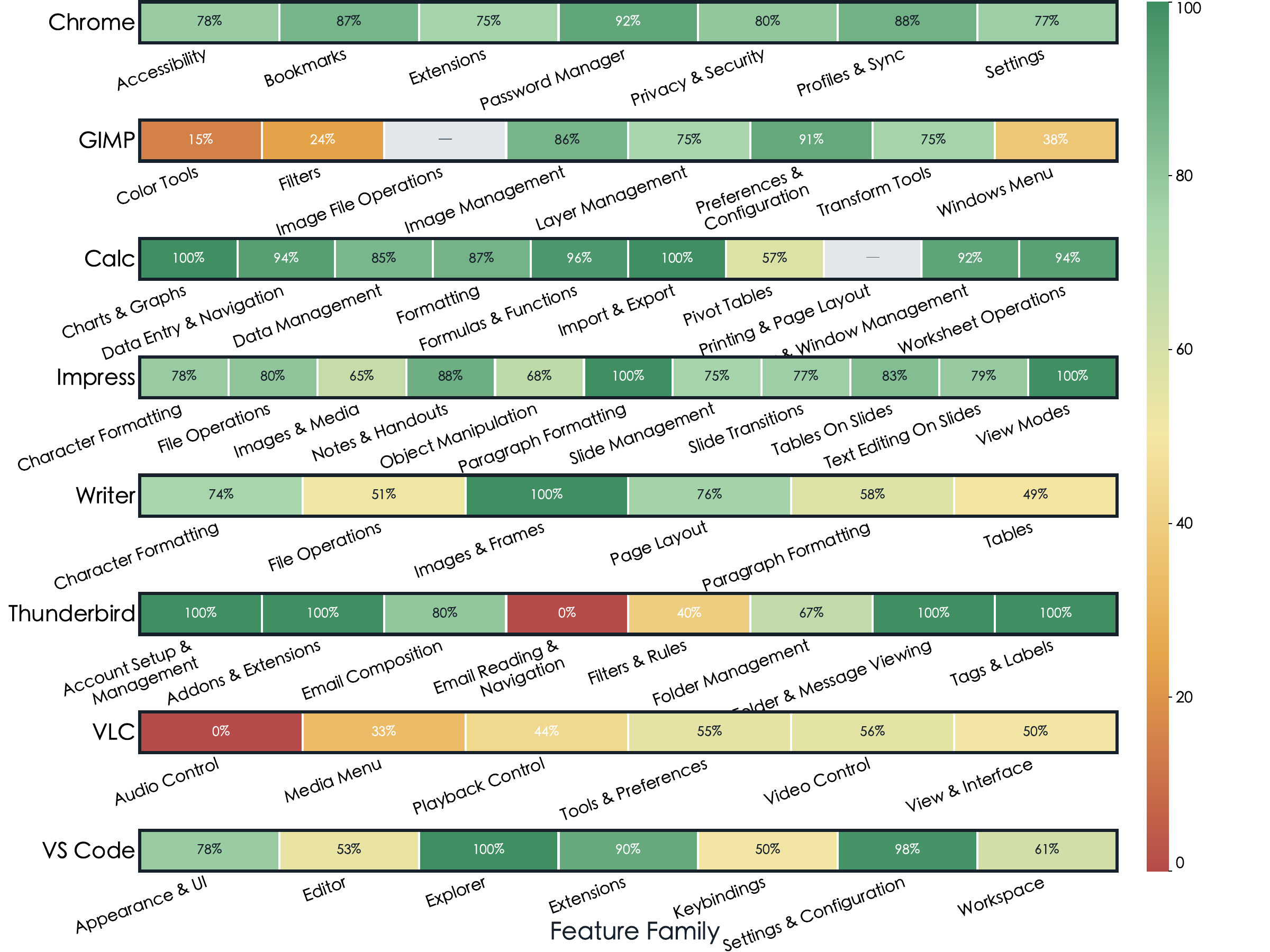}
  \phantomsubcaption
  \label{fig:diagnose-gemini-software-feature}
\end{subfigure}
\\[-1pt]

\rotatebox{90}{\textbf{Kimi}}
&
\begin{subfigure}[c]{0.225\textwidth}
  \centering
  \includegraphics[width=\linewidth]
  {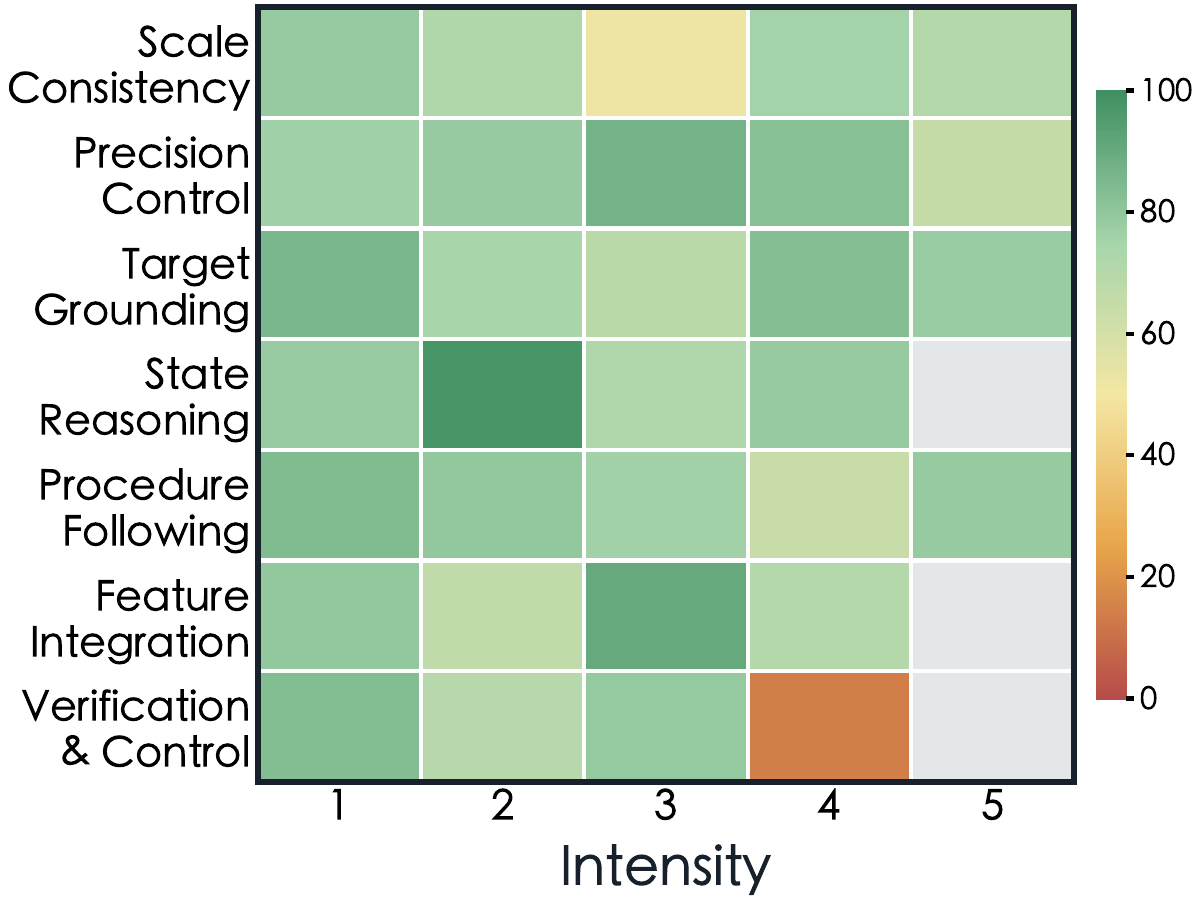}
  \phantomsubcaption
  \label{fig:diagnose-kimi-capability-intensity}
\end{subfigure}
&
\begin{subfigure}[c]{0.225\textwidth}
  \centering
  \includegraphics[width=\linewidth]
  {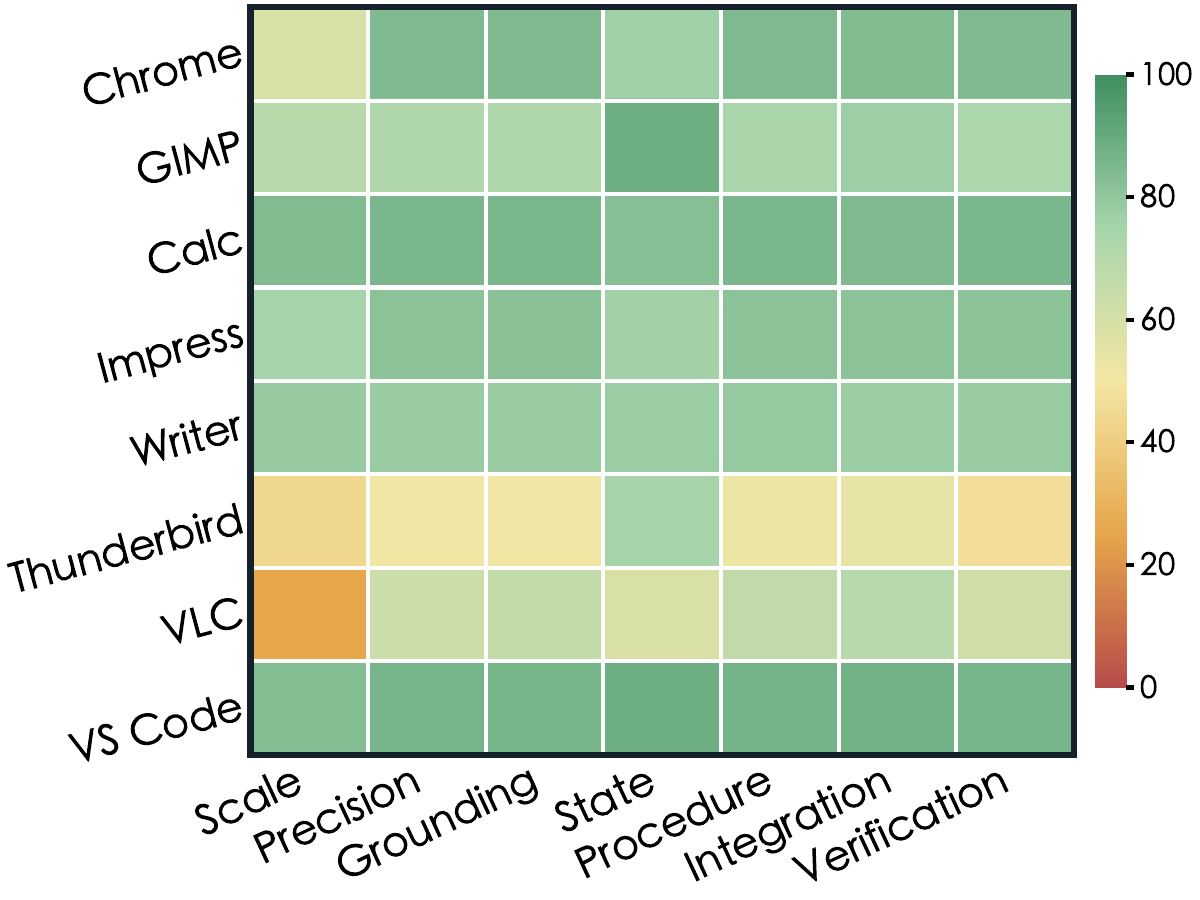}
  \phantomsubcaption
  \label{fig:diagnose-kimi-software-capability}
\end{subfigure}
&
\begin{subfigure}[c]{0.225\textwidth}
  \centering
  \includegraphics[width=\linewidth]
  {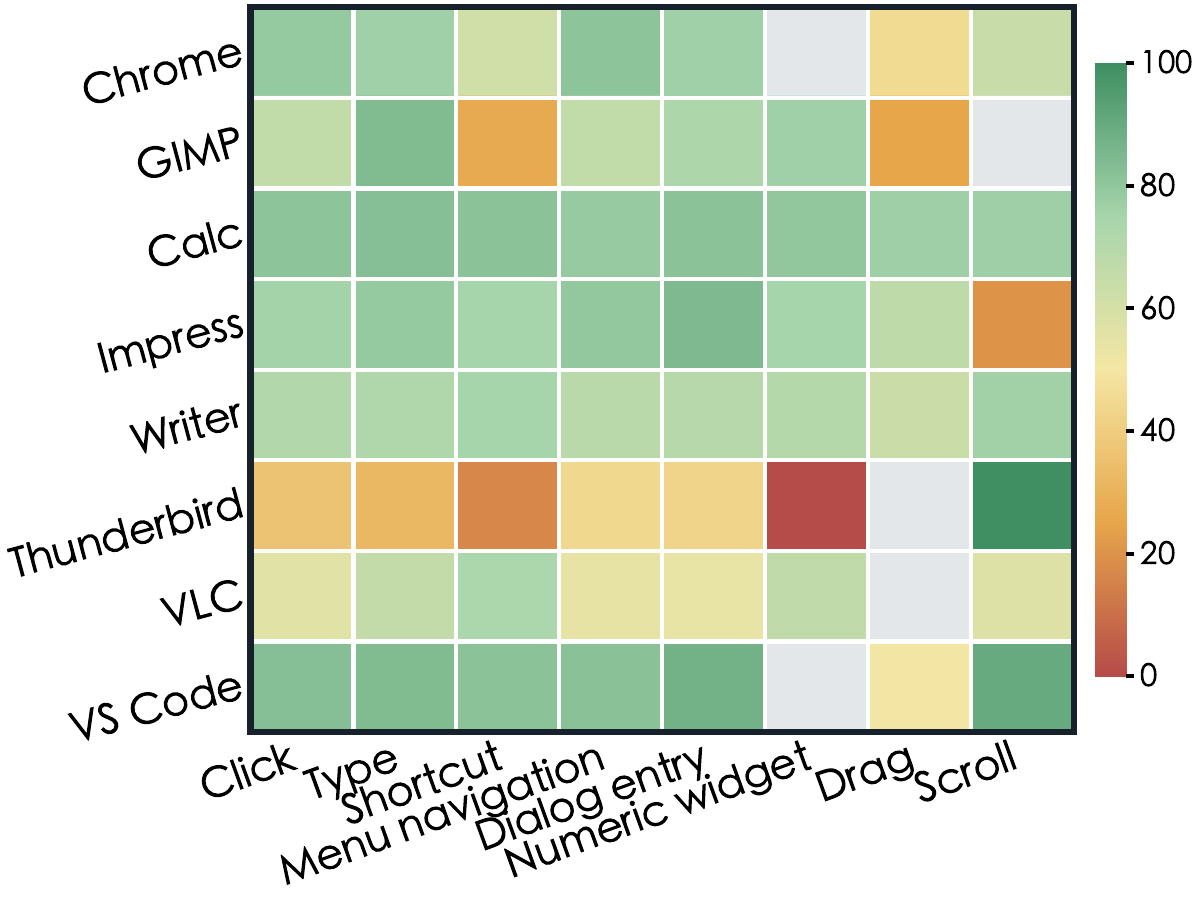}
  \phantomsubcaption
  \label{fig:diagnose-kimi-software-action}
\end{subfigure}
&
\begin{subfigure}[c]{0.225\textwidth}
  \centering
  \includegraphics[width=\linewidth]
  {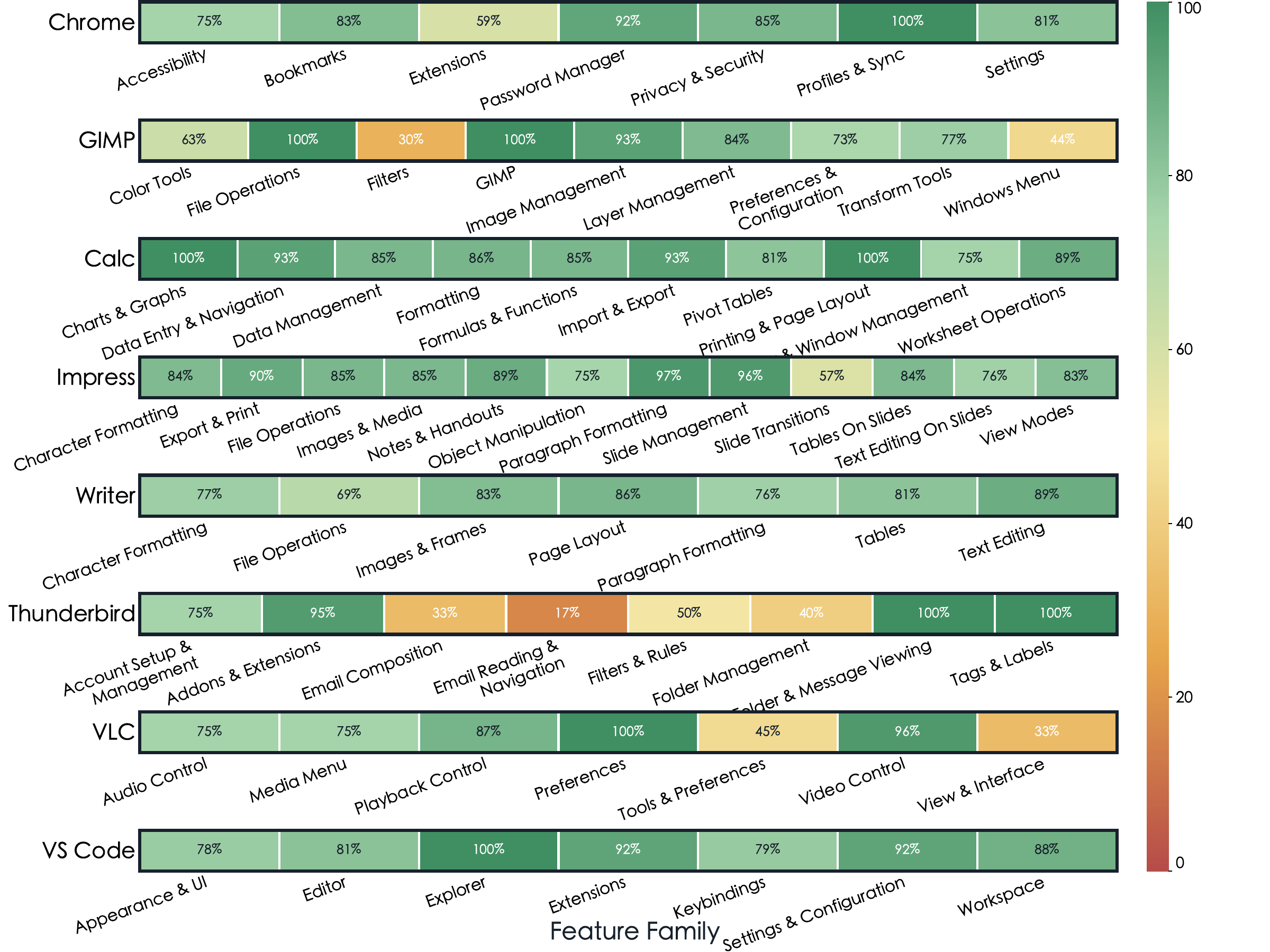}
  \phantomsubcaption
  \label{fig:diagnose-kimi-software-feature}
\end{subfigure}
\\[-1pt]

\rotatebox{90}{\textbf{Seed}}
&
\begin{subfigure}[c]{0.225\textwidth}
  \centering
  \includegraphics[width=\linewidth]
  {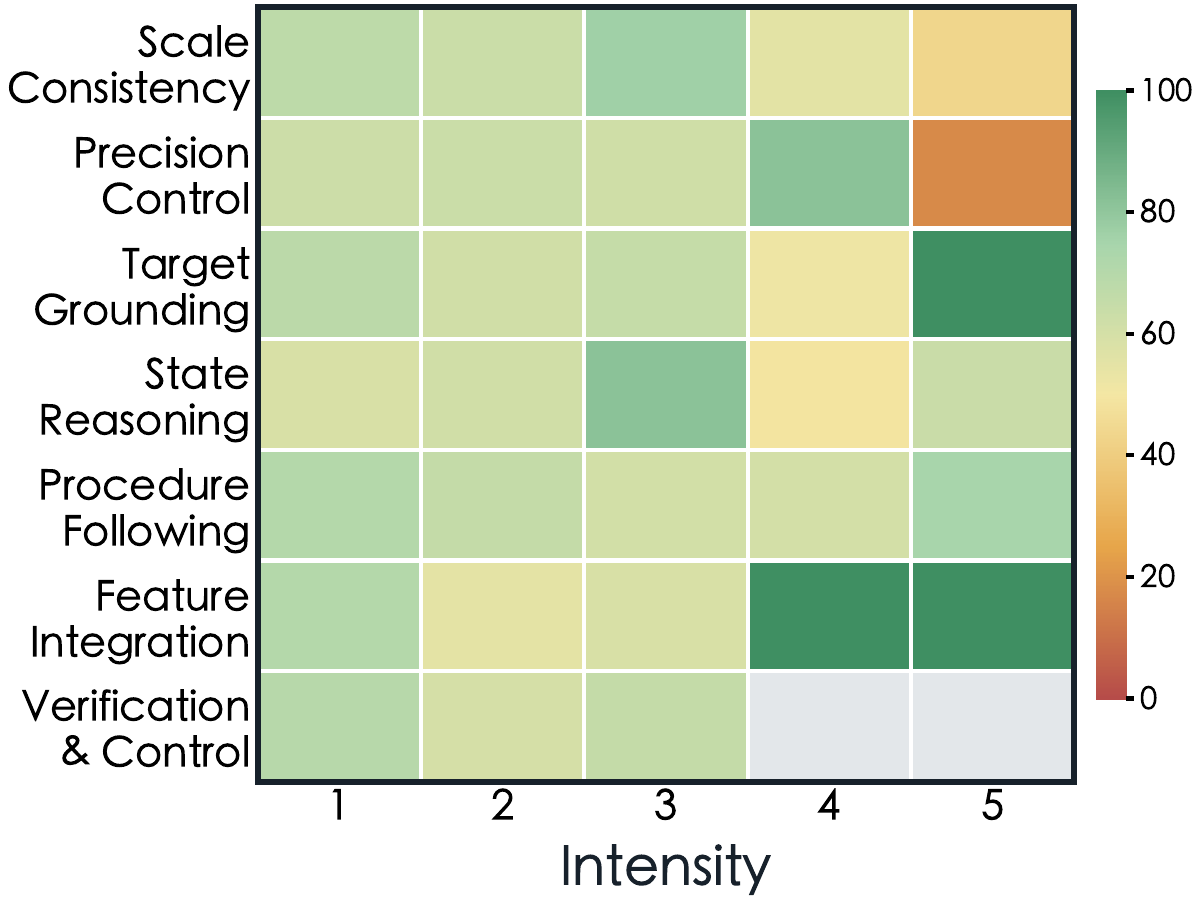}
  \phantomsubcaption
  \label{fig:diagnose-seed-capability-intensity}
\end{subfigure}
&
\begin{subfigure}[c]{0.225\textwidth}
  \centering
  \includegraphics[width=\linewidth]
  {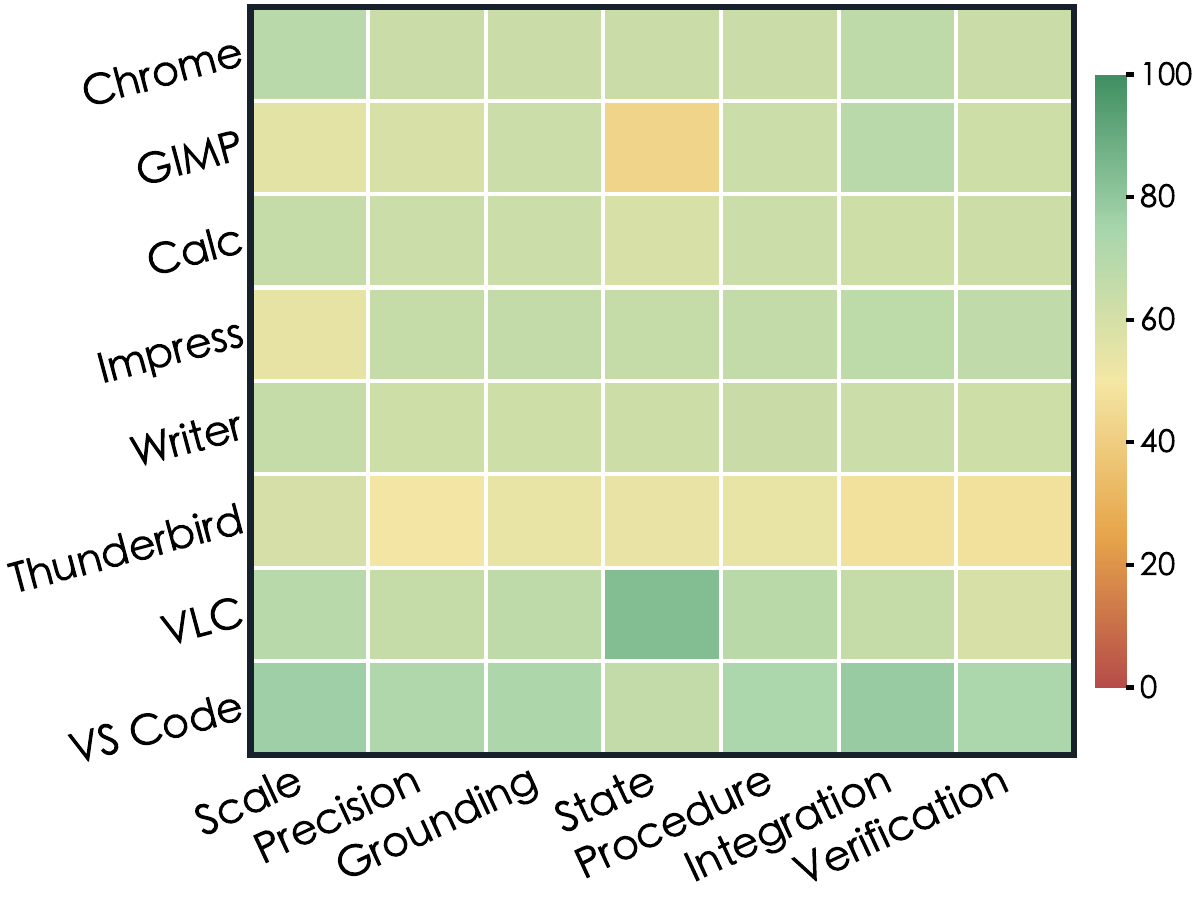}
  \phantomsubcaption
  \label{fig:diagnose-seed-software-capability}
\end{subfigure}
&
\begin{subfigure}[c]{0.225\textwidth}
  \centering
  \includegraphics[width=\linewidth]
  {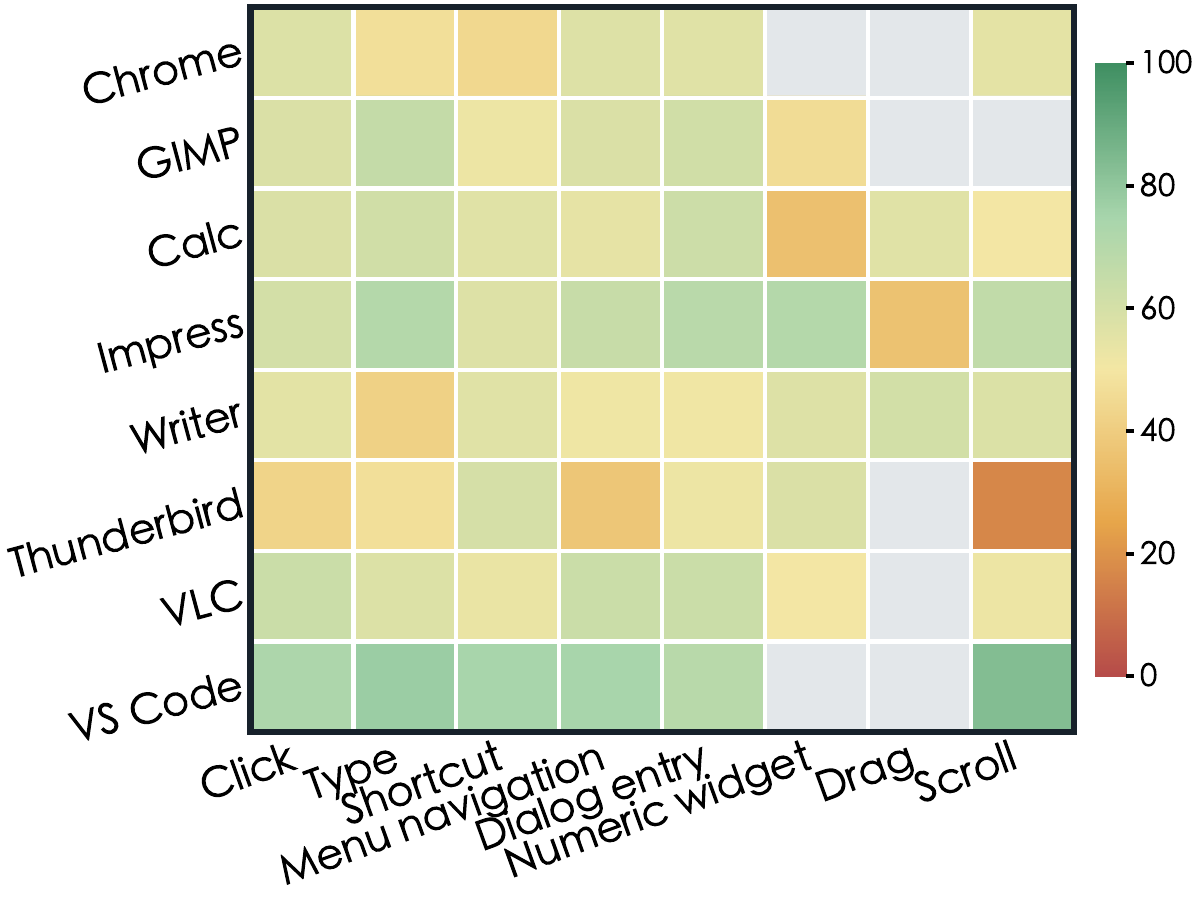}
  \phantomsubcaption
  \label{fig:diagnose-seed-software-action}
\end{subfigure}
&
\begin{subfigure}[c]{0.225\textwidth}
  \centering
  \includegraphics[width=\linewidth]
  {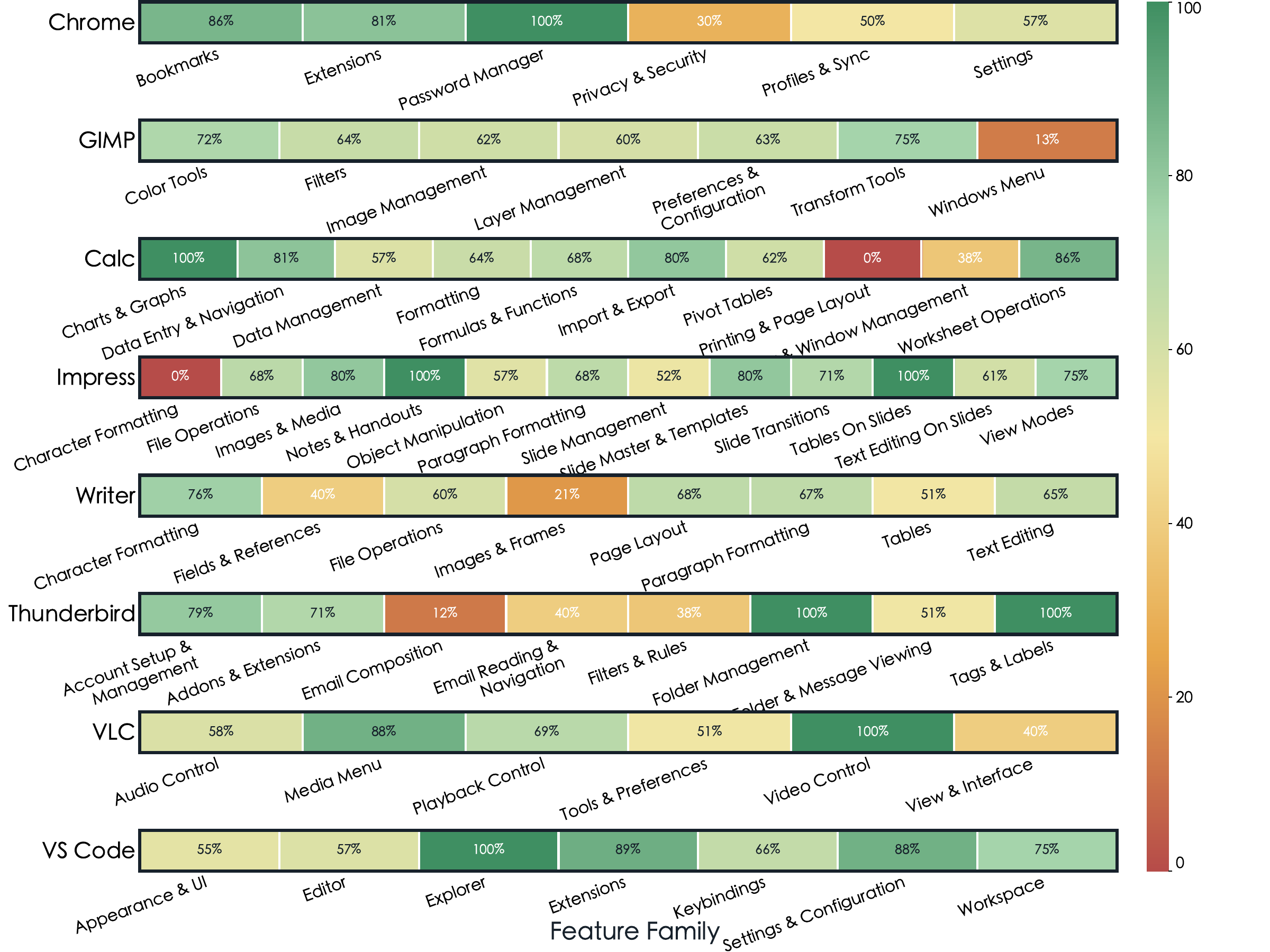}
  \phantomsubcaption
  \label{fig:diagnose-seed-software-feature}
\end{subfigure}
\end{tabular}

\caption{Diagnostic performance aggregated over $R0$--$R3$ for Gemini, Kimi, and Seed. Rows correspond to the evaluated agents, while columns report capability intensity, software--capability performance, software--action performance, and software--feature performance. Colors denote mean checkpoint scores; gray cells indicate unavailable evidence.}
\label{fig:diagnostic-matrices-all-agents}
\end{figure*}




\begin{figure*}[t]
\centering

\begin{subfigure}[t]{0.49\textwidth}
\centering
\includegraphics[width=\linewidth]{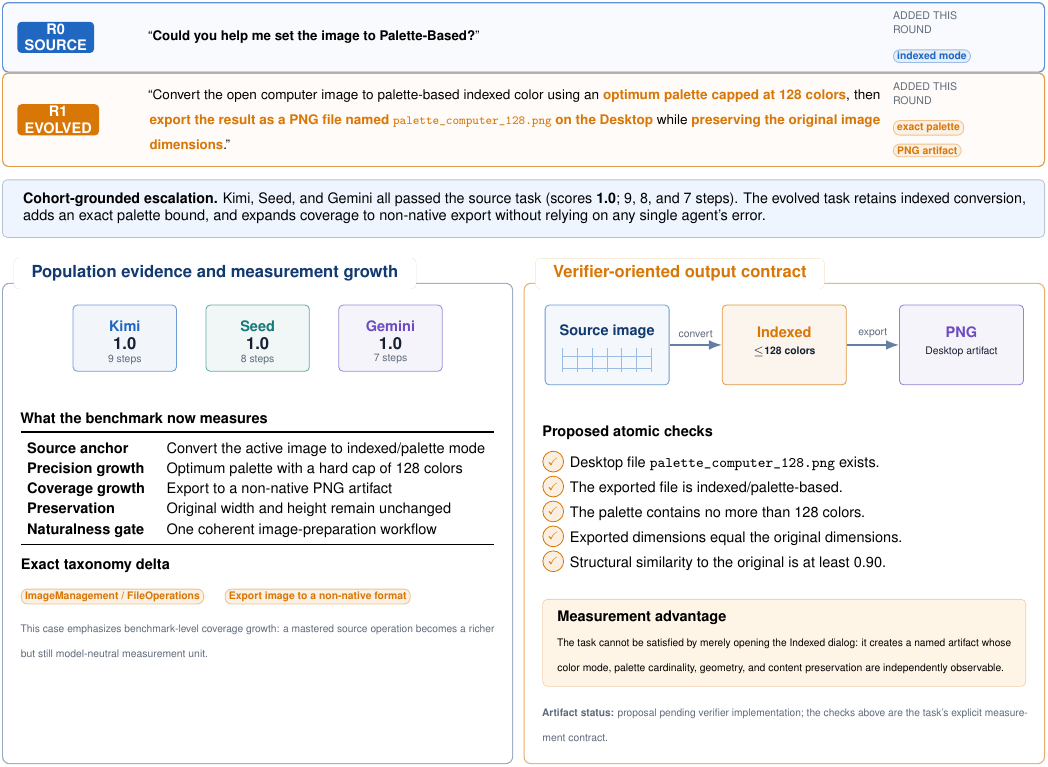}
\caption{Additional benchmark-evolution case study based on a GIMP indexed-color export task.}
\label{fig:appendix_case_gimp_export}
\end{subfigure}
\hfill
\begin{subfigure}[t]{0.49\textwidth}
\centering
\includegraphics[width=\linewidth]{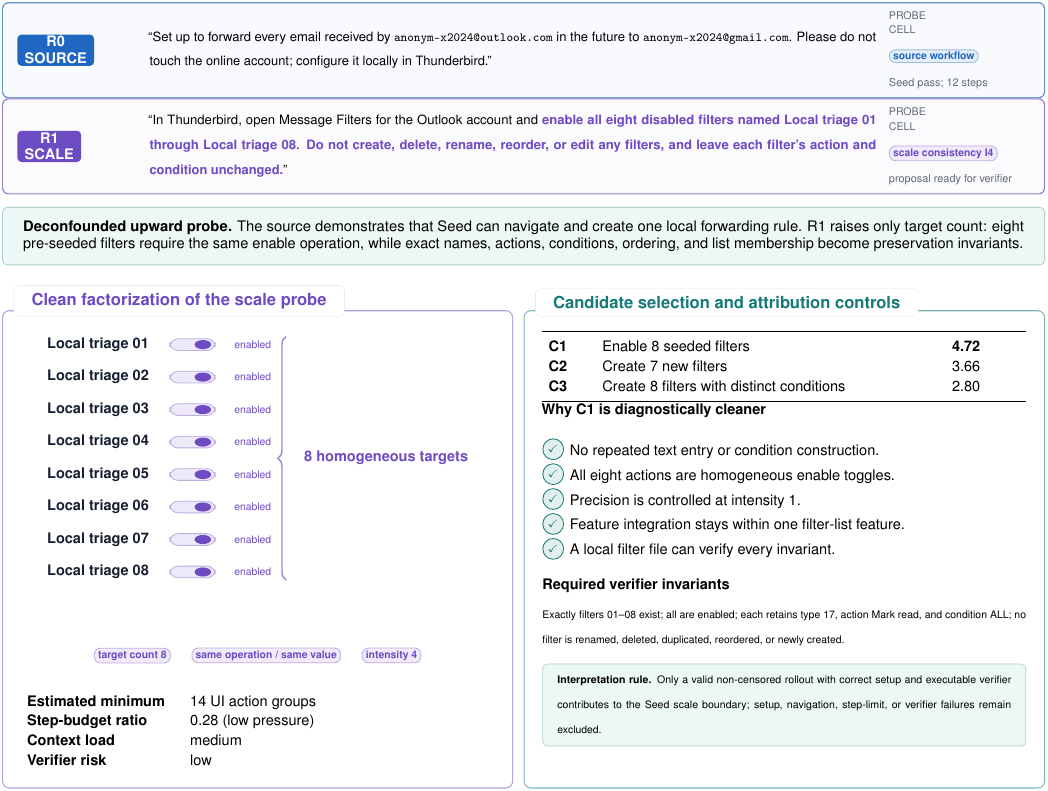}
\caption{Additional diagnostic-evolution case study for Seed using a Thunderbird message-filter
task.}
\label{fig:appendix_case_seed_thunderbird}
\end{subfigure}

\caption{Additional case studies illustrating benchmark evolution and agent-conditioned diagnostic evolution.}
\label{fig:appendix_evolution_cases}
\end{figure*}

 \begin{figure*}[t]
  \centering

  \begin{subfigure}[t]{0.49\textwidth}
  \centering
  \includegraphics[width=\linewidth]{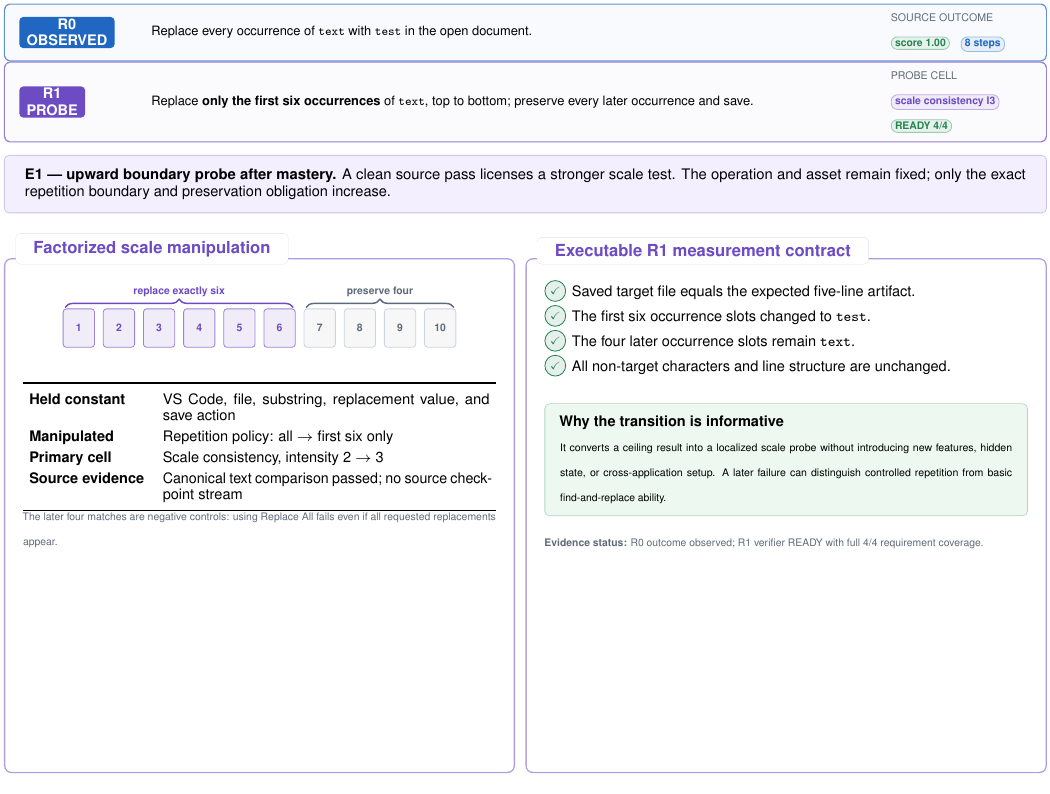}
  \caption{Round-$R1$ diagnostic evolution for Kimi, strengthening a mastered VS Code replacement task
  into a controlled scale-consistency probe.}
  \label{fig:appendix_case_e1_kimi_vscode}
  \end{subfigure}
  \hfill
  \begin{subfigure}[t]{0.49\textwidth}
  \centering
  \includegraphics[width=\linewidth]{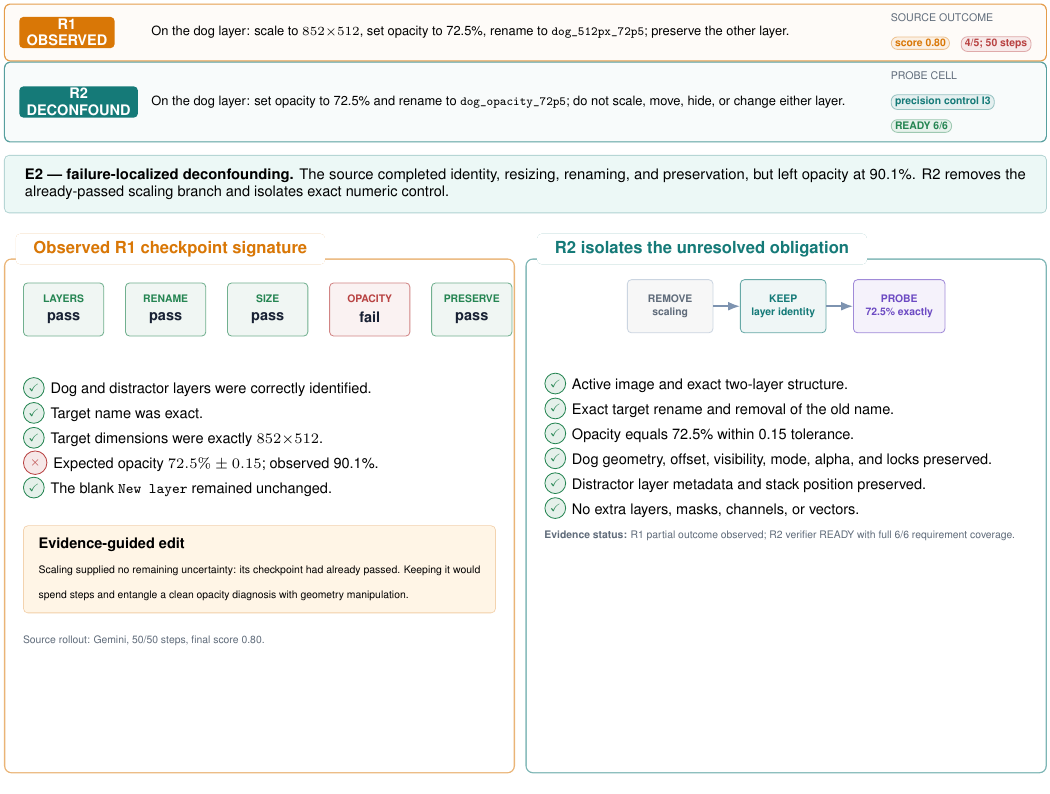}
  \caption{Round-$R2$ diagnostic evolution for Gemini, isolating exact opacity control from an otherwise
  successful GIMP task.}
  \label{fig:appendix_case_e2_gemini_gimp}
  \end{subfigure}

  \vspace{0.8em}

  \begin{subfigure}[t]{0.49\textwidth}
  \centering
  \includegraphics[width=\linewidth]{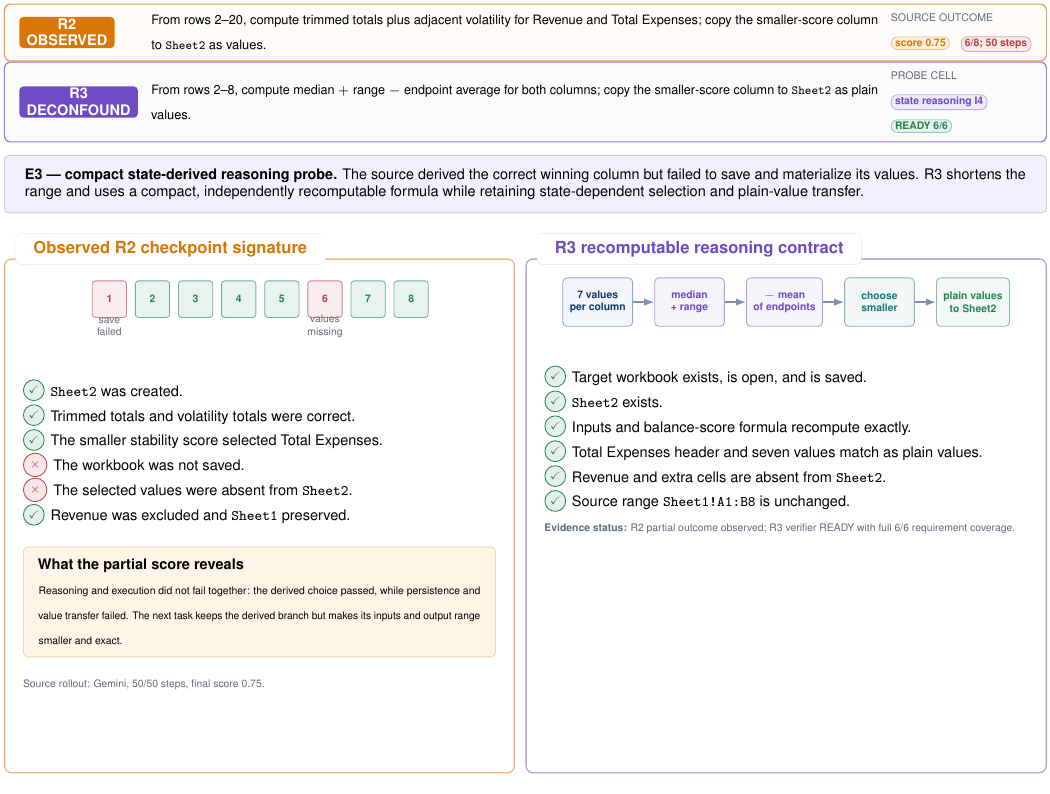}
  \caption{Round-$R3$ diagnostic evolution for Gemini, separating state-dependent reasoning from value-
  transfer failures in Calc.}
  \label{fig:appendix_case_e3_gemini_calc}
  \end{subfigure}
  \hfill
  \begin{subfigure}[t]{0.49\textwidth}
  \centering
  \includegraphics[width=\linewidth]{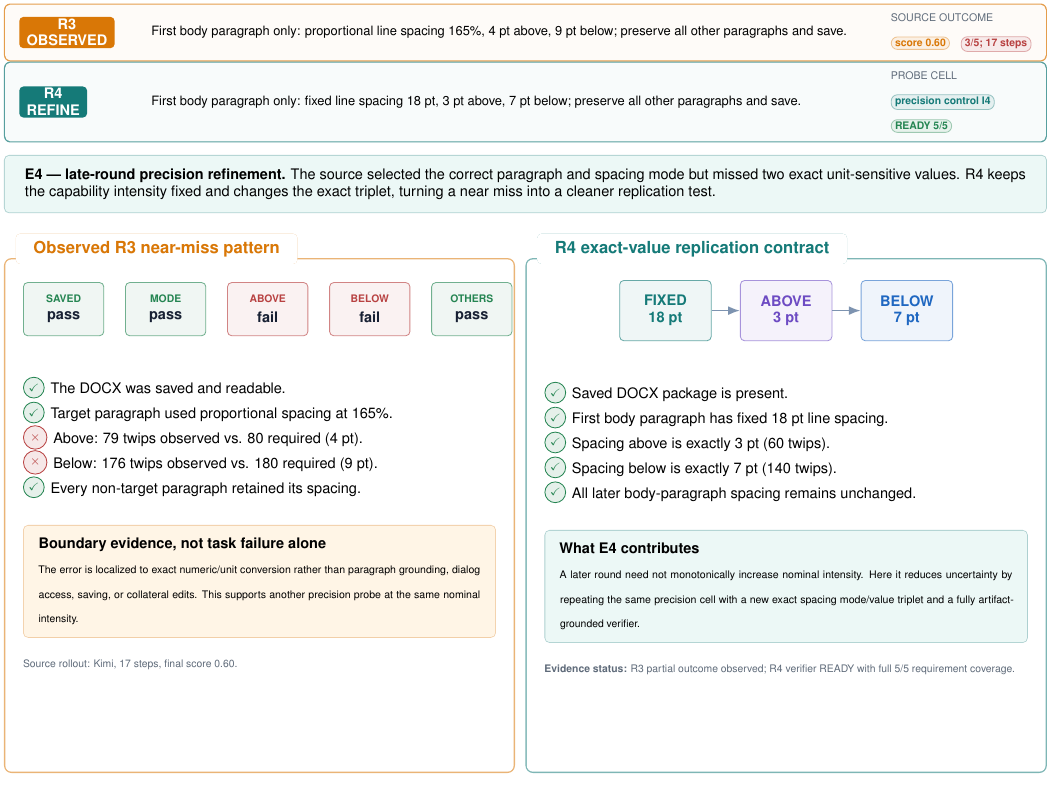}
  \caption{Round-$R4$ diagnostic evolution for Kimi, refining an exact, unit-sensitive paragraph-
  formatting boundary in Writer.}
  \label{fig:appendix_case_e4_kimi_writer}
  \end{subfigure}

  \caption{Diagnostic-evolution case studies across four successive rounds. Each transition uses the
  preceding rollout outcome to strengthen a mastered task, remove confounding requirements, or refine
  an uncertain capability boundary.}
  \label{fig:appendix_round_transition_cases}
  \end{figure*}

\paragraph{Rollout parameters.} Each episode is limited to 50 environment-interaction steps. The environment records the complete action trajectory, the terminal normalized task score, and state-hook checkpoint traces. We classify scores of at least $0.99$ as pass and scores of at most $0.01$ as fail; intermediate scores retain partial credit. A zero-score episode that reaches the 50-step limit is classified as censored rather than as clean failure. Unless otherwise stated, partially observable but sound verifiers are retained and their uncovered requirements are reported explicitly; the stricter full-verifier-only filter is used only in designated sensitivity analyses. For retried executions, aggregation selects the latest completed run within each published task instance.

\paragraph{Compute and deployment.} Task synthesis, verifier construction, and result aggregation are executed on two CPU servers. The primary OSW server runs Ubuntu 22.04 with 96 Intel Xeon Platinum 8163 vCPUs and 376~GiB RAM; the auxiliary SCI server runs Ubuntu 20.04 with 256 AMD EPYC 7T83 vCPUs and 1008~GiB RAM. Neither server uses a local GPU because model inference is provided by remote Codex subscription or API services. Evolved task JSON files, dataset manifests, and executable verifiers are published, whose rollout controller runs on the OSW server at \texttt{http://localhost:8086}. The controller launches isolated Ubuntu desktop environments, applies task setup, executes the selected agent, invokes the state-hook verifier after interaction steps, and stores \texttt{traj.json}, \texttt{checkpoints.jsonl}, and \texttt{result.txt} for subsequent analysis.

\subsection{Implementation}
\label{sec:implementation}

\paragraph{Skill-driven synthesis.} Our implementation represents each methodological component as a versioned Codex skill rather than as an unconstrained one-shot prompt. The benchmark evolver consumes aligned multi-agent trajectories and the shared coverage registry; the diagnostic batch planner converts longitudinal evidence into a round-level dispatch; the diagnostic task evolver realizes one controlled probe per dispatch row; and the verifier evolver reconstructs and validates the corresponding executable measurement contract. Every worker is instructed to read the relevant skill specification and schema before accessing task artifacts. Thus, capability rubrics, admissible evolution operators, confound limits, output schemas, and validation gates remain fixed across parallel workers and rounds.

\paragraph{Concurrent orchestration.} Batch analysis is performed once before each diagnostic round, after which independent task workers consume the resulting \texttt{dispatch.jsonl} in parallel. Benchmark tasks and task-specific verifiers are parallelized similarly, while shared feature and capability registries use atomic claims to prevent duplicated coverage gains. The reusable runners default to three concurrent Codex workers, a two-second launch stagger, a 45-minute per-task timeout, and resumable execution that skips already validated terminal artifacts; bounded repair and verifier jobs may raise concurrency to ten. Machine-readable run summaries record task status, duration, token usage, retries, validation errors, and the exact output directory.

\paragraph{Hybrid Codex backends.} We use GPT-5.5 for task and verifier synthesis and explicitly set reasoning effort to \texttt{high} for the reported runs. Workloads are distributed across two isolated execution modes: a ChatGPT/Codex subscription backend authenticated through a persistent \texttt{CODEX\_HOME}, and an API backend using an OpenAI-compatible Responses interface through Routify. A worker uses exactly one backend; credentials and provider-specific environment variables from the other backend are removed, and there is no silent cross-backend fallback. This hybrid deployment allows subscription and API capacity to be combined at the campaign level without changing the skill, prompt contract, model identifier, or output schema.

\paragraph{Validation and provenance.} Task workers must produce structured task, capability-profile, analysis, and task-card artifacts, all of which pass schema and semantic consistency checks before entering a round. Verifier workers additionally run syntax, interface-contract, schema-grounding, PRE/PASS discrimination, and negative-fixture checks before a task is marked rollout-ready. Source identifiers, lineage operations, model/backend settings, token accounting, and rollout paths are preserved in machine-readable summaries. This artifact-first design makes every plotted value traceable to a task profile, one selected rollout, and one validated verifier.

\subsection{Metrics}
\label{sec:evaluation_metrics}

\paragraph{Normalized task score and pass rate.} Let $s_{a,t}^{(r)}\in[0,1]$ denote the normalized evaluator score obtained by agent $a$ on task lineage $t$ in round $r$. The benchmark-level agent score is the macro mean over the fixed agent--task panel, $100\,|\mathcal{A}|^{-1}|\mathcal{T}|^{-1}\sum_{a\in\mathcal{A}}\sum_{t\in\mathcal{T}}s_{a,t}^{(r)}$. Per-agent and per-software scores apply the same arithmetic mean to the corresponding subset. Pass rate is the percentage of episodes with $s_{a,t}^{(r)}=1$, so partially completed tasks affect mean score but not pass rate. In the diagnostic analysis, the episode score is the normalized state-hook checkpoint score stored in the selected rollout's \texttt{result.txt}; averaging these scores preserves substantially more information than binarizing every episode.

\paragraph{Interaction length and token usage.} Episode length is the number of recorded environment-interaction steps in \texttt{traj.json}. The round-trend figure reports $100\bar{L}/50$, i.e., the mean fraction of the 50-step budget, whereas result tables report the unnormalized mean number of steps. Input and output tokens are final-selected-run, per-task means reported in thousands, not totals over a round. Failed and superseded retry attempts are excluded; input tokens include cached input tokens when the provider reports them, while output tokens use the recorded output-token field directly. Seed $R0$ lacks token logs, and Gemini $R0$ token means are computed from the 90 selected runs for which logs are available.

\paragraph{Empirical difficulty and boundary informativeness.} Empirical difficulty is $D^{(r)}=100(1-\bar{s}^{(r)})$, where $\bar{s}^{(r)}$ is the macro agent score before multiplication by 100. To measure whether a task lies near the cohort's capability boundary, we first compute $p_t^{(r)}=|\mathcal{A}|^{-1}\sum_a s_{a,t}^{(r)}$ and then report $I^{(r)}=100|\mathcal{T}|^{-1}\sum_t4p_t^{(r)}(1-p_t^{(r)})$. This quantity is maximal when the cohort obtains a mean score of $0.5$ and vanishes for universally solved or universally failed tasks. It therefore measures potential information near the population boundary rather than cross-agent variance.

\paragraph{Capability demand and coverage.} Each task has an intensity $z_{t,d}^{(r)}\in\{0,\ldots,5\}$ for each of seven capability dimensions. Mean capability demand is $100|\mathcal{T}|^{-1}\sum_t\sum_d z_{t,d}^{(r)}/35$. A capability--intensity cell is a nonzero pair $(d,z)$ realized by at least one task. Snapshot coverage counts unique cells within one round, while cumulative coverage counts the set union from $R0$ through the current round and divides by the fixed universe of $7\times5=35$ cells. These are task-demand statistics and must not be interpreted as agent ability scores.

\paragraph{Feature and software coverage.} The finest feature cell is defined by an exact software, taxonomy-category, and operation tuple. \emph{Current} coverage counts unique canonical tuples in one round; $\Delta$ is the change in this snapshot count from the preceding round; cumulative feature coverage is the normalized set union through the current round; and \emph{Frontier} denotes the committed shared-registry state used to coordinate concurrent evolution. The cumulative profile union and registry frontier can differ slightly because they are produced at different validation stages and because legacy cells are canonicalized before plotting. Category breadth collapses operations to unique software--taxonomy-category cells, feature-family coverage further collapses a taxonomy path to its top-level family, and features per task is the mean number of distinct operation tuples required by one task. Software breadth is the number of distinct applications represented in the benchmark.

\paragraph{Display scaling in the benchmark comparison.} Agent score, empirical difficulty, boundary informativeness, capability demand, and capability--intensity coverage naturally lie on a 0--100 scale. Average steps are divided by the 50-step budget only for plotting. Software breadth and feature-operation breadth are absolute counts; in the paired bar figure, each count pair is independently scaled so that the larger bar has height 95 while preserving the exact ratio, and the unscaled counts are printed as labels. These pair-scaled bars must therefore not be compared vertically with percentage-valued metrics.

\paragraph{Validity-aware diagnostic outcomes.} Diagnostic observations are assigned one of six mutually exclusive classes. \emph{Pass}, \emph{partial}, and \emph{fail} correspond to scores in $[0.99,1]$, $(0.01,0.99)$, and $[0,0.01]$, respectively, provided the task and verifier are valid. \emph{Censored} denotes a zero-score run that exhausts the step budget; \emph{confounded} denotes a profile with a blocking nuisance factor; and \emph{invalid} denotes missing scores, infeasible profiles, invalid setup, or a missing/unready verifier. Only pass, partial, and fail are \emph{clean} observations eligible for matrix values and capability-boundary estimation. Partial-verifier observations remain eligible in the reported main figures when their observable subset is sound, but their coverage mode is retained as metadata.

\paragraph{Diagnostic matrix value.} Figure~\ref{fig:diagnostic-matrices-all-agents} pools eligible observations from $R0$ through $R3$ separately for each target agent. For a matrix cell $C$, let $\mathcal{E}_C$ be the clean episodes assigned to that cell and let $q_i\in[0,1]$ be episode $i$'s checkpoint score. The displayed value is $M_C=100\sum_{i\in\mathcal{E}_C}w_{i,C}q_i/\sum_{i\in\mathcal{E}_C}w_{i,C}$. Primary-dimension evidence has weight one and supporting-dimension evidence also uses weight one in the reported figures; the implementation permits a lower supporting weight for sensitivity analysis. Consequently, the reported setting is the task-weighted mean over all contributing episodes, not an average of per-round means. For example, a displayed value of $75\%$ means that clean episodes associated with that cell received an average normalized checkpoint score of $0.75$; it does not mean that 75\% of them fully passed. A pass-rate variant replaces $q_i$ with $\mathbf{1}[q_i\geq0.99]$ but is not used for the main matrices.

\paragraph{Capability--intensity matrix.} Rows are the seven capability dimensions and columns are task-demand intensities 1--5. A task contributes to its primary dimension and to every supporting dimension with a positive recorded intensity. Thus, a cell answers the question ``How well did the agent perform on tasks that required capability $d$ at intensity $z$?'' It does not estimate the causal effect of increasing intensity, because tasks in adjacent columns may differ in software, feature, and lineage. Intensity zero is omitted because it denotes no meaningful demand on that dimension.

\paragraph{Software--capability matrix.} Rows are applications and columns are capability dimensions. A task contributes to every positive capability demand recorded in its profile, using the same primary/supporting attribution rule as the capability--intensity matrix. The value therefore summarizes performance on tasks coupling a particular application with a particular capability demand. It is not a software-only score and does not imply that failures are uniquely caused by that capability.

\paragraph{Software--feature matrix.} Each row is an application and each cell corresponds to one top-level taxonomy feature family required by at least one profile. All required features are included, with primary and supporting roles weighted equally; leaf operations are collapsed to the top-level family only for readability. The printed percentage is the rounded mean checkpoint score over clean tasks exposing that software--family pair. A task requiring multiple feature families contributes its episode score to each corresponding cell.

\paragraph{Software--action matrix.} Columns represent required interaction families such as click, type, shortcut, menu navigation, dialog entry, numeric-widget manipulation, drag, and scroll. The matrix records planned action exposure from the task profile, not actions inferred from a trajectory. Its color is the mean whole-task checkpoint score for tasks requiring that action family; it is therefore not an action-localized success rate. One task may contribute to several action columns, and $R0$ contributes no action cells because the original tasks do not have compatible action-exposure annotations.

\paragraph{Missing cells and evidence maturity.} In the compact paper matrices, a gray cell means that no eligible clean score is available; it may be genuinely unexposed or may have planned profiles whose rollouts are censored, confounded, invalid, or missing. The annotated companion plots separate these cases: $P$ is the number of designed profiles assigned to a cell, $V$ is the number of clean rollout observations, gray denotes $P=0$, and light blue denotes $P>0$ but $V=0$. We additionally report three coverage layers. \emph{Exposed} means that at least one task profile occupies the cell, \emph{valid} means that at least one clean observation occupies it, and \emph{well-sampled} requires at least three clean observations drawn from at least two distinct base-task families.

\paragraph{Matched $R0$ reconstruction.} The original OSWorld tasks were not authored with diagnostic primary-dimension or action labels. For each agent, the $R0$ capability axis is reconstructed from the one-to-one $R1$ profile: we use the dimension selected as the $R1$ primary probe and that task's recorded source-demand intensity. The full source-demand vector supplies supporting dimensions, while feature cells come from the curated $R0$ taxonomy sidecar. This produces a matched baseline with one observation per lineage, but it should not be interpreted as an independently annotated $R0$ diagnostic plan. No $R0$ action matrix is reconstructed.

\paragraph{Diagnostic-resolution metrics.} The separate diagnostic-resolution plots use manual task-level judgments and explicitly exclude checkpoint scores. Only judgments marked eligible for capability analysis are included. Boundary-region density is the fraction of scores strictly between 20 and 80. Upward-probe transition yield is the fraction of clean-pass source tasks whose higher-intensity successors become partial or fail. A strict empirical boundary bracket occurs when an upward probe changes pass to partial/fail or a downward probe changes partial/fail to pass; bracket width is the absolute difference between the two capability intensities, and a width-one event localizes the boundary to adjacent levels. These quantities evaluate adaptive resolution rather than overall task difficulty.

\paragraph{Verifier metrics.} Requirement coverage is the number of atomic task requirements grounded in executable checkpoints divided by the total number of extracted requirements. A \textsc{Full} verifier covers every requirement, whereas a \textsc{Partial} verifier preserves a nonempty sound observable subset and records the remaining gaps; \textsc{Blocked} tasks have no meaningful observable checkpoint and are not rolled out. Checkpoint count is the number of callable checkpoint functions and is reported both as a total and as a per-task mean. \texttt{READY} indicates that the verifier passed syntax, interface-contract, schema-grounding, and static validation gates; it is an implementation-validity status, while empirical PRE/PASS and negative-fixture tests provide the behavioral discrimination evidence.

\paragraph{Uncertainty estimates.} For paired $R0$--$R4$ benchmark comparisons, confidence intervals are obtained by 5,000 task-level bootstrap resamples. Intervals for changes resample paired per-task differences, preserving lineage correspondence across rounds. Diagnostic matrices report cell means together with evidence counts or evidence-maturity layers rather than confidence intervals because many cells are intentionally sparse during adaptive exploration.

\subsection{Prompt Templates}
Figure~\ref{fig:rollout-prompt} illustrates the unified agent--environment rollout prompt contract shared across all evaluated models.

\definecolor{RolloutBlue}{HTML}{1F5AA6}
\definecolor{RolloutBlueLight}{HTML}{F3F8FE}
\definecolor{RolloutTeal}{HTML}{0F8B8D}
\definecolor{RolloutTealLight}{HTML}{F1FAFA}
\definecolor{RolloutOrange}{HTML}{D97706}
\definecolor{RolloutOrangeLight}{HTML}{FFF7E8}
\definecolor{RolloutPurple}{HTML}{6D4CB3}
\definecolor{RolloutPurpleLight}{HTML}{F7F4FC}
\definecolor{RolloutInk}{HTML}{172033}
\definecolor{RolloutMuted}{HTML}{5E6879}

\tcbset{
  rollout/card/.style={
    enhanced,
    boxrule=0.55pt,
    arc=1.3mm,
    left=1.5mm,right=1.5mm,top=1.1mm,bottom=1.1mm,
    fonttitle=\bfseries\sffamily\scriptsize,
    coltitle=white,
    before skip=0pt,after skip=0pt,
  },
  rollout/field/.style={
    enhanced,
    boxrule=0.45pt,
    arc=1.1mm,
    colback=white,
    colframe=RolloutBlue!45,
    left=1.2mm,right=1.2mm,top=1.0mm,bottom=0.9mm,
    fonttitle=\bfseries\sffamily\fontsize{6.2}{7.2}\selectfont,
    coltitle=RolloutBlue,
    attach boxed title to top left={xshift=1.0mm,yshift=-1.2mm},
    boxed title style={
      colback=white,colframe=white,boxrule=0pt,
      left=0.4mm,right=0.4mm,top=0pt,bottom=0pt
    },
  }
}

\begin{figure}[t]
\centering
\begin{tcolorbox}[
  enhanced,
  width=\columnwidth,
  colback=white,
  colframe=RolloutBlue,
  boxrule=0.8pt,
  arc=1.8mm,
  left=2.0mm,right=2.0mm,top=2.8mm,bottom=2.0mm,
  title={\sffamily\bfseries\footnotesize Agent--Environment Rollout Prompt},
  fonttitle=\sffamily\bfseries\footnotesize,
  coltitle=white,
  colbacktitle=RolloutBlue,
  attach boxed title to top left={xshift=2.4mm,yshift=-2.0mm},
  boxed title style={
    enhanced,arc=1.1mm,boxrule=0pt,
    colback=RolloutBlue,left=1.8mm,right=1.8mm,top=0.8mm,bottom=0.8mm
  }
]

\vspace{0.8mm}
\begin{tcolorbox}[
  rollout/card,
  title={\strut ROLE},
  colback=RolloutBlueLight,
  colframe=RolloutBlue,
  colbacktitle=RolloutBlue
]
\footnotesize
You are a computer-use agent. Given a task instruction, the current visual
observation, and the preceding interaction history, produce the next executable
action block that advances the task in the desktop environment.
\end{tcolorbox}

\vspace{1.4mm}
\begin{tcbraster}[
  raster columns=3,
  raster equal height=rows,
  raster column skip=1.2mm
]
\begin{tcolorbox}[rollout/field,title={TASK}]
\scriptsize\raggedright
\textbf{$\mathcal{I}$:} User-facing instruction \& target state.
\end{tcolorbox}
\begin{tcolorbox}[rollout/field,title={OBSERVATION}]
\scriptsize\raggedright
\textbf{$o_t$:} Current screenshot \& UI context.
\end{tcolorbox}
\begin{tcolorbox}[rollout/field,title={HISTORY}]
\scriptsize\raggedright
\textbf{$h_t$:} Prior observations \& executed actions.
\end{tcolorbox}
\end{tcbraster}

\vspace{1.3mm}
\begin{tcolorbox}[
  rollout/card,
  title={\strut ACTION POLICY},
  colback=RolloutTealLight,
  colframe=RolloutTeal,
  colbacktitle=RolloutTeal
]
\footnotesize
\begin{enumerate}[leftmargin=1.35em,itemsep=0.18em,parsep=0pt,topsep=0.1em]
  \item Ground the next action in $(\mathcal{I},o_t,h_t)$; do not assume an
        unobserved interface state.
  \item Emit an executable model-native action block
        $a_t\!\in\!\mathcal{A}_m$ (e.g., click, drag, scroll, type, hotkey,
        wait, or an enabled GUI/CLI program).
  \item If repeated execution leaves the screen unchanged, revise the action
        or choose an alternative interaction path.
\end{enumerate}
\end{tcolorbox}

\vspace{1.3mm}
\begin{tcolorbox}[
  rollout/card,
  title={\strut RESPONSE CONTRACT},
  colback=RolloutPurpleLight,
  colframe=RolloutPurple,
  colbacktitle=RolloutPurple
]
\footnotesize
\begin{tabular}{@{}p{0.20\linewidth}p{0.75\linewidth}@{}}
\textbf{Rationale} & $\rho_t$: concise description of intended state transition.\\[0.2em]
\textbf{Action} & $a_t$: one model-native tool call or action block.\\[0.2em]
\textbf{Status} & $z_t\in\{\textsc{Continue},\textsc{Success},$\\
& $\textsc{Failure},\textsc{Infeasible}\}$.
\end{tabular}

\vspace{0.5mm}
\textcolor{RolloutMuted}{\scriptsize Success is emitted only after the requested
state is achieved; infeasibility is reserved for tasks unexecutable under available software, state, or tool constraints.}
\end{tcolorbox}

\vspace{1.3mm}
\begin{tcolorbox}[
  enhanced,
  colback=RolloutOrangeLight,
  colframe=RolloutOrange!65,
  boxrule=0.5pt,
  arc=1.2mm,
  left=0.8mm,right=0.8mm,top=1.0mm,bottom=1.0mm
]
\centering
\resizebox{0.98\linewidth}{!}{%
\begin{tikzpicture}[
  baseline=(current bounding box.center),
  node distance=2.4mm,
  every node/.style={font=\sffamily\scriptsize},
  stage/.style={
    draw=RolloutOrange!65,fill=white,rounded corners=1.0mm,
    minimum height=4.8mm,inner xsep=1.6mm
  },
  flow/.style={-{Latex[length=1.5mm]},semithick,draw=RolloutOrange}
]
\node[stage] (observe) {Observe $o_t$};
\node[stage,right=of observe] (reason) {Condition $(\mathcal{I},h_t)$};
\node[stage,right=of reason] (act) {Execute $a_t$};
\node[stage,right=of act] (transition) {Transition};
\draw[flow] (observe) -- (reason);
\draw[flow] (reason) -- (act);
\draw[flow] (act) -- (transition);
\draw[flow,rounded corners=1.8mm]
  (transition.south) -- ++(0,-2.7mm) -| (observe.south)
  node[pos=0.48,below=0.2mm,text=RolloutMuted] {$o_{t+1}$ until terminal status};
\end{tikzpicture}%
}
\end{tcolorbox}

\vspace{0.5mm}
\textcolor{RolloutMuted}{\scriptsize\textit{Implementation note.}
Task, observation, history, and termination semantics are shared across agents;
only low-level serialization of $\mathcal{A}_m$ follows each model's native
interface. Sensitive credentials and verbose schemas are omitted.}
\end{tcolorbox}
\caption{Model-agnostic prompt contract used for agent--environment rollouts.
At every step, the agent conditions on the same semantic inputs and returns a
native executable action together with an explicit continuation or terminal
status.}
\label{fig:rollout-prompt}
\end{figure}

\end{document}

%% file: preamble.tex
\usepackage{float}
\usepackage{algorithm}
\usepackage{algorithmic}

\usepackage{multirow}
\usepackage[most]{tcolorbox}
\usepackage{tikz}
\usepackage{textcomp} 
  \usepackage{makecell}

\input{case/evolution-case-style}

\newcommand{\Qwenemoji}{\includegraphics[height=1.5\fontcharht\font`\B]{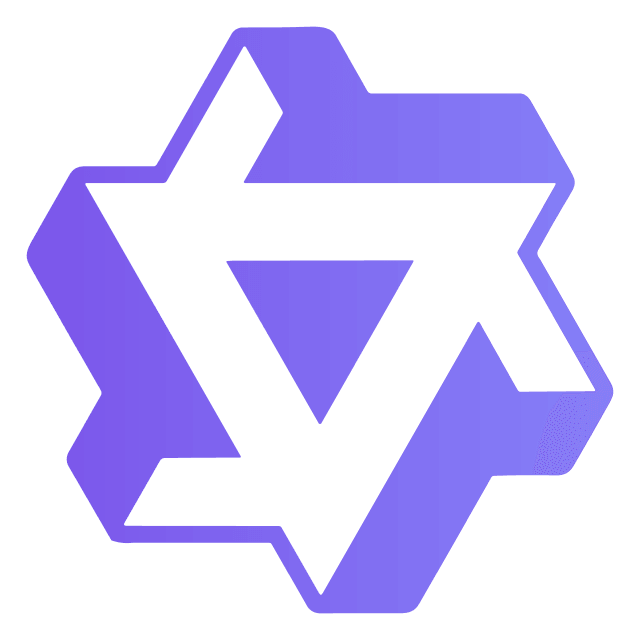}}
\newcommand{\Geminiemoji}{\includegraphics[height=1.5\fontcharht\font`\B]{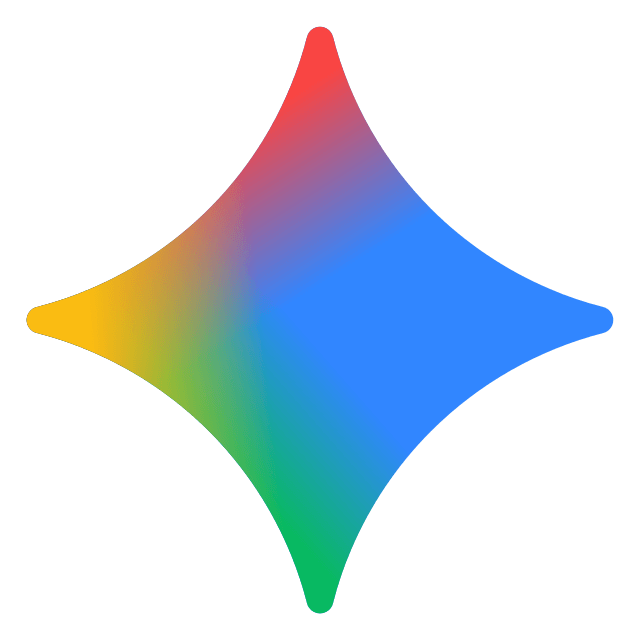}}
\newcommand{\Openaiemoji}{\includegraphics[height=1.5\fontcharht\font`\B]{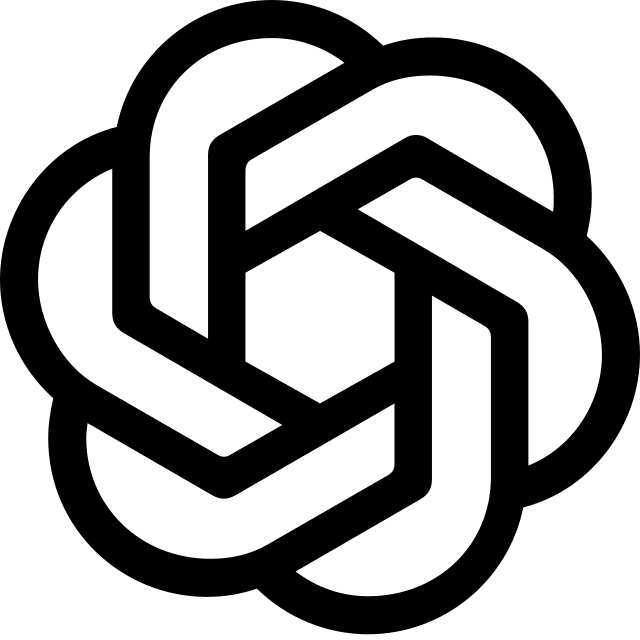}}
\newcommand{\kimimoji}{\includegraphics[height=1.5\fontcharht\font`\B]{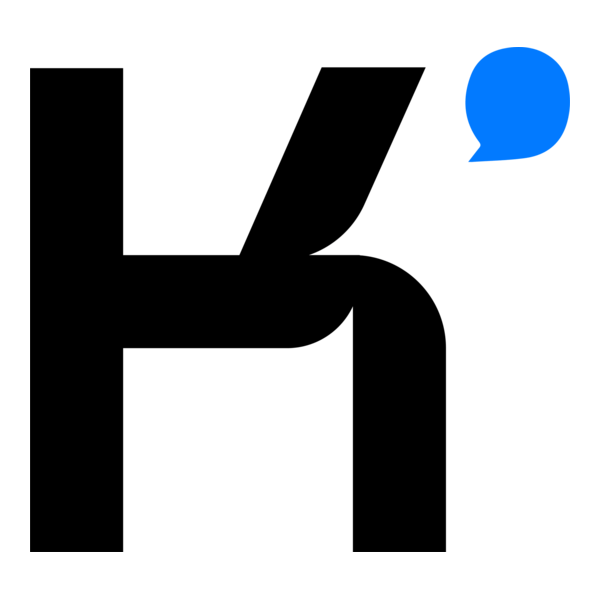}}
\newcommand{\seedmoji}{\includegraphics[height=1.5\fontcharht\font`\B]{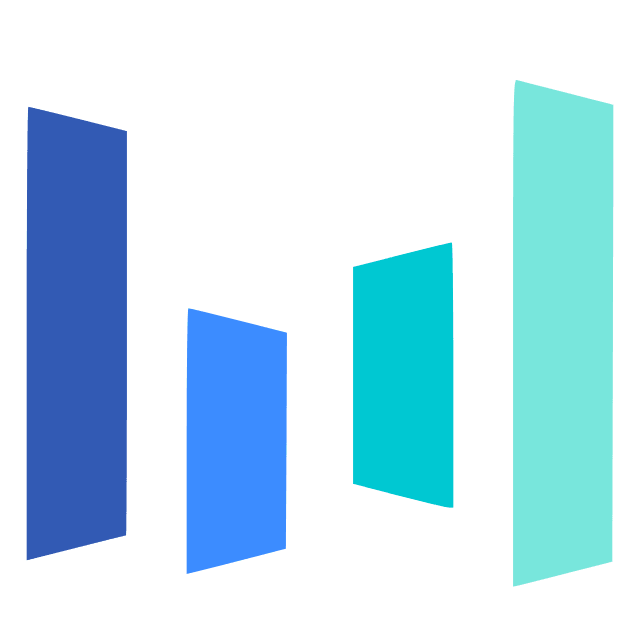}}

\newcommand{\cmark}{\textcolor{green!60!black}{\checkmark}}
\newcommand{\xmark}{\textcolor{red!75!black}{\ensuremath{\times}}}

\usetikzlibrary{arrows.meta,positioning}

\usepackage{adjustbox}

%% file: case/evolution-case-style.tex
\usepackage{xcolor}
\usepackage{graphicx}
\usepackage{tikz}
\usetikzlibrary{arrows.meta,positioning,calc}
\usepackage[most]{tcolorbox}
\usepackage{tabularx,booktabs,array}

\definecolor{evoNavy}{HTML}{123A70}
\definecolor{evoBlue}{HTML}{1F67C1}
\definecolor{evoBlueFill}{HTML}{EEF5FF}
\definecolor{evoTeal}{HTML}{157A78}
\definecolor{evoTealFill}{HTML}{ECF8F6}
\definecolor{evoOrange}{HTML}{D97706}
\definecolor{evoOrangeFill}{HTML}{FFF5E6}
\definecolor{evoPurple}{HTML}{6D4BC3}
\definecolor{evoPurpleFill}{HTML}{F3EFFF}
\definecolor{evoInk}{HTML}{172033}
\definecolor{evoMuted}{HTML}{5F6B7A}